\documentclass[journal]{IEEEtran}
\usepackage{graphicx}
\usepackage{subcaption}
\usepackage{multirow}
\usepackage[table]{xcolor}
\usepackage{color}
\usepackage{colortbl}
\usepackage{tabularx}
\usepackage{bm}
\usepackage{booktabs}
\usepackage{url}
\usepackage{stackengine}
\usepackage{float}
\usepackage{verbatim}
\usepackage{array}
\usepackage{cite}
\usepackage{algorithm}
\usepackage{algorithmic}
\usepackage[colorlinks,linkcolor=blue]{hyperref}
\usepackage{amsmath,amssymb} 

\begin{document}
\title{Bi-Temporal Semantic Reasoning for the Semantic Change Detection in HR Remote Sensing Images}

\author{Lei~Ding, Haitao Guo, Sicong Liu,~\IEEEmembership{Member,~IEEE,}~ Lichao Mou, Jing Zhang and Lorenzo~Bruzzone,~\IEEEmembership{Fellow,~IEEE}~

\thanks{L. Ding, J. Zhang, and L. Bruzzone are with the Department
of Information Engineering and Computer Science, University of Trento,
38123 Trento, Italy (E-mail: lei.ding@unitn.it, jing.zhang-1@studenti.unitn.it, lorenzo.bruzzone@unitn.it).}

\thanks{H. Guo is with the Geospatial Information Institute, Strategic Support Force Information Engineering University, Zhengzhou, China (E-mail: ghtgjp2002@163.com).}

\thanks{S. Liu is with the College of Surveying and Geoinformatics, Tongji University, No.1239 Siping Road, Shanghai, China (E-mail: sicongliu.rs@gmail.com).}

\thanks{L. Mou is with the Remote Sensing Technology Institute, German Aerospace Center, and Data Science in Earth Observation, Technical University of Munich, Germany (E-mail: lichao.mou@dlr.de).}

\thanks{This document is funded by the National Natural Science Foundation of China (No. 41876105, 41671410, 42071324). It is also funded by the scholarship from China Scholarship Council (grant NO.201703170123).}}

\markboth{Manuscript under review.}%
{Shell \MakeLowercase{\textit{et al.}}: Bare Demo of IEEEtran.cls for IEEE Journals}

\maketitle

\begin{abstract}
Semantic change detection (SCD) extends the multi-class change detection (MCD) task to provide not only the change locations but also the detailed land-cover/land-use (LCLU) categories before and after the observation intervals. This fine-grained semantic change information is very useful in many applications. Recent studies indicate that the SCD can be modeled through a triple-branch Convolutional Neural Network (CNN), which contains two temporal branches and a change branch. However, in this architecture, the communications between the temporal branches and the change branch are insufficient. To overcome the limitations in existing methods, we propose a novel CNN architecture for the SCD, where the semantic temporal features are merged in a deep CD unit. Furthermore, we elaborate on this architecture to reason the bi-temporal semantic correlations. The resulting Bi-temporal Semantic Reasoning Network (Bi-SRNet) contains two types of semantic reasoning blocks to reason both single-temporal and cross-temporal semantic correlations, as well as a novel loss function to improve the semantic consistency of change detection results. Experimental results on a benchmark dataset show that the proposed architecture obtains significant accuracy improvements over the existing approaches, while the added designs in the Bi-SRNet further improves the segmentation of both semantic categories and the changed areas. The codes in this paper are accessible at: \href{github.com/ggsDing/Bi-SRNet}{\textit{https://github.com/ggsDing/Bi-SRNet}}.
\end{abstract}




\section{Introduction}\label{sc1}

Change detection (CD) refers to the task of identifying the areas in remote sensing images (RSIs) where changes have occurred during the observation intervals \cite{bovolo2015time}. CD is useful for various kinds of real-world applications, such as urban management, environment monitoring, crop monitoring and damage assessment. Although binary change detection (BCD) algorithms allow us to automatically monitor and analyze the region of interests in RSIs, the information provided is coarse-grained and does not describe the detailed change types. In many applications we are interested in not only 'where' the changes occurred, but also 'what' are the changes. To overcome this limitation, multi-class change detection (MCD) techniques \cite{Bovolo2007p-CVA, Bovolo2012C2VA, bovolo2015time} and approaches to the detection of LCLU transitions~\cite{bruzzone2004detection, Bruzzone2000diff, bruzzone1999neural, bruzzone1997iterative} have been presented in the literature. They have been referred as Semantic Change Detection (SCD) in recent literature~\cite{kataoka2016semantic, daudt2019multitask} and provide not only the change information, but also the detailed LCLU maps before and after the change events. This allows representation of richer and more complex semantic change information.

Recently with the development of Convolutional Neural Networks (CNNs) \cite{long2015fully}, great improvements have been achieved in terms of CD. Instead of extracting difference information (which is the common practice in statistical and image processing methods), CNNs learn to directly segment multi-temporal images \cite{peng2019end}. CNNs typically have a hierarchy bottom-up design, where the bi-temporal features are embedded and down-scaled through stacked convolutional layers. The change information is modelled through weighted combination and transformation of the features. Compared with statistical and image processing methods, CNN-based methods have the advantages of: i) Improved robustness. The CNN-based CD methods extract numerous features while being free of hyper-parameters (such as weights and thresholds), thus can stably process large volumes of data; ii) Modelling more complex changes. CNNs can learn to model some complex change types which can not be well described by hand-crafted features.

However, the CNN-based SCD of RSIs has been rarely studied in existing works. In the perspective of image processing, BCD is essentially a binary segmentation task where a binary map is produced to represent the changed/ unchanged regions. However, the SCD is a complex task containing two underlying sub-tasks: i) Semantic segmentation (SS) of the LCLU classes. It is required to segment the bi-temporal semantic labels of either all the RSI or the changed areas; ii) binary CD of the changed areas. Therefore, the results of SCD should be either two temporal LCLU maps and a change map \cite{daudt2019multitask} or two semantic change maps \cite{yang2020asymmetric}. Previous CNN-based CD methods may not be suitable for the SCD, since they typically contain only a single branch to embed the difference features \cite{daudt2018fully}. Although several recent works have proposed task-specific methods for the SCD \cite{daudt2019multitask, yang2020asymmetric}, they are based on a triple-branch architecture where the sub-tasks are separately modelled. Moreover, the intrinsic correlations between the two sub-tasks have not been considered. 

To fill the research gap in SCD, in this paper, we exploit the spatial and temporal semantic correlations in SCD to improve the accuracy. The major contributions in this paper are as follows:

\begin{enumerate}
    \item Proposing a novel CNN-based architecture for the SCD. The sub-tasks in SCD (SS and CD) are disentangled, whereas their features are shared and are deeply fused. The loss functions are also disentangled to supervise respectively SS and CD in the SCD. The resulting architecture shows significant accuracy improvements over existing approaches;
    \item Proposing a Bi-temporal Semantic Reasoning Network (Bi-SRNet) for the SCD. Building on top of the novel CNN architecture, the Bi-SRNet further integrates i) two Siamese Semantic Reasoning (Siam-SR) blocks to model the semantic information in each temporal branch; ii) a Cross-temporal SR (Cot-SR) block to model the temporal correlations, and iii) a Semantic Consistency Loss (SCLoss) function to align the semantic and change representations. These designs are verified in an ablation study, whereas the resulting Bi-SRNet is evaluated in comparisons with state-of-the-art (SOTA) methods.
\end{enumerate}

The remainder of this paper is organized as follows. Section \ref{sc2} introduces the literature works on CD of RSIs. Section \ref{sc3} elaborates on the proposed CNN architecture, as well as the Bi-SRNet. Section \ref{sc4} describes the experimental settings and the evaluation metrics. Section \ref{sc5} reports results of the ablation study and comparative experiments. Section \ref{sc6} summarizes this work and draws the conclusions.
\section{Related Work}\label{sc2}
This section is organized following the development of CD methods. The pre-CNN and CNN-based methods are separately introduced. Recent methods for the SCD are also reviewed.

\subsection{Expert knowledge-based Change Detection}

In the past decades, CD techniques have experienced a rapid development due to the increasing availability of remote sensing images, the importance of CD applications, and the evolution of machine learning. There are many excellent review works in literature (e.g.,~\cite{singh1990digital,lu2004change,Ban2016,Liu2019reviewHSI}), focusing on the analyzing of typical CD problems and corresponding methods. Among the traditional statistical and image processing unsupervised CD techniques, in \cite{Bruzzone2000diff} an approach based on the expectation-maximization (EM) algorithm and Markov Random Fields (MRF) was proposed for automatically solving binary CD problems by analyzing a difference image generated by Change Vector Analysis (CVA). CVA \cite{malila1980change, Bovolo2007p-CVA} and its different variation versions (e.g., \cite{Bovolo2012C2VA, liu2015sequential,Liu2017M2C2VA}) starting from a theoretical definition of the CD problem, provided different successful solutions for CD in multispectral and hyperspectral images also considering the detection of multiple changes. Another popular transformation-based unsupervised CD approach is the Multivariate Alteration Detection (MAD) \cite{Nielsen1998MAD} and its iterative reweighted version (IR-MAD) \cite{Nielsen2007IRMAD}, which exploits the nature of changes in multiple spectral bands. For the supervised CD, approaches are usually designed by taking advantages from robust supervised classifiers such as the Support Vector Machine (SVM), the Random Forest (RF) and the Extreme Learning Machine (ELM) in order to achieve better CD performance with a high accuracy \cite{nemmour2010support, Wang2018RS,Wang2020RS}. Other attempts developed from the semi-supervised perspective combined the merits of both unsupervised and supervised methods in order to improve the automation and robustness of the CD process \cite{Zhang2018semi-CD,liu2017novel}. On the other hand, CD performance can be also enhanced by considering different features, for example the spectral-spatial features \cite{Liu2019bexpand}, the kernel-based features \cite{Volpi2012kernel}, and the target-driven features \cite{Du2012IGARSS}. However, the conventional hand-designed features usually represent the low-level descriptions of the change objects, whereas the use of deep learning based approaches provides the capability to learn more complex and effective high-level change features from the data.

\subsection{CNN-based Change Detection}

The development of convolutional neural networks (CNNs) in the filed of computer vision provides insights into CD. By modelling the CD in hyper/multi-spectral imagery as a classification task, \cite{ZhangL19} presents a 2D CNN to learn spectral-spatial feature representation and \cite{mou2018learning} proposes a recurrent CNN that is able to learn spectral-spatial-temporal features and produce accurate results. Furthermore, in \cite{song2018change}, the authors introduce 3D CNN. For high spatial resolution remote sensing images, CD is usually deemed as a dense prediction problem and solved by semantic segmentation CNNs. For instance, in~\cite{peng2019end} an improved UNet++ is designed for the CD. \cite{ZhangYue20} presents a network with a hierarchical supervision. In \cite{zhan2017change}, the authors introduce a Siamese CNN to extract features of bi-temporal images and utilize a weighted contrastive loss to alleviate the influence of imbalanced data. In \cite{zhang2020deeply} and \cite{zhang2020feature}, Siamese CNNs are employed to extract bi-temporal features, while CNN decoders are designed to fuse the features and to learn change information. To tackle pseudo changes caused by seasonal transitions in CD tasks, \cite{ZhaoMCBE20} proposes a metric learning-based generative adversarial network (GAN), termed MeGAN, to learn seasonal invariant feature representations. In addition, some works recently focused on devising novel CNN architectures for unsupervised CD \cite{saha2019unsupervised,li2019deep,li2019unsupervised,yang2019transferred,wang2019getnet,yuan2018robust,hou2017change}. For example, \cite{saha2019unsupervised} proposes an unsupervised deep learning-based change vector analysis model for MCD in very high resolution images. In \cite{saha2020building} this method is also used for CD in SAR images, after converting them into optical-like features. The authors of \cite{hou2017change} incorporate deep neural networks and low-rank decomposition for predicting saliency maps where high values indicate large change probabilities. Another important branch in CNN for CD is heterogeneous CD, also known as multimodal CD, which aims to detect changes between heterogeneous images. In \cite{luppino2021deep}, the authors propose two novel network architectures using an affinity-based change prior learnt from the input data for heterogeneous CD. \cite{LiuGQZ18} introduces a CNN entitled symmetric convolutional coupling network (SCCN) for CD in heterogeneous optical and SAR images. In \cite{li2021deep}, the optical images are translated into the SAR domain to reduce differences, before performing CD with the SAR images.

\subsection{Semantic Change Detection}

Binary CD produces as outputs binary maps to represent the change information, which are often not enough informative in practical applications. In many applications there are multiple change types, such as seasonal changes, urbanization, damages, deforestation and pollution. To elaborate the change information, the MCD~\cite{bruzzone1997iterative, liu2015sequential} not only detects the changes, but also classifies the LCLU transition types. In \cite{singh1990digital} an intuitive method to the MCD, i.e., the post-classification comparison (PCC) is presented, which directly compares the LCLU classification maps to produce transition statistics. A major limitation of this approach is that it omits the temporal correlation between the two RSIs, as each LCLU classification map is produced independently. Multi-date direct classification (MDC) \cite{singh1990digital} overcomes this limitation by jointly training on the two RSIs. However, it requires multitemporal training data that model all possible LCLU transitions. To address this problem, in~\cite{bruzzone1997iterative, bruzzone1999neural, bruzzone2004detection, demir2011detection} the compound classification (CC) technique is introduced for MCD, which computes and maximizes the posterior joint probabilities of LC transitions. This method models the temporal dependence between two RSIs through an iterative semi-supervised estimation of the probability of transitions without requiring training data for each possible transition. One of the representative CNN-based methods for MCD is the work in \cite{mou2018learning} which integrates RNN units into CNN: the CNNs are employed to extract spatial information, whereas the RNN units detect multi-class changes from temporal features.

Recently there are CNN-based methods developed for SCD. In \cite{daudt2018fully} two SCD methods with triple embedding branches are introduced. Two of the branches segment temporal images into LCLU maps, while a CD branch detects the difference information. In \cite{yang2020asymmetric}, the triple-branch CNN is further extended by introducing gating and weighting designs in the decoders to improve the feature representations. In this work, a benchmark dataset for the SCD is also released together with task-specific evaluation metrics. However, two major problems remained: i) the bi-temporal LCLU features are separately embedded without considering their temporal coherence; and ii) frequent inconsistences occur between the detected changes and the segmented LCLU maps.
\section{Proposed Bi-temporal Semantic Reasoning Network (Bi-SRNet) for SCD}\label{sc3}
In this section we introduce the Bi-SRNet for SCD. First, we summarize the existing CNN approaches for the SCD and introduce a novel task-specific architecture. Then, we introduce the Bi-SRNet built on top of this proposed SCD architecture. Finally, we introduce the semantic reasoning designs and the loss functions in the Bi-SRNet.

\subsection{Task Formulation and Possible Approaches} \label{sc3.archs}

Before introducing the proposed approach, let us first define the task of semantic change detection (SCD) and its connections with the semantic segmentation (SS) and the binary change detection (BCD). Given an input image $I$, the task of SS is to find a mapping function $f_{s}$ that projects $I$ into a semantic map $P$:
\begin{equation}
    f_s(p_{i,j}) = c_{i,j}
\end{equation}
where $p_{i,j}$ is a pixel on $I$, $c_{ij}$ is the projected LCLU class. Meanwhile, a BCD function $f_{bcd}$ projects two temporal images $I_1, I_2$ into a binary change map $C$. For two image pixels $p_{1_{i,j}}, p_{2_{i,j}}$ on $I_1, I_2$ that are related to the same spatial location, the calculation is:
\begin{equation}
    f_{bcd}(p_{1_{i,j}}, p_{2_{i,j}}) =
    \begin{cases}
      0, & c_{1_{i,j}} = c_{2_{i,j}}\\
      1, & c_{1_{i,j}} \neq c_{2_{i,j}}
     \end{cases} 
\end{equation}
where the projected signal (0 or 1) reports if there is change in LCLU classes ($c_{1_{i,j}}$ and $c_{2_{i,j}}$) or other change types (e.g., damage and status change). The SCD function $f_{scd}$ is a combination of $f_s$ and $f_{bcd}$:
\begin{equation}
    f_{scd}(p_{1_{i,j}}, p_{2_{i,j}}) =
    \begin{cases}
      (0, 0), & c_{1_{i,j}} = c_{2_{i,j}}\\
      (c_{1_{i,j}}, c_{2_{i,j}}), & c_{1_{i,j}} \neq c_{2_{i,j}}
     \end{cases}
\end{equation}
It produces two semantic change maps $S_1$ and $S_2$ which indicate both the change areas and the bi-temporal LCLU classes. The required training labels in the SCD can be either i) two ground truth (GT) semantic change maps $L_1$ and $L_2$ \cite{yang2020asymmetric}, or alternatively ii) one GT binary CD map $L_c$ and two GT LCLU maps $L_{S_1}$and $L_{S_2}$ \cite{daudt2019multitask}. In the later , $L_1$ and $L_2$ can be easily generated by masking $L_{S_1}$ and $L_{S_2}$ with $L_c$.

An intuitive approach to the SCD is the post-classification comparison (PCC) \cite{singh1990digital}, which first classifies the LCLU maps and then compare the difference information. However, this approach has been proved sub-optimal since it neglects the temporal correlations and may cause accumulation of errors \cite{bruzzone1997iterative, daudt2019multitask}. Alternatively, we employ CNNs to learn directly the semantic changes, which is often referred as multidate direct classification (MDC) in MCD tasks \cite{bruzzone1997iterative}. The possible CNN-based approaches can be summarized as follows:

\begin{figure}[t]
\centering
        \subcaptionbox{}
        {\includegraphics[width=2.7cm]{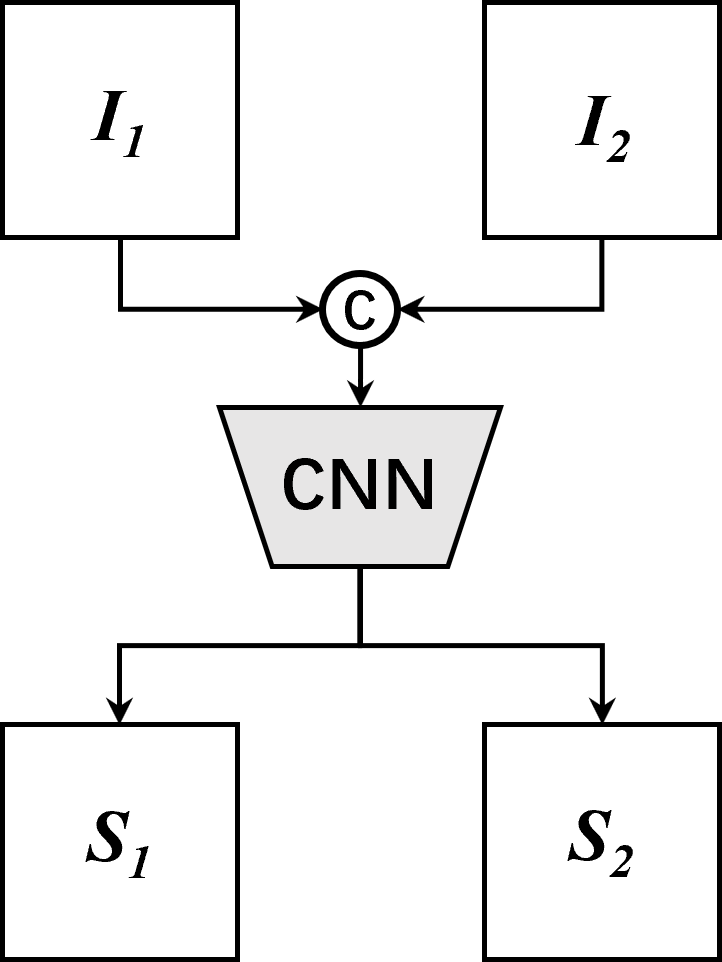}}
        \subcaptionbox{}
        {\includegraphics[width=3cm]{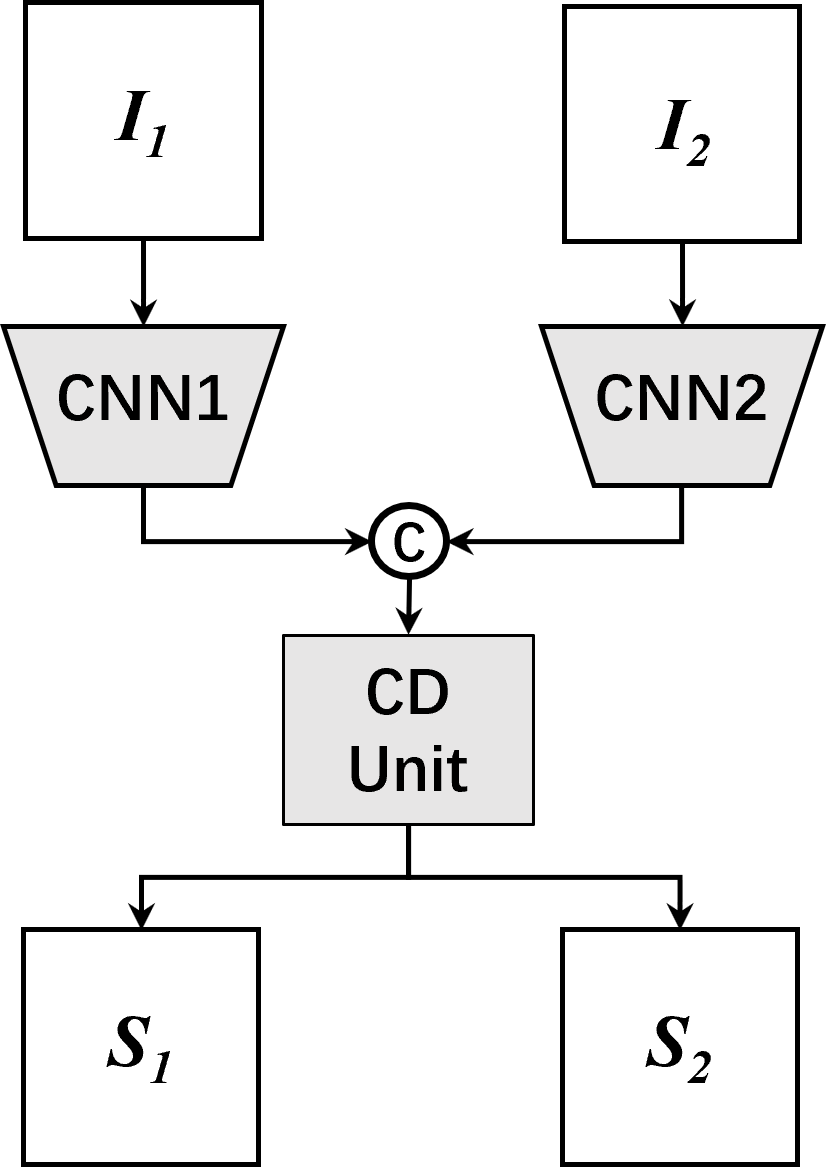}}\\
        \subcaptionbox{}
        {\includegraphics[height=4cm]{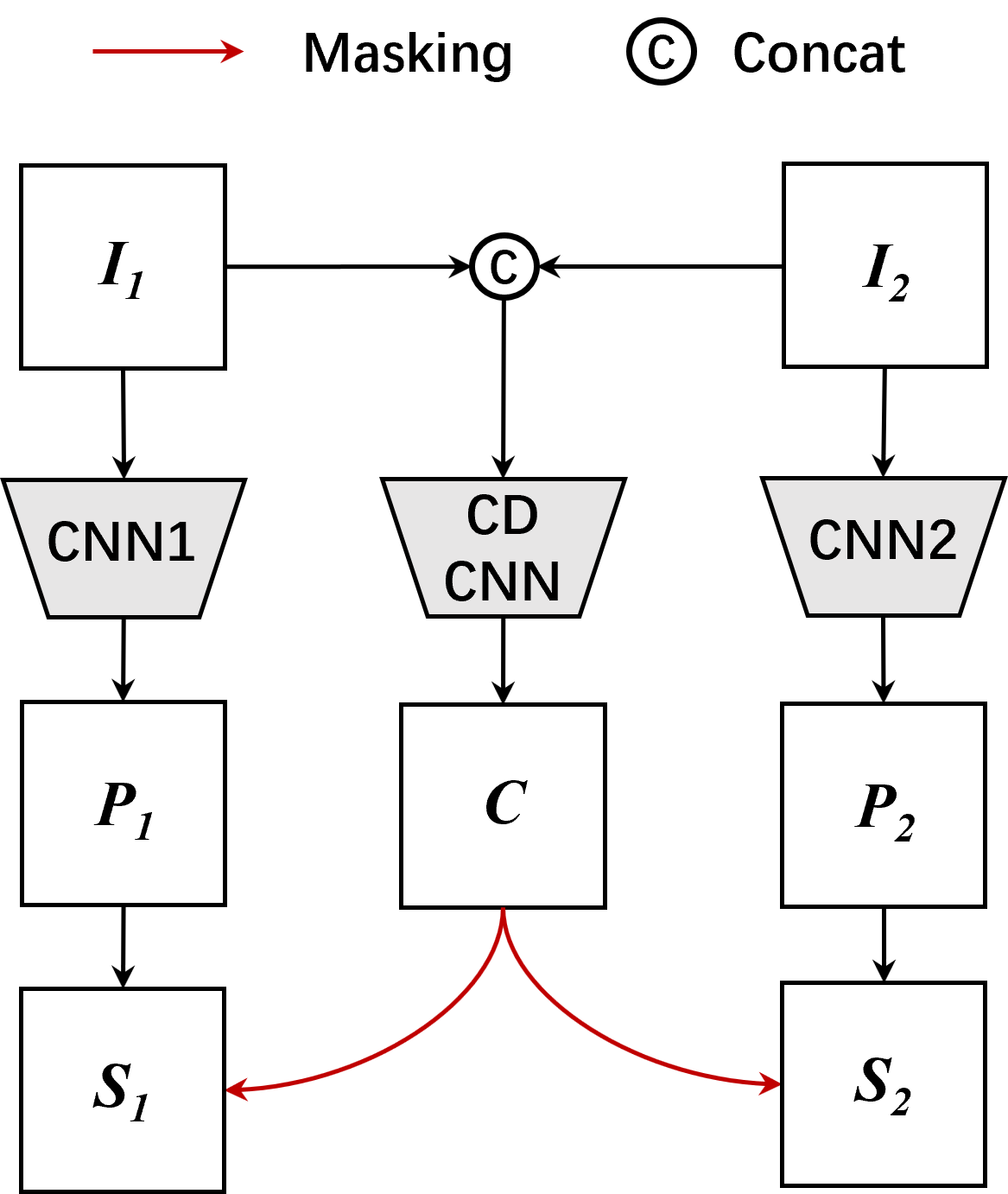}}
        \subcaptionbox{}
        {\includegraphics[height=4cm]{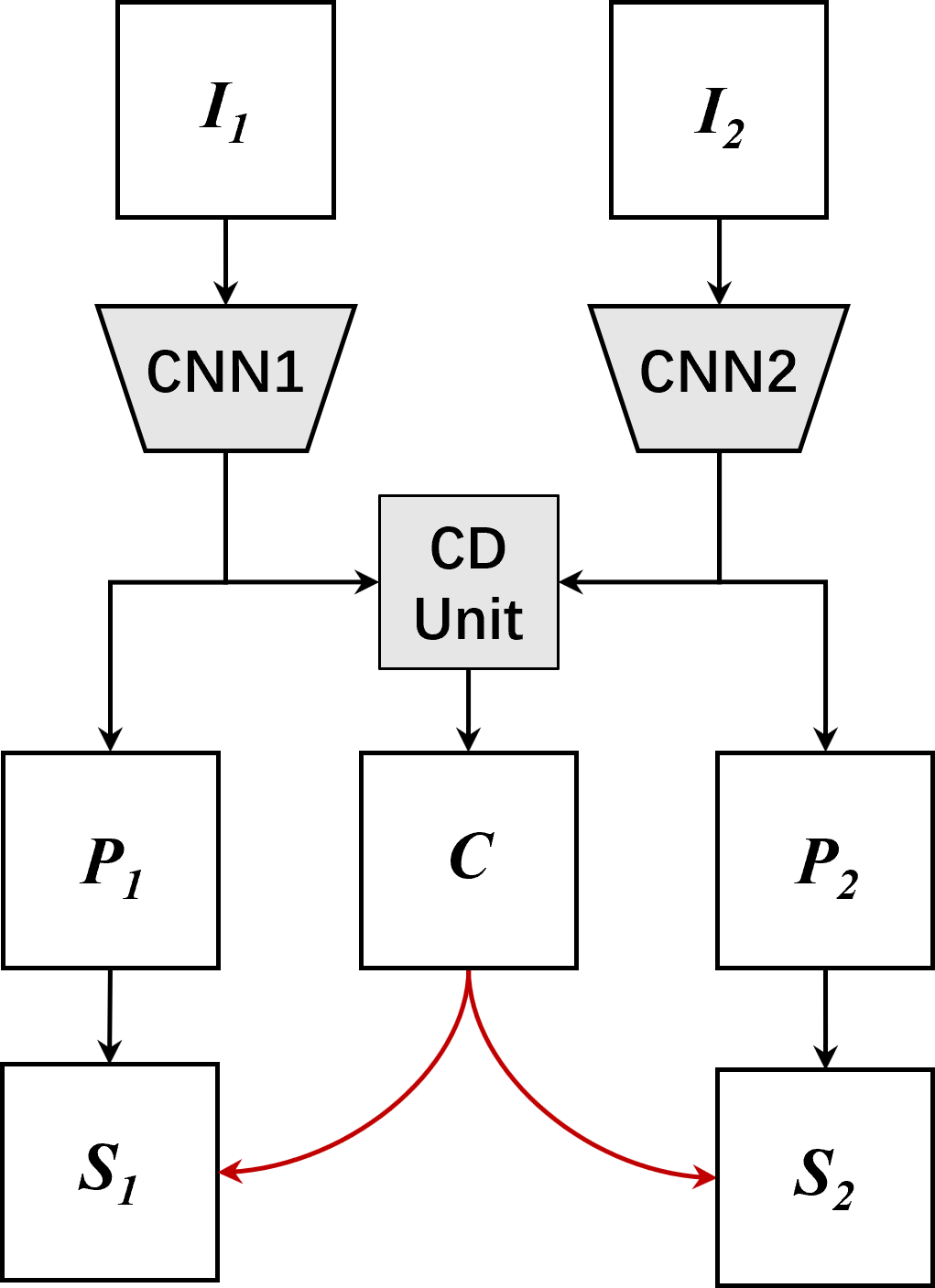}}
    \caption{Four possible CNN architectures for the SCD: (a) Direct SCD, early fusion (DSCD-e); (b) Direct SCD, late fusion (DSCD-l), (c) Disentangled SS and CD, early fusion (SSCD-e), and (d) Disentangled SS and CD, late fusion (SSCD-l).}\label{Fig.Archs}
\end{figure}

\textit{1) Direct SCD, early fusion} (DSCD-e, Fig.\ref{Fig.Archs}(a)). The temporal images $I_1$ and $I_2$ are concatenated as input data. A single CNN encoder $\mathcal{E}$ is employed to directly learn the $f_{scd}$:
\begin{equation}
    S_1, S_2 = \mathcal{E}(I_1, I_2).
\end{equation}
Many literature works that treat CD as a semantic segmentation task can be included into this category, including the FC-EF in \cite{daudt2018fully} and the UNet++ in \cite{peng2019end}. A major limitation of this architecture is that LCLU information in each temporal branch is not fully exploited \cite{daudt2019multitask}. Since the changed areas are minority, $\mathcal{E}$ is driven to pay more attention to the unchanged areas.

\textit{2) Direct SCD, late fusion} (DSCD-l, Fig.\ref{Fig.Archs}(b)). $I_1$ and $I_2$ are separately fed as inputs into two CNN encoders $\mathcal{E}_1$ and $\mathcal{E}_2$ (which can be weight-sharing, i.e., siamese \cite{daudt2018fully}, if $I_1$ and $I_2$ belong to the same domain). The encoded features are then fused and modelled through a convolutional CD unit $\mathcal{D}$. $\mathcal{E}_1$ and $\mathcal{E}_2$ serve as feature extractors, whereas the $f_{scd}$ is learned by $\mathcal{D}$ with the embedded semantic features:
\begin{equation}
    S_1, S_2 = \mathcal{D}[\mathcal{E}(I_1), \mathcal{E}(I_2)].
\end{equation}

This architecture may also include decoder networks and skip connections. In the FC-Siam-conc and the FC-Siam-diff \cite{daudt2018fully} the CD units are multi-scale concatenation blocks, whereas in the ReCNN\cite{mou2018learning} they are Recurrent Neural Networks (RNNs). However, in this architecture, the \textit{no-change} class is still competing with other classes during the network inference, which does not correctly reflect their intrinsic correlations.

\textit{3) Disentangled SS and CD, early fusion} (SSCD-e, Fig.\ref{Fig.Archs}(c)). Three CNN encoders $\mathcal{E}_1$, $\mathcal{E}_2$ and $\mathcal{E}_c$ are separately trained with $I_1$, $I_2$ and $(I_1, I_2)$ as the inputs. $\mathcal{E}_1$ and $\mathcal{E}_2$ can be siamese if $I_1$ and $I_2$ belong to the same domain. The semantic and change information are separately modelled. $\mathcal{E}_1$ and $\mathcal{E}_2$ produce the semantic maps $P_1$ and $P_2$, while $\mathcal{E}_c$ produces a binary change map $C$. $S_1, S_2$ are then generated by masking $P_1, P_2$ with $C$. The calculations are as follows:

\begin{gather}
    P_1, P_2, C= \mathcal{E}_1(I_1), \mathcal{E}_2(I_2), \mathcal{E}_c(I_1, I_2)\\
    S_1, S_2 = C \cdot (P_1, P_2).
\end{gather}

A representative of this architecture is the HRSCD-str.3 in \cite{daudt2019multitask}. However, a disadvantage of this architecture is that it neglects the temporal correlations, since there are limited communications between the temporal branches and the CD branch. Although there can be connections between $\mathcal{E}_1$, $\mathcal{E}_2$ and $\mathcal{E}_c$ (e.g., the skip-connections in the HRSCD-str.4 \cite{daudt2019multitask} and the gating operations in the ASN \cite{yang2020asymmetric}), they do not fully exploit the LCLU information in temporal features. Additionally, training the $f_{cd}$ from scratch with input images is not computation-efficient.

\textit{4) Disentangled SS and CD, late fusion} (SSCD-l, Fig.\ref{Fig.Archs}(c)). This is a novel architecture proposed to address the limitations of above-mentioned approaches. In the SSCD-l, two CNN encoders $\mathcal{E}_1$ and $\mathcal{E}_2$ are employed to extract semantic information from $I_1$ and $I_2$. The extracted semantic features are further merged to train a CD unit $\mathcal{D}$, which exploits the difference information. The calculations can be represented as follows:
\begin{gather}
    P_1, P_2 = \mathcal{E}_1(I_1), \mathcal{E}_2(I_2)\\
    C = \mathcal{D}[\mathcal{E}(I_1), \mathcal{E}(I_2)]\\
    S_1, S_2 = C \cdot (P_1, P_2).
\end{gather}

The advantages of the SSCD-l architecture are three-fold: i) Both the LCLU information and the change information are explicitly modelled. This enables the SSCD-l to embed more oriented features for the underlying sub-tasks (SS and CD); ii) Instead of producing two separate change maps (such as in the DSCD-e and DSCD-l), where there may be discrepancies, the SSCD-l learns to produce a single change map. The change information is therefore more consistent in the two masked temporal predictions $S_1$ and $S_2$; iii) Instead of extracting change features from scratch in the two RSIs (such as in the SSCD-e), the SSCD-l learns more semantic changes from the extracted features of the two temporal branches.

To find out the optimal CNN architecture for the SCD, experimental comparisons have been conducted in Sec.\ref{sc5.comareArch} and Sec.\ref{sc5.compareSOTA}.

\subsection{Bi-temporal Semantic Reasoning Network}

\begin{figure*}
\centering
    \includegraphics[width=1\linewidth]{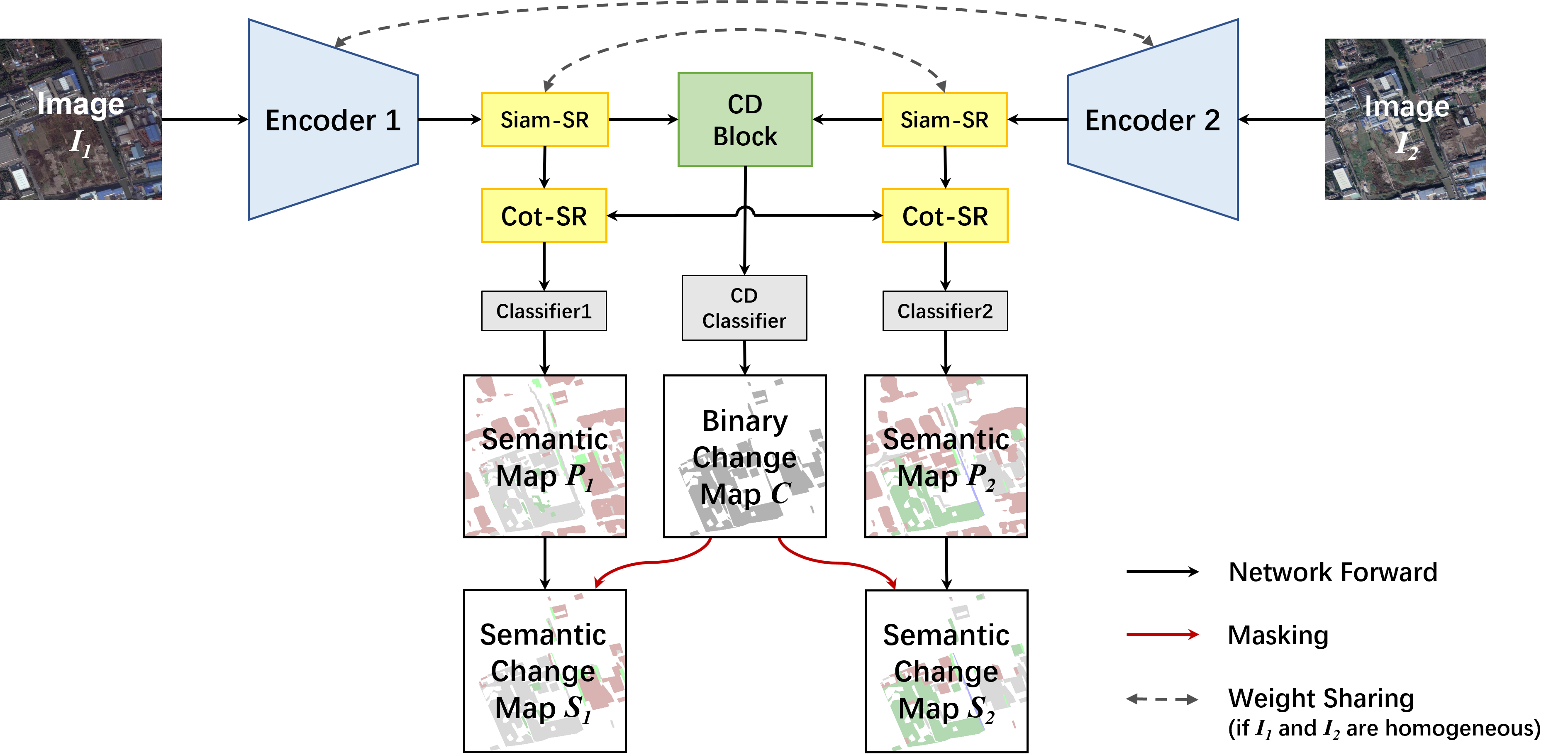}
    \caption{Architecture of the proposed \textbf{Bi}-temporal \textbf{S}emantic \textbf{R}easoning \textbf{Net}work (Bi-SRNet) for SCD. The modelling of semantic and changed information is disentangled to enhance the feature exploitation, whereas the semantic representations in the two temporal branches are aligned through the SR blocks.}
    \label{Fig.Overview}
\end{figure*}

In the proposed SSCD-l architecture, the semantic features are separately extracted through two temporal branches and are merged through a deep CD unit. However, one of the remaining problems is to model the temporal correlation and coherence between the two feature extraction branches in many applications. The majority of image regions remain unchanged through the observation intervals, thus they exhibit similar visual patterns. To better exploit this information, we propose the Bi-temporal Semantic Reasoning Network (Bi-SRNet) as illustrated in Fig.\ref{Fig.Overview}.

The Bi-SRNet is built on top of the SSCD-l architecture by introducing two extra Semantic Reasoning (SR) blocks \cite{mou2019relation, zhang2020multi} and a semantic consistency loss. Given 2 input temporal images $I_1$ and $I_2$, the Bi-SRNet first employs 2 CNN encoders $\mathcal{E}_1$ and $\mathcal{E}_2$ to extract the semantic features $X_1$ and $X_2$. Differently from the SSCD-l, $X_1$ and $X_2$ are then processed by two Siamese SR (Siam-SR) blocks to improve their semantic representations. Under the circumstance that there is no significant domain difference, the weights of both $\mathcal{E}_1, \mathcal{E}_2$ and the Siam-SRs are shared to reduce over-fitting risks. The enhanced features $\hat{X}_1$ and $\hat{X}_2$ are further sent to a Cross-temporal SR (Cot-SR) block to model their semantic correlations. These correlations between the two temporal branches are also supervised by the semantic consistency loss (introduced in Sec.\ref{sc2.loss}). The temporally aligned features (denoted as $\tilde{X}_1$ and $\tilde{X}_2$) are then projected into two semantic maps $P_1$ and $P_2$. Meanwhile, the CD block models change information through the unaligned features $\hat{X}_1$ and $\hat{X}_2$, before them being projected into a binary change map $C$. All the projections are made through \textit{$1\times1$} convolutional layers whose weights are not shared. Same as the SSCD-l, the Bi-SRNet produces as outputs 3 direct maps: the semantic maps $P_{1}, P_{2}$ and a binary CD map $C$. Finally, the semantic change maps $S_1$ and $S_2$ are generated by masking $P_{1}$ and $P_{2}$ with $C$. $\mathcal{E}_1$, $\mathcal{E}_2$ and $\mathcal{D}$ in the Bi-SRNet are identical to those in the SSCD-l. The simplified calculations (omitting the convolutional classifiers) are:

\begin{gather}
    \hat{X}_1 = {\rm SiamSR}[\mathcal{E}_1(I_1)], \hat{X}_2 = {\rm SiamSR}[\mathcal{E}_2(I_2)]\\
    P_1, P_2 = {\rm CotSR}(\hat{X}_1, \hat{X}_2)\\
    C = \mathcal{D}(\hat{X}_1, \hat{X}_2)\
    S_1, S_2 = C \cdot (P_1, P_2).
\end{gather}

Since the focus of this work is to investigate architecture designs and to model the semantic correlations for SCD, no decoder structures are adopted. The SR blocks operate on a spatial scale of $1/8$ of the input resolution. This spatial resolution is suggested for enhancing the semantic features in the literature \cite{chen2018deeplabv3+}\cite{ding2020lanet}, as it achieves a balance between spatial accuracy and context modelling distance. The network outputs are directly enlarged as the results.

\begin{figure}[t]
\centering
        \subcaptionbox{}
        {\includegraphics[height=4.6cm]{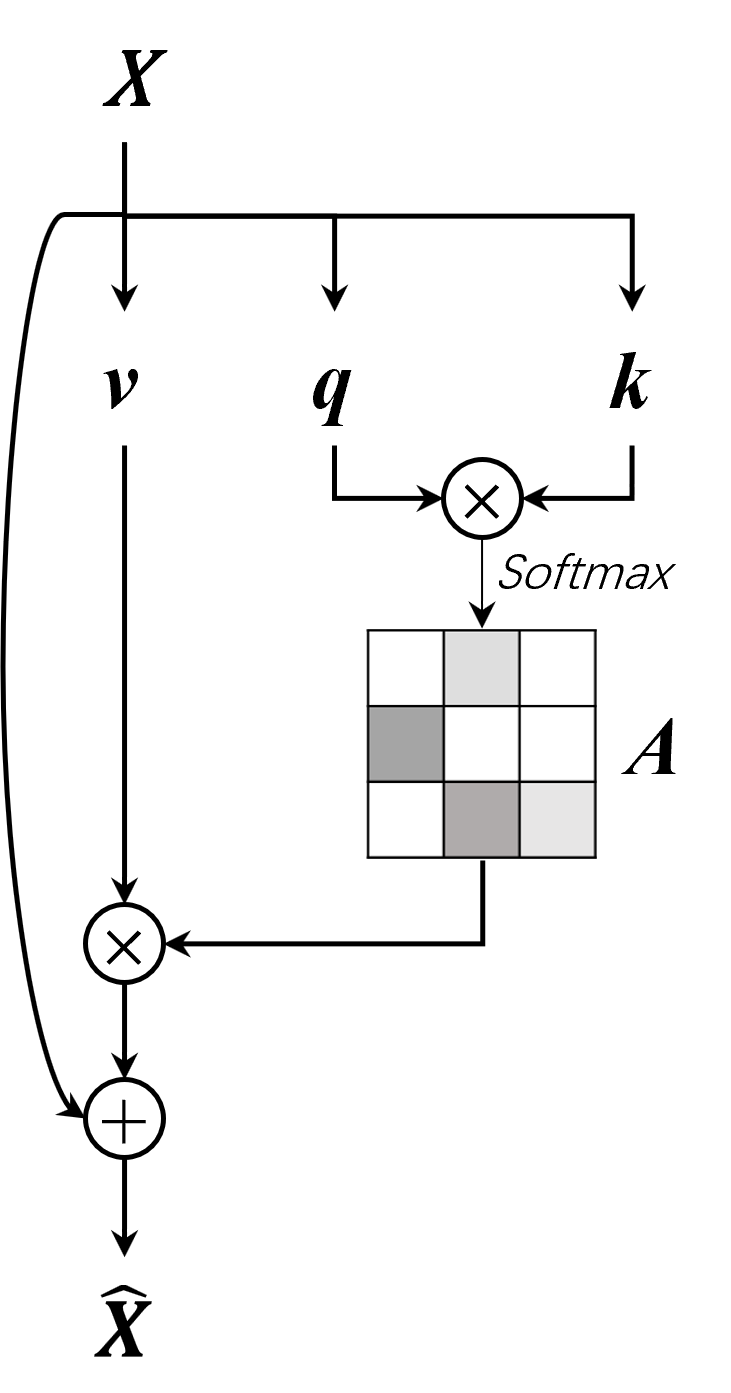}}
        \subcaptionbox{}
        {\includegraphics[height=4.6cm]{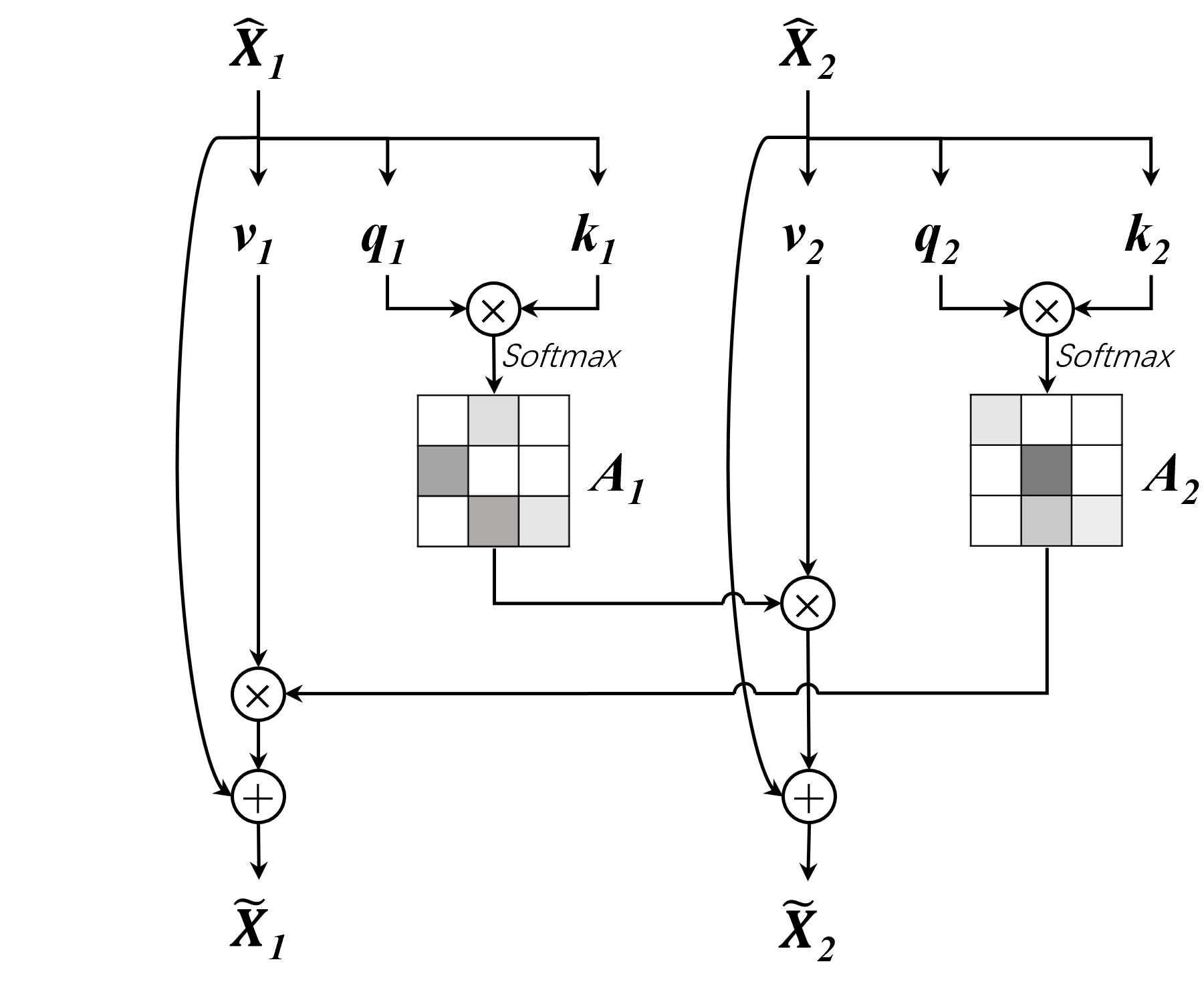}}
    \caption{Structures of the (a) Siamese Semantic Reasoning (Siam-SR) and (b) Cross-temporal Semantic Reasoning (Cot-SR) blocks.}\label{Fig.SR_block}
\end{figure}

\subsection{Semantic Reasoning Blocks}\label{sc2.SR}

Non-local units \cite{wang2018nonlocal} have been proved effective to model long-range dependencies in images and have been widely used in semantic segmentation tasks~\cite{nam2017dual, mou2019relation, zhang2020multi}. In the SCD task it is beneficial to take into account both i) the spatial correlations within each temporal image and ii) the semantic correlation and consistency between the bi-temporal images. In the Bi-SRNet these information are learned through the Siam-SR blocks and the Cot-SR block, respectively.

The Siam-SR blocks are two standard non-local units that share the same weights. Fig.\ref{Fig.SR_block}(a) illustrates the operations inside a Siam-SR block. Given an input feature $\textbf{X} \in \mathbb{R}^{c \times h \times w}$ where $c$ is the number of channels and $h, w$ are the spatial sizes, a Siam-SR block first projects it into three vectors $\textbf{q} \in \mathbb{R}^{H \times c'}, \textbf{k} \in \mathbb{R}^{c' \times H}$ and $\textbf{v} \in \mathbb{R}^{c \times H}$, where $H = hw$ and $c' = c/r$, $r$ is a channel reduction factor (normally set to 2). An attention matrix $\textbf{A} \in \mathbb{R}^{H \times H}$ is then calculated as:
\begin{equation}
    \textbf{A} = \phi (\textbf{q} \times \textbf{k})
\end{equation}
where $\phi$ is a \textit{softmax} normalization function along the row dimension. $\textbf{A}$ records the correlations between each pair of spatial positions. An enhanced feature $\hat{\textbf{X}}$ is then obtained with:
\begin{equation}
    \hat{\textbf{X}} = \textbf{X} + \textbf{v} \times \textbf{A}
\end{equation}

Intuitively, the bi-temporal information provides more clues of the image context. Therefore, we propose the Cot-SR block, which is an extension of the non-local unit to model cross-temporal information. As illustrated in Fig.\ref{Fig.SR_block}(b), the Cot-SR block simultaneously enhances two temporal features $\hat{\textbf{X}}_1, \hat{\textbf{X}}_2 \in \mathbb{R}^{c \times h \times w}$. First, 6 vectors $\textbf{q}_1, \textbf{q}_2 \in \mathbb{R}^{H \times c'}$, $\textbf{k}_1, \textbf{k}_2 \in \mathbb{R}^{c' \times H}$ and $\textbf{v}_1, \textbf{v}_2 \in \mathbb{R}^{c \times H}$ are projected from $\hat{\textbf{X}}_1$ and $\hat{\textbf{X}}_2$. Second, two attention matrices $\textbf{A}_1, \textbf{A}_2 \in \mathbb{R}^{H \times H}$ are generated to record the semantic focuses in each temporal branch:
\begin{gather}
    \textbf{A}_1 = \phi (\textbf{q}_1 \times \textbf{k}_1)\\
    \textbf{A}_2 = \phi (\textbf{q}_2 \times \textbf{k}_2)
\end{gather}

Finally, each attention map operates on its opposite temporal branch to project cross-temporal correlations:
\begin{gather}
    \tilde{\textbf{X}}_1 = \hat{\textbf{X}}_1 + \textbf{v}_1 \times \textbf{A}_2\\
    \tilde{\textbf{X}}_2 = \hat{\textbf{X}}_2 + \textbf{v}_2 \times \textbf{A}_1
\end{gather}
where $\tilde{\textbf{X}}_1$, $\tilde{\textbf{X}}_2$ are the enhanced features.

The Siam-SR blocks aggregate spatial information to embed semantic focuses into each temporal branch, whereas the Cot-SR learns cross-temporal semantic consistency to enhance the features in unchanged areas.

\subsection{Loss Functions}\label{sc2.loss}

\begin{figure}[t]
    \centering
    \includegraphics[width=1\linewidth]{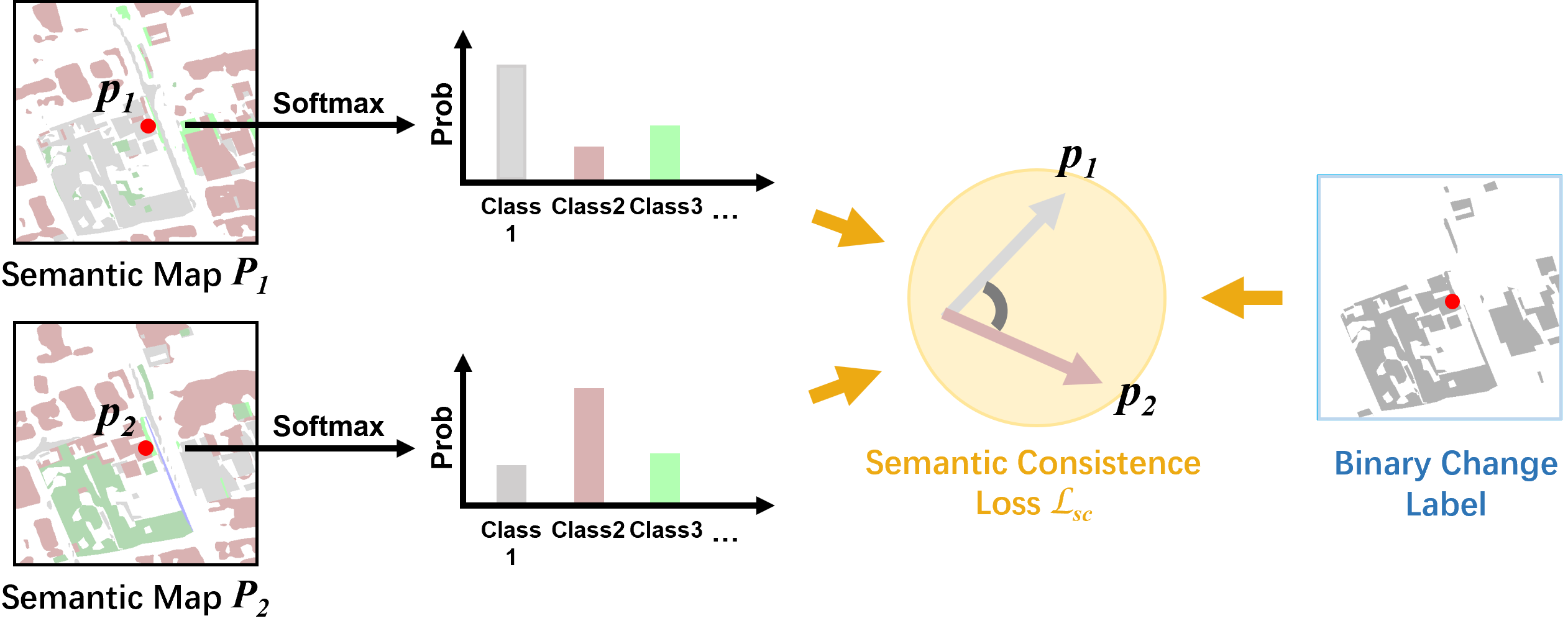}
    \caption{Illustration of the calculation of the Semantic consistency Loss (SCLoss).}
    \label{Fig.SCLoss}
\end{figure}

We use three loss functions to train the Bi-SRNet: the semantic class loss $\mathcal{L}_{sem}$, the binary change loss $\mathcal{L}_{change}$, and a proposed semantic consistency loss $\mathcal{L}_{sc}$.

The semantic loss $\mathcal{L}_{sem}$ is the mutli-class cross entropy loss between the semantic segmentation results $P_1, P_2$ and the GT semantic change maps $L_1, L_2$. The calculation of $\mathcal{L}_{sem}$ on each pixel is:

\begin{equation}\label{Fml.LossSeg}
    \mathcal{L}_{sem} = -\frac{1}{N}\sum_{i=1}^{N} y_{i}log(p_{i})
\end{equation}
where $N$ is the number of semantic classes, $y_{i}$ and $p_{i}$ denote the GT label and the predicted probability of the $i$-th class, respectively. $N$ is set according to the number of LCLU classes in the dataset. The \textit{no-change} class (indexed as '0') is excluded from loss calculation to encourage the temporal branches to focus on extracting the semantic features.

\begin{figure}[t]
    \centering
    \includegraphics[width=7cm]{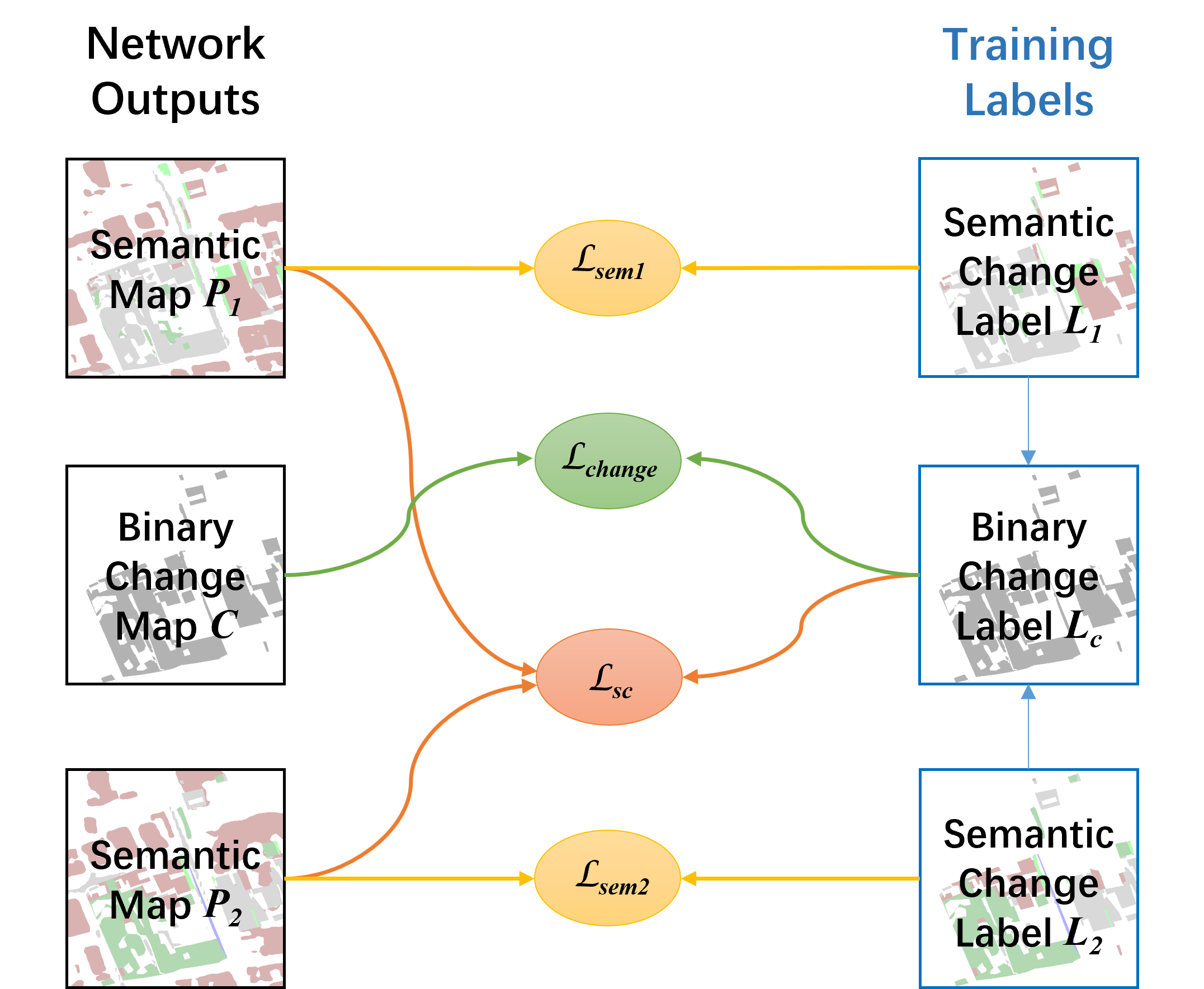}
    \caption{The calculation of supervision losses in the Bi-SRNet.}
    \label{Fig.Loss}
\end{figure}

The change loss $\mathcal{L}_{change}$ is the binary cross entropy loss between the predicted binary change map $C$ and the reference change map $L_c$. The $L_c$ is generated with $L_1$ or $L_2$ by replacing all their non-zero labels with a \textit{changed} label (indexed as '1'). The $\mathcal{L}_{change}$ for each pixel is calculated as:
\begin{equation}
    \mathcal{L}_{change} = -y_{c}log(p_{c})-(1-y_{c})log(1-p_{c})
\end{equation}
where $y_{c}$ and $p_{c}$ denote the GT label and the predicted probability of change, respectively.

$\mathcal{L}_{sem}$ and $\mathcal{L}_{change}$ are designed to drive the learning of the semantic information and of the CD, respectively. We further propose a task-specific semantic consistency loss (SCLoss) to link SS with CD. It aligns the semantic representations between the two temporal branches, as well as guides training of the Cot-SR block. As illustrated in Fig.\ref{Fig.SCLoss}, the SCLoss awards predictions with similar probability distributions in the \textit{no-change} areas, whereas punishing those in the changed areas. This aligns the bi-temporal semantic and change information in the SCD task. The SCLoss $\mathcal{L}_{sc}$ is calculated between the predicted semantic maps $P_1, P_2$ and the change label $L_c$ using the Cosine loss function:
\begin{equation}
    \mathcal{L}_{sc} =\left\{\begin{array}{lr}
                                1-cos(x_1, x_2), & y_{c}=1  \\
                                cos(x_1, x_2), & y_{c}=0
                            \end{array}
                      \right.
\end{equation}
where $x_1, x_2$ are feature vectors of a pixel on $P_1$ and $P_2$, respectively. $y_{c}$ is the value at the same position on $L_c$.

Training of the two feature embedding branches is directly supervised by $L_1$ and $L_1$ and is assisted by $L_c$ (through the $\mathcal{L}_{sc}$), while training of the CD block is directly supervised by $L_c$. The relationships between the 3 outputs $P_1, P_2, C$ and the GT maps $L_1$, $L_2$ and $L_c$ are illustrated in Fig.\ref{Fig.Loss}. The total loss $\mathcal{L}_{scd}$ is calculated as:

\begin{equation}
    \mathcal{L}_{scd} = (\mathcal{L}_{sem_1} + \mathcal{L}_{sem_2})/2 + \mathcal{L}_{change} + \mathcal{L}_{sc}
\end{equation}
where $\mathcal{L}_{sem_1}$ and $\mathcal{L}_{sem_2}$ are the semantic loss of each temporal branch. They are added and averaged to represent the $\mathcal{L}_{sem}$. Since the learning of LCLU information and change information is disentangled in the SSCD-l architecture, no hyper-parameters are used to balance the loss functions.

\section{Dataset Description and Experimental Settings}\label{sc4}
In this section we describe the dataset, the evaluation metrics and the experimental settings.

\subsection{Dataset}
The experiments in this study are conducted on the SEmantic Change detectiON Dataset (SECOND)\cite{yang2020asymmetric}, a benchmark dataset for the SCD. The SECOND is constructed with bi-temporal HR optical images (containing RGB channels) collected by several aerial platforms and sensors. The observed regions include several cities in China, including Hangzhou, Chengdu and Shanghai. Each image has the same size of $512 \times 512$ pixels. The spatial resolution varies from 0.5m to 3m (per pixel).

The LC categories before and after the change events are provided. In each GT semantic change map, one change class and six LC classes are annotated, including: \textit{no-change, non-vegetated ground surface, tree, low vegetation, water, buildings} and \textit{playgrounds}. These LC classes are selected considering the commonly interesting LC classes and the frequent geographical changes \cite{tong2020land}. The bi-temporal LC transitions raise a total of 30 LC change types. The changed pixels account for 19.87\% of the total image pixels. Among the 4662 pairs of temporal images, 2968 ones are openly available. We further split them into a training set and a test set with the numeric ratio of $4:1$ (i.e., 2375 image pairs for training, 593 ones for testing).

\subsection{Evaluation Metrics}

In this study, 3 evaluation metrics are adopted to evaluate the SCD accuracy, including: overall accuracy (OA), mean Intersection over Union (mIoU) and Separated Kappa (SeK) coefficient. OA has been commonly adopted in both semantic segmentation \cite{ding2020lanet, ding2020diresnet} and CD \cite{daudt2019multitask} tasks. Let us denote $Q=\{q_{i,j}\}$ as the confusion matrix where $q_{i,j}$ represents the number of pixels that are classified into class $i$ while their GT index is $j$ ($i, j \in \{0, 1, ..., N\}$ ($0$ represents \textit{no-change}). OA is calculated as:
\begin{equation}
    OA = \sum_{i=0}^{N} q_{ii} / \sum_{i=0}^{N}\sum_{j=0}^{N} q_{ij}.
\end{equation}
Since OA is mostly determined by the identification of \textit{no-change} pixels, it cannot well evaluate the segmentation of LCLU classes. Additionally, it does not count the pixels that are identified as \textit{changed} but are predicted into wrong LCLU classes. Alternatively, mIoU and SeK are suggested in the SECOND \cite{yang2020asymmetric} to evaluate the discrimination of changed/\textit{no-change} regions and the segmentation of LC classes, respectively.

mIoU is the mean value of the IoU of \textit{no-change} regions ($IoU_{nc}$) and that of the changed regions ($IoU_{c}$):
\begin{gather}
    mIoU = (IoU_{nc} + IoU_{c})/2,\\
    IoU_{nc} = q_{00}/ (\sum_{i=0}^{N} q_{i0} + \sum_{j=0}^{N} q_{0j} - q_{00}),\\
    IoU_{c} = \sum_{i=1}^{N} \sum_{j=1}^{N} q_{ij} / (\sum_{i=0}^{N} \sum_{j=0}^{N} q_{ij} - q_{00}),
\end{gather}

The SeK coefficient is calculated based on the confusion matrix $\hat{Q} = \{\hat{q}_{ij}\}$ where $\hat{q}_{ij} = q_{ij}$ except that $\hat{q}_{00} = 0$. This is to exclude the true positive \textit{no-change} pixels, whose number is dominant. The calculations are as follows:
\begin{gather}
    \rho = \sum_{i=0}^{N}\hat{q}_{ii} / \sum_{i=0}^{N}\sum_{j=0}^{N}\hat{q}_{ij},\\
    \eta = \sum_{i=0}^{N} (\sum_{j=0}^{N}\hat{q}_{ij} * \sum_{j=0}^{N}\hat{q}_{ji}) / (\sum_{i=0}^{N}\sum_{j=0}^{N}\hat{q}_{ij}) ^2,\\
    SeK = e^{IoU_c-1} \cdot (\rho - \eta) / (1 - \eta).
\end{gather}

The mIoU and SeK directly evaluate the sub-tasks in SCD, i.e., the CD and the SS of LCLU classes, respectively. Additionally, to evaluate more intuitively the segmentation of LCLU classes in changed areas, we introduce a new metric $F_{scd}$ (derived from the $F_1$ score in segmentation and CD tasks \cite{daudt2018fully}\cite{lei2021aslnet}):

\begin{gather}
    P_{scd} = \sum_{i=1}^{N} q_{ii} / \sum_{i=1}^{N}\sum_{j=0}^{N} q_{ij},\\
    R_{scd} = \sum_{i=1}^{N} q_{ii} / \sum_{i=0}^{N}\sum_{j=1}^{N} q_{ij},\\
    F_{scd} = \frac{2*P_{scd}*R_{scd}}{P_{scd}+R_{scd}}
\end{gather}
where $P_{scd}$ and $R_{scd}$ are variants of the $Precision$ and $Recall$ \cite{ding2020lanet} which focus only on the changed areas. $F_{scd}$ describes the segmentation accuracy of LCLU classes in the changed areas.

Finally, three metrics are provided to measure the computational costs, including the size of parameters (Params), the floating point operations (FLOPs) and the inference (Infer) time for 100 epochs. The FLOPs and Infer time are measured considering the calculations for a pair of input images each with $512 \times 512$ pixels.

\subsection{Experimental settings}

The experiments in this study are conducted on a workstation with a NVIDIA Quadro P6000 GPU. All the CNN models are implemented with the PyTorch library. The same experimental parameters are used in all the experiments, including batch size (8), running epochs (50) and initial learning (0.1). The gradient descent optimization method is Stochastic Gradient Descent (SGD) with Nesterov momentum. The augmentation strategy include random flipping and rotating while loading the image pairs. For simplicity, no test-time augmentation operation is applied. 
\section{Experimental Results}\label{sc5}
In this section, a series of experiments are conducted to verify the effectiveness of the proposed architecture for SCD (the SSCD-l) and the components in the Bi-SRNet. Finally the proposed methods are compared with several recent methods in both SCD and binary CD.

\subsection{Comparison of SCD Architectures}\label{sc5.comareArch}

\begin{table*}[t]
    \centering
    \caption{Comparison of the results provided by different CNN architectures for SCD.}
    \resizebox{1\linewidth}{!}{%
        \begin{tabular}{r|ccc|cccc}
        \toprule
            \multirow{2}*{Method} & \multicolumn{3}{c|}{Computational Costs} & \multicolumn{4}{c}{Accuracy}\\
            \cline{2-8}
            & Params (Mb) & FLOPS (Gbps) & Infer. Time (s/100e) & OA(\%) & mIoU(\%) & Sek(\%) & $F_{scd}$(\%) \\
            \hline
            DSCD-e & 21.36 & 91.39 & 1.32 & 86.46 & 68.55 & 16.01 & 56.22\\
            DSCD-l & 23.31 & 189.54 & 2.74 & 86.58 & 68.86 & 16.43 & 56.67 \\
            SSCD-e & 42.72 & 272.95 & 3.89 & 85.66 & 69.55 & 17.05 & 56.96 \\
            proposed SSCD-l & 23.31 & 189.57 & 2.75 & \textbf{87.19} & \textbf{72.60} & \textbf{21.86} & \textbf{61.22} \\
        \bottomrule
        \end{tabular} }\label{Table.CompareArchs}
\end{table*}

\begin{figure*}
\centering
    {\includegraphics[width=14cm]{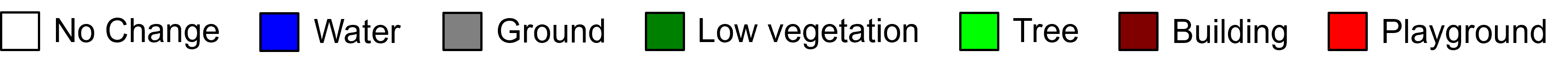}}\\
    \setlength{\tabcolsep}{1pt}
    \begin{tabular}{>{\centering\arraybackslash}m{0.6cm}>{\centering\arraybackslash}m{2.0cm}>{\centering\arraybackslash}m{2.0cm}>{\centering\arraybackslash}m{2.0cm}>{\centering\arraybackslash}m{2.0cm}>{\centering\arraybackslash}m{2.0cm}>{\centering\arraybackslash}m{2.0cm}>{\centering\arraybackslash}m{2.0cm}>{\centering\arraybackslash}m{2.0cm}}
        (a1) &
        \includegraphics[width=2.0cm]{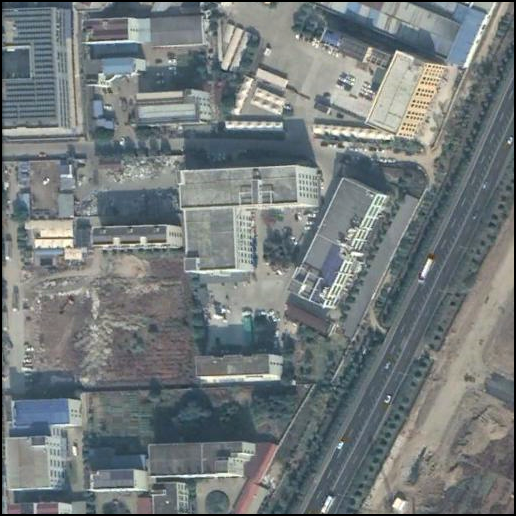} &
        \includegraphics[width=2.0cm]{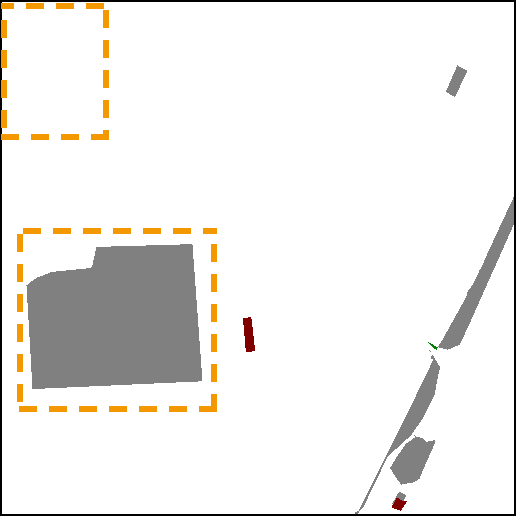} &
        \includegraphics[width=2.0cm]{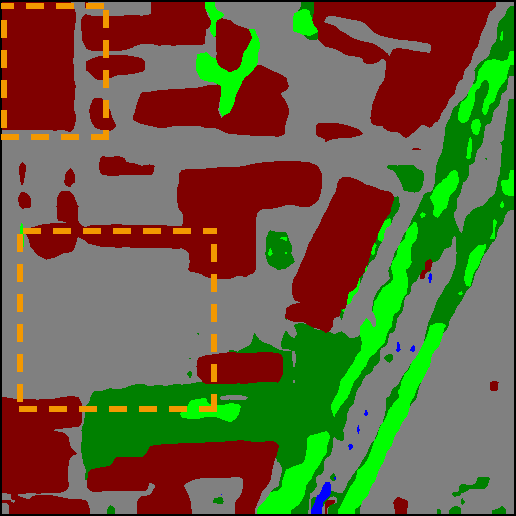} &
        \multirow{2}*{\includegraphics[width=2.0cm]{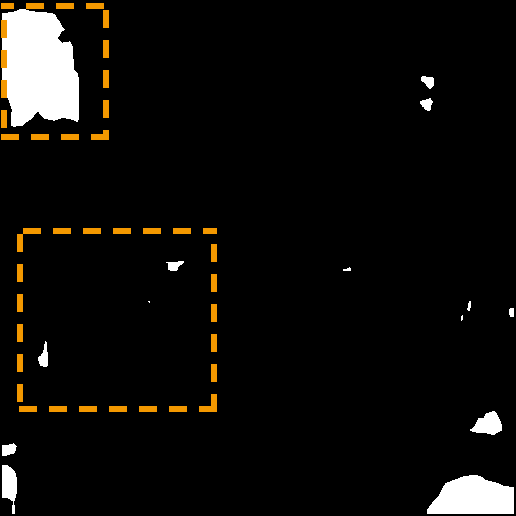}} &
        \includegraphics[width=2.0cm]{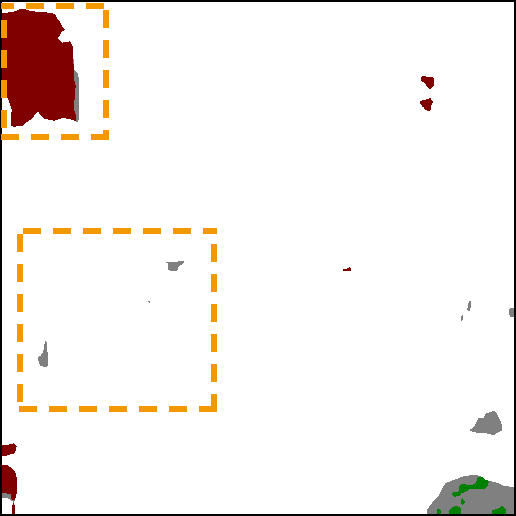} &
        \includegraphics[width=2.0cm]{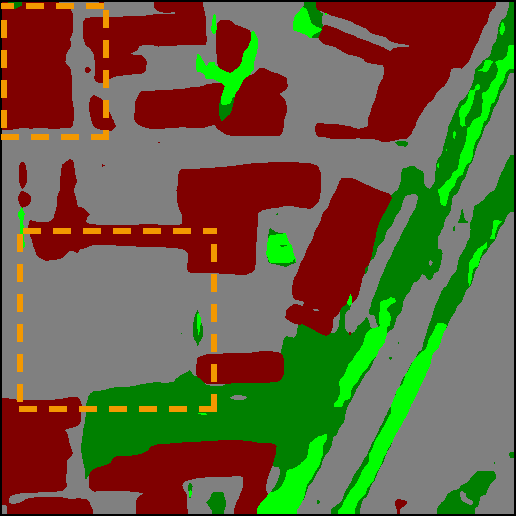} &
        \multirow{2}*{\includegraphics[width=2.0cm]{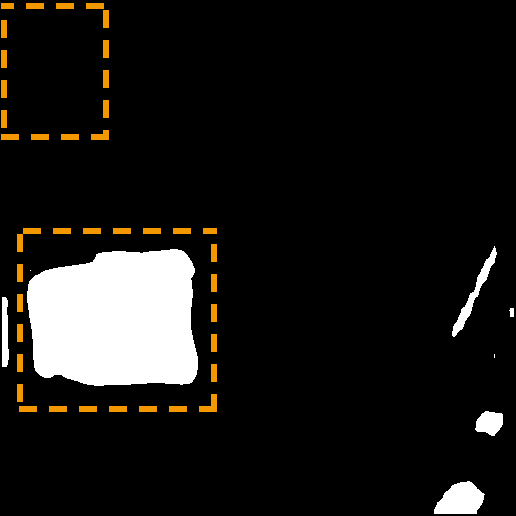}} &
        \includegraphics[width=2.0cm]{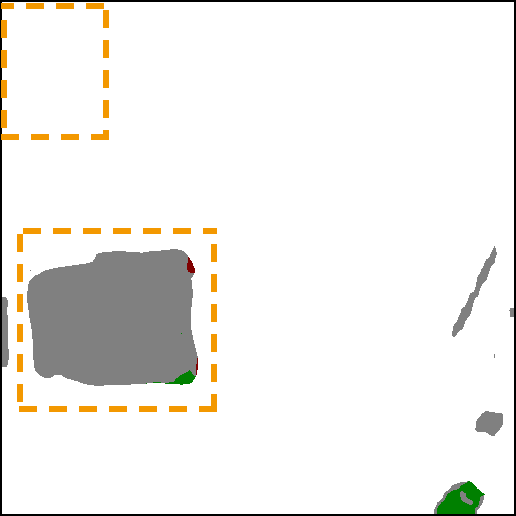}\\
        (a2) &
        \includegraphics[width=2.0cm]{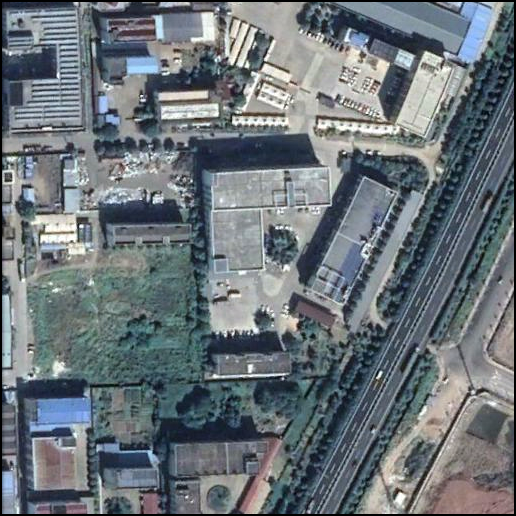} &
        \includegraphics[width=2.0cm]{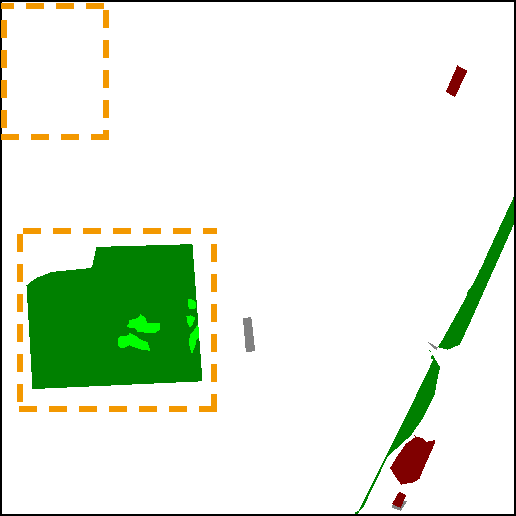} &
        \includegraphics[width=2.0cm]{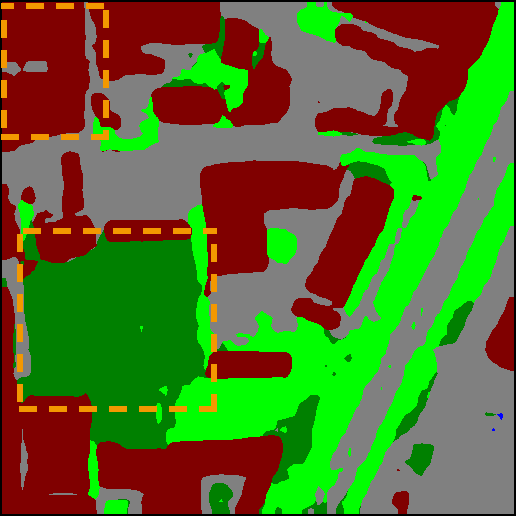} & &
        \includegraphics[width=2.0cm]{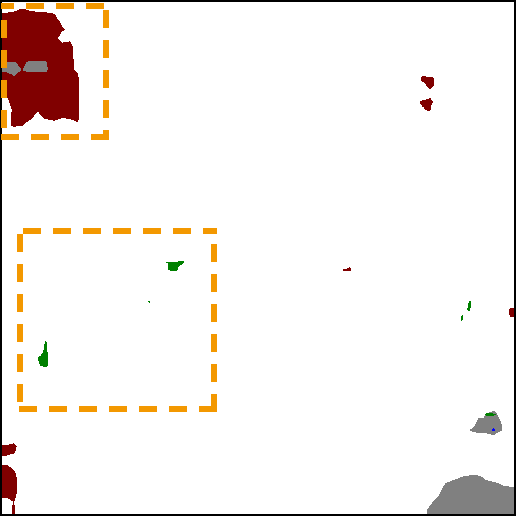} &
        \includegraphics[width=2.0cm]{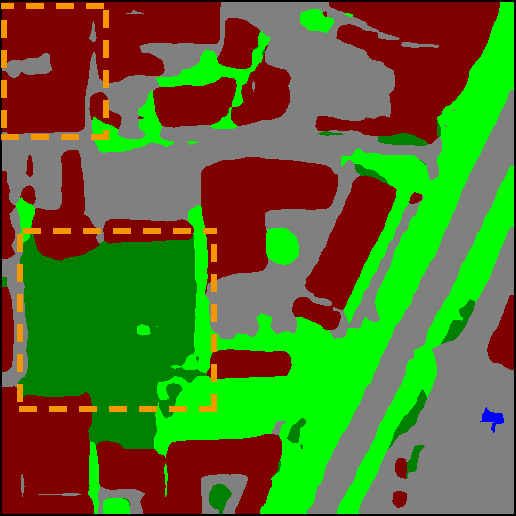} & &
        \includegraphics[width=2.0cm]{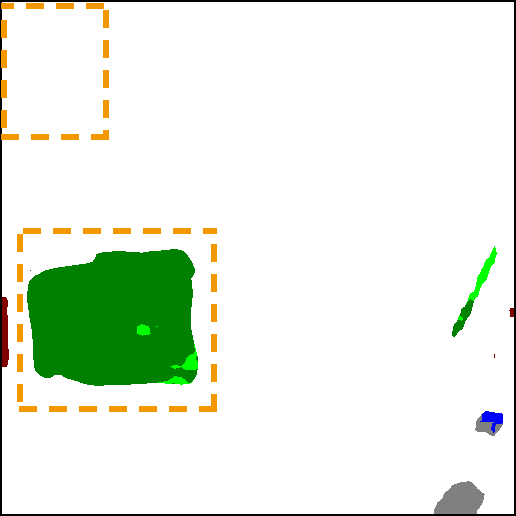}\\
        \hline\\
        (b1) &
        \includegraphics[width=2.0cm]{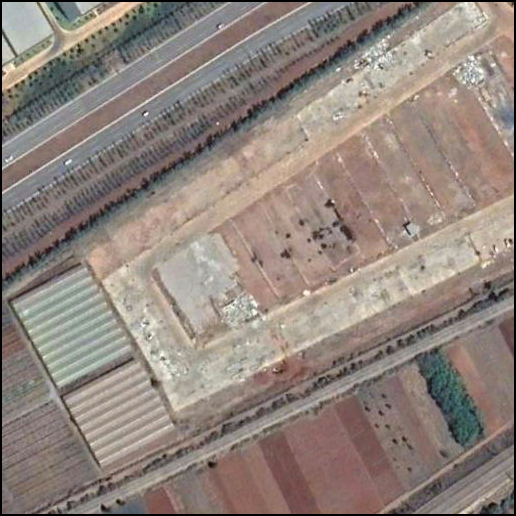} &
        \includegraphics[width=2.0cm]{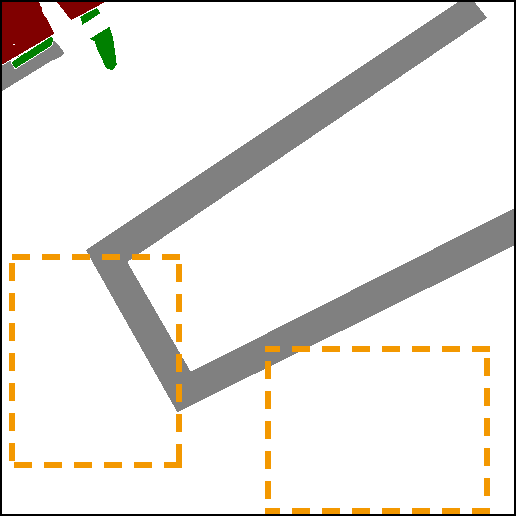} &
        \includegraphics[width=2.0cm]{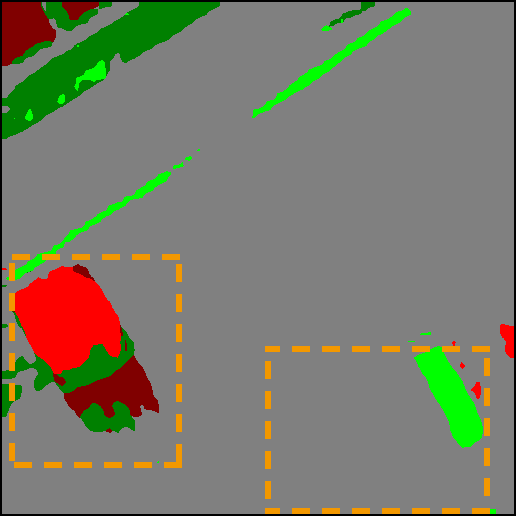} &
        \multirow{2}*{\includegraphics[width=2.0cm]{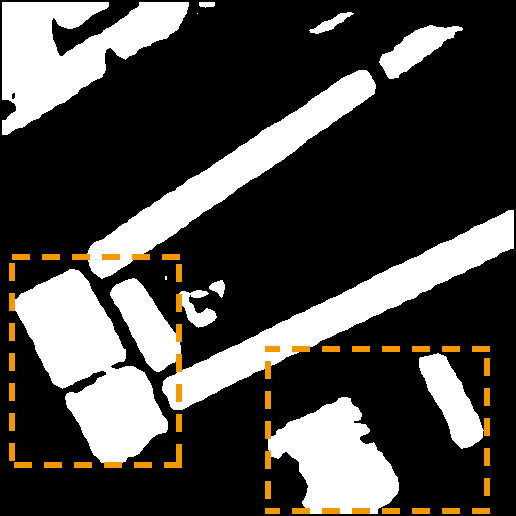}} &
        \includegraphics[width=2.0cm]{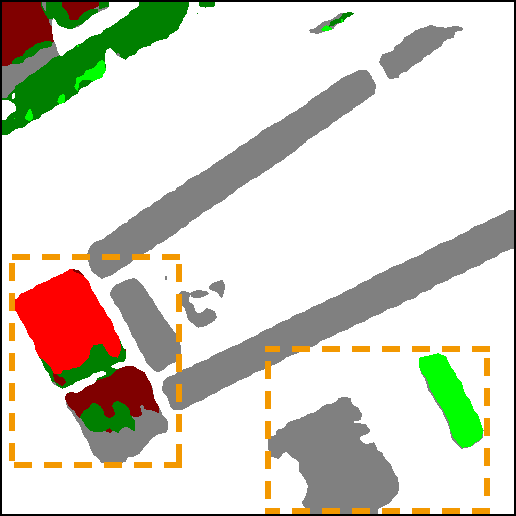} &
        \includegraphics[width=2.0cm]{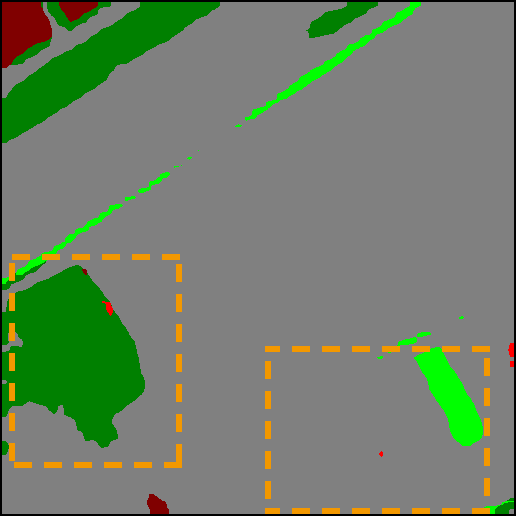} &
        \multirow{2}*{\includegraphics[width=2.0cm]{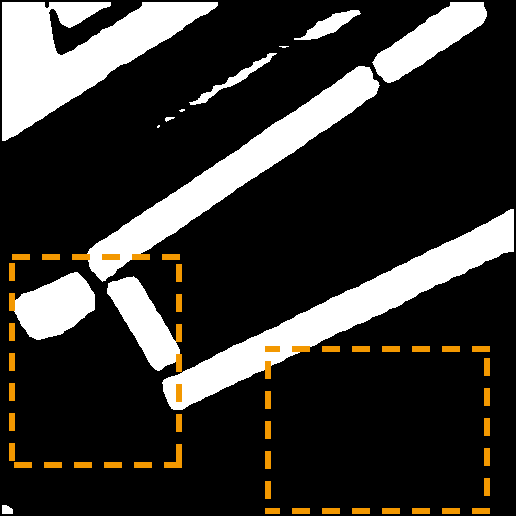}} &
        \includegraphics[width=2.0cm]{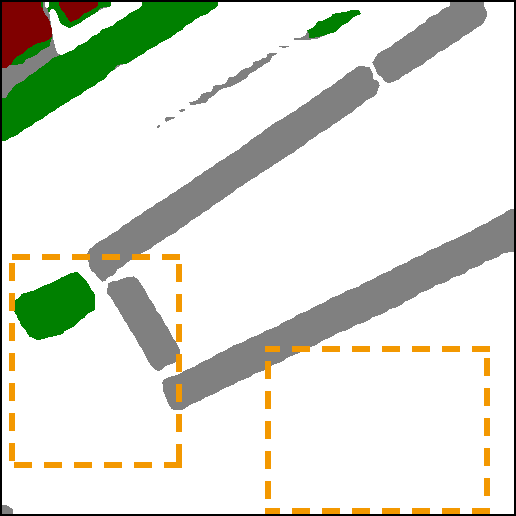}\\
        (b2) &
        \includegraphics[width=2.0cm]{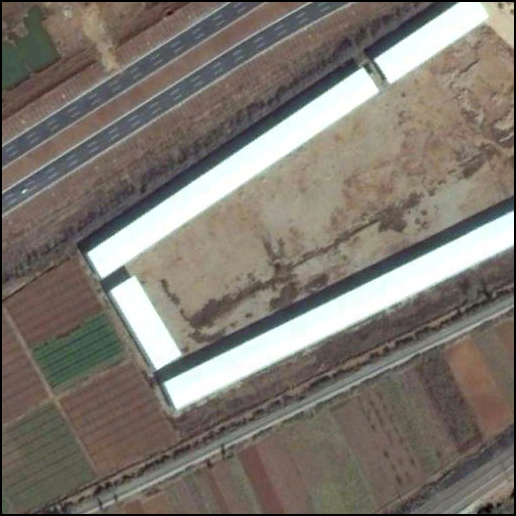} &
        \includegraphics[width=2.0cm]{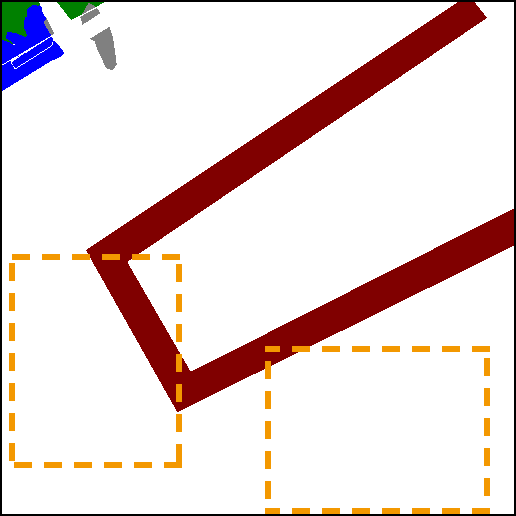} &
        \includegraphics[width=2.0cm]{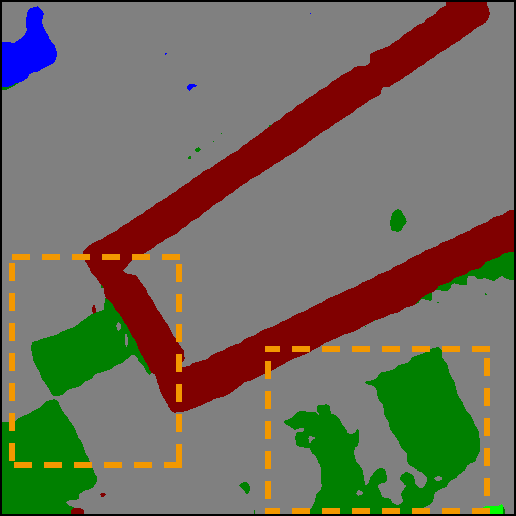} & &
        \includegraphics[width=2.0cm]{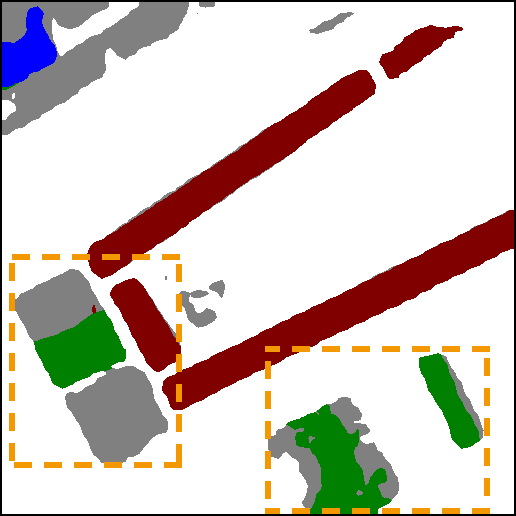} &
        \includegraphics[width=2.0cm]{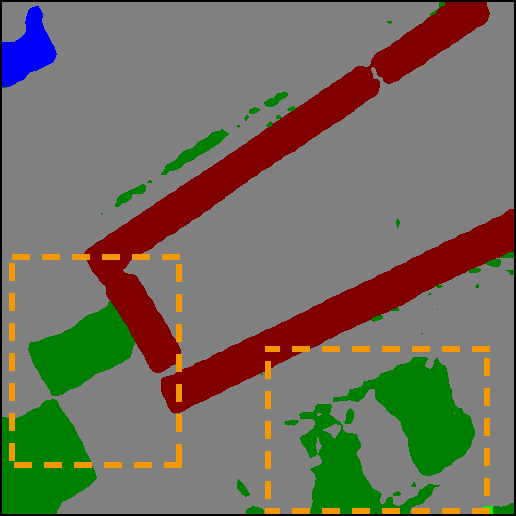} & &
        \includegraphics[width=2.0cm]{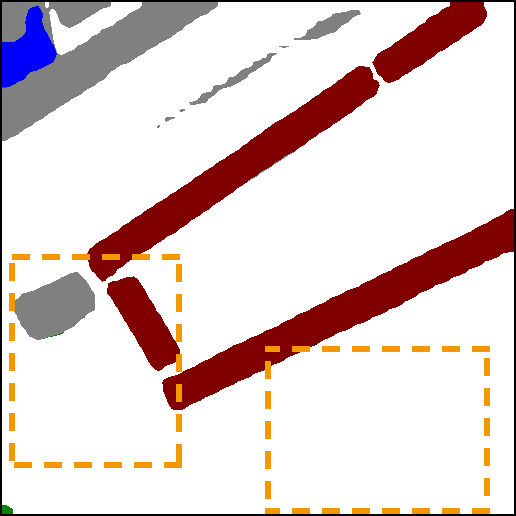}\\
        \hline\\
        (c1) &
        \includegraphics[width=2.0cm]{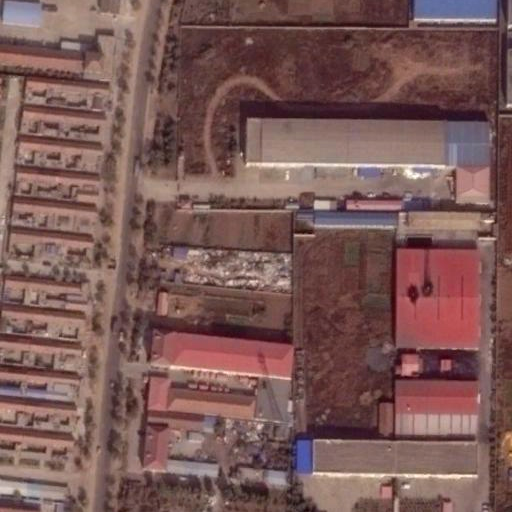} &
        \includegraphics[width=2.0cm]{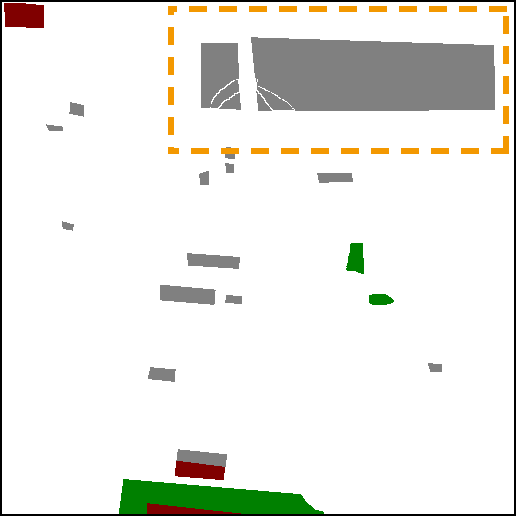} &
        \includegraphics[width=2.0cm]{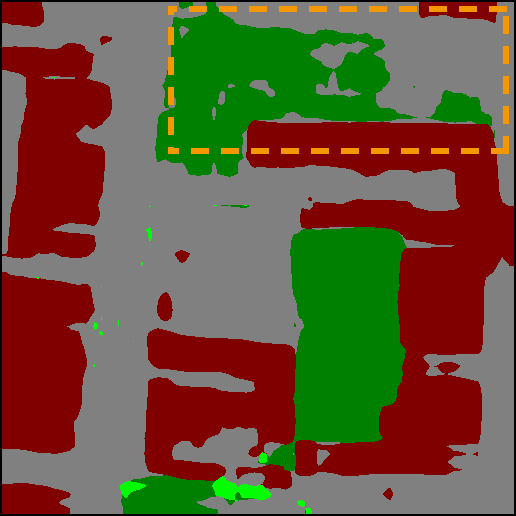} &
        \multirow{2}*{\includegraphics[width=2.0cm]{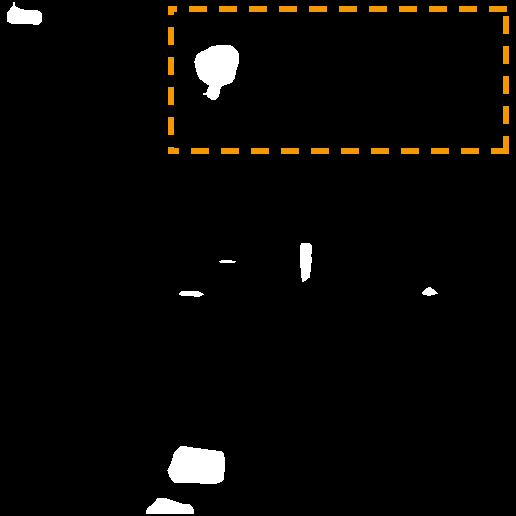}} &
        \includegraphics[width=2.0cm]{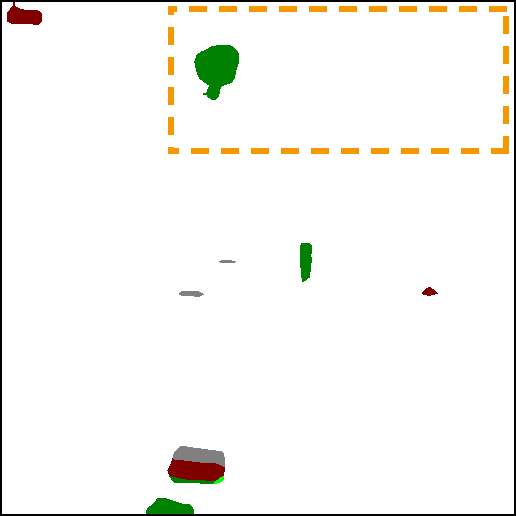} &
        \includegraphics[width=2.0cm]{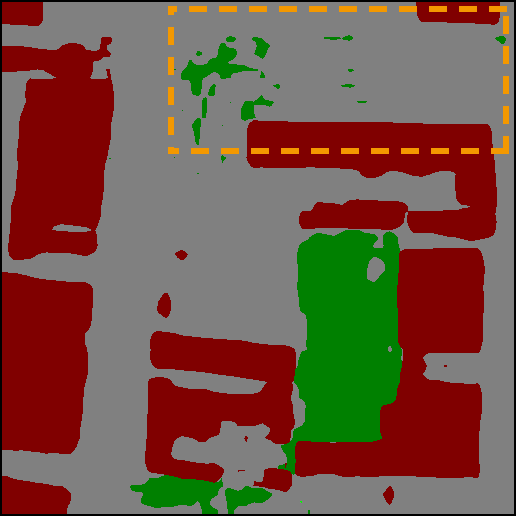} &
        \multirow{2}*{\includegraphics[width=2.0cm]{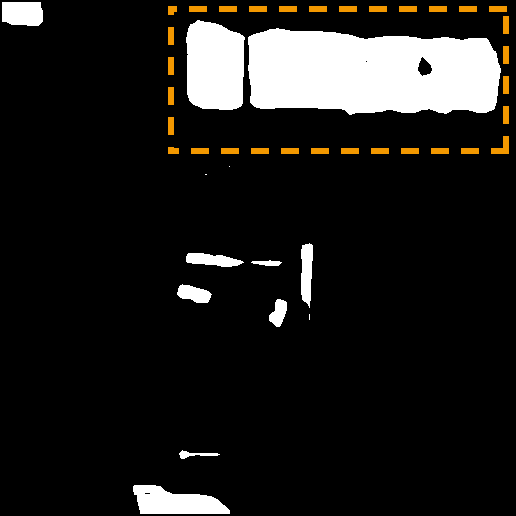}} &
        \includegraphics[width=2.0cm]{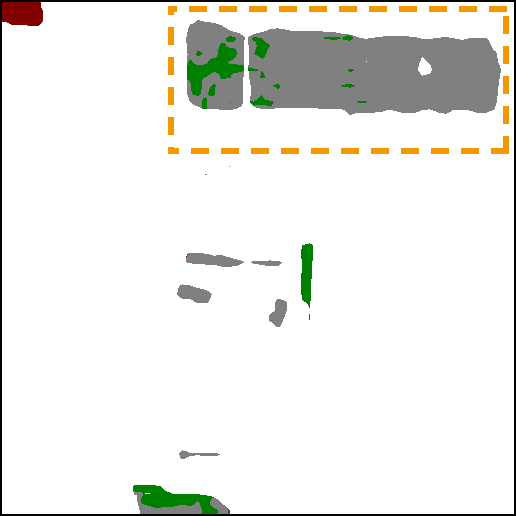}\\
        (c2) &
        \includegraphics[width=2.0cm]{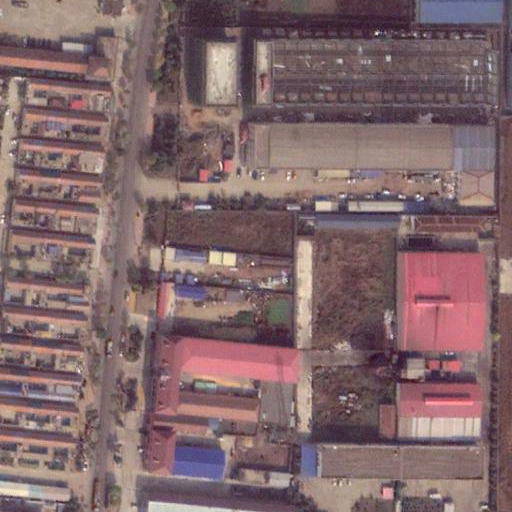} &
        \includegraphics[width=2.0cm]{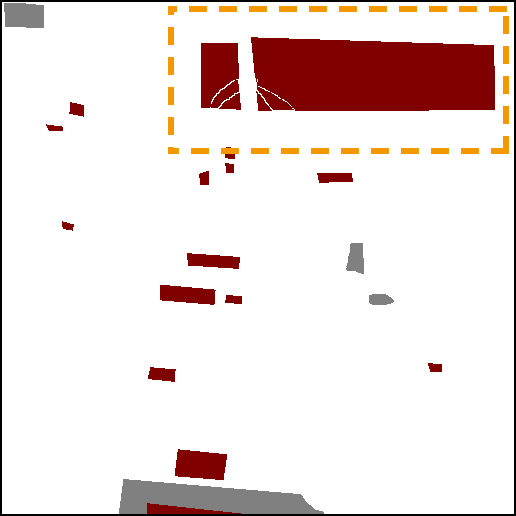} &
        \includegraphics[width=2.0cm]{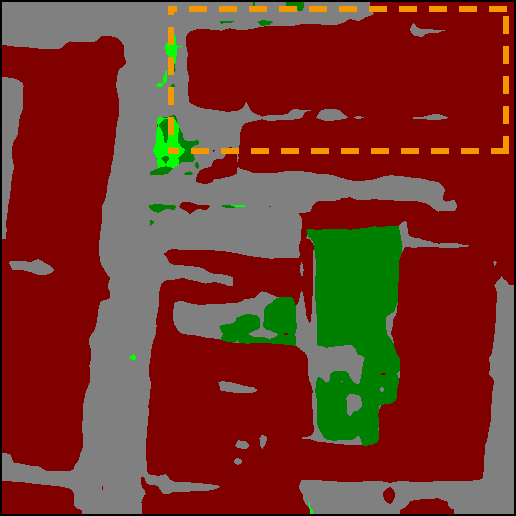} & &
        \includegraphics[width=2.0cm]{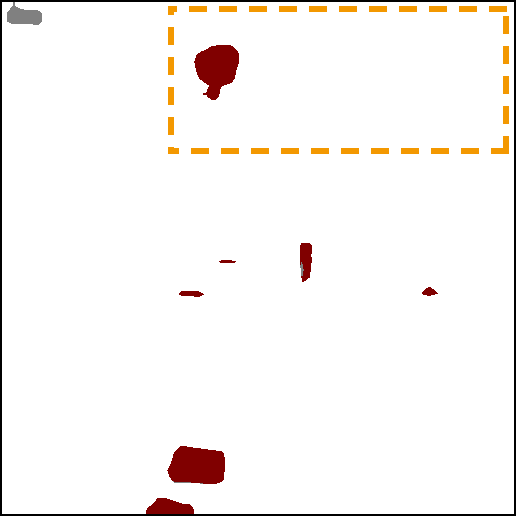} &
        \includegraphics[width=2.0cm]{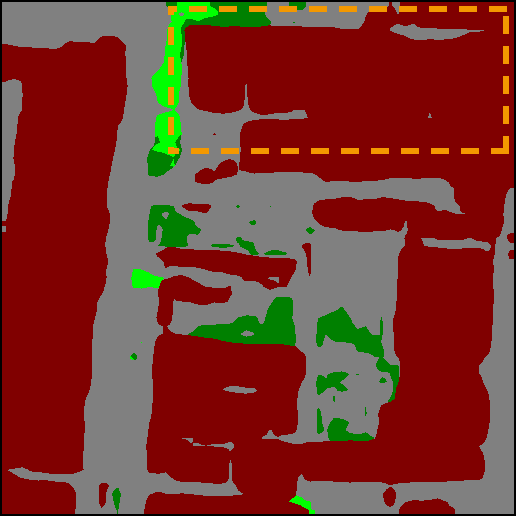} & &
        \includegraphics[width=2.0cm]{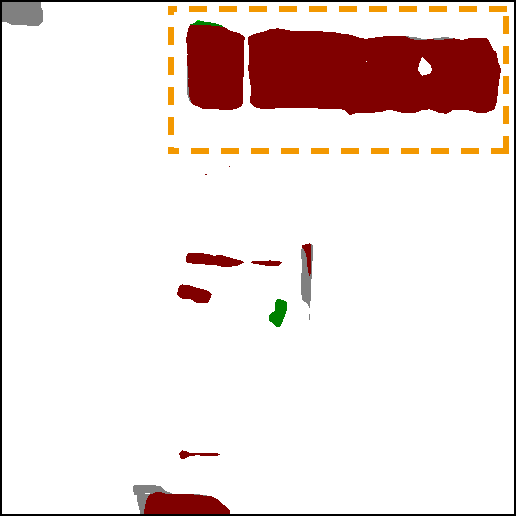}\\
        & \multirow{2}*{Test image} & \multirow{2}*{GT} & \multicolumn{3}{|c|}{$\underbrace{\rm LC\:map \;\;\;\;\; Change\:map \;\;\;\;\; SCD\:map}$} &  \multicolumn{3}{c}{$\underbrace{\rm LC\:map \;\;\;\;\; Change\:map \;\;\;\;\; SCD\:map}$}\\
        & & & \multicolumn{3}{|c|}{SSCD-e} & \multicolumn{3}{c}{SSCD-l (proposed)}\\
    \end{tabular}
    \caption{Comparisons of the results provided by the SSCD-e and SSCD-l architectures.} \label{Fig.ResultsArch}
\end{figure*}

To find the optimal CNN architecture for the SCD, we compare the 4 approaches discussed in Sec.\ref{sc3.archs}: the DSCD-e, DSCD-l, SSCD-e and SSCD-l. For simplicity and fairness, all the tested architectures are implemented with only the basic convolutional designs (i.e., sophisticated designs such as encoder-decoder structures, dilated convolutions and attention units are not adopted). They employ the same CNN encoder (the ResNet34 \cite{he2016resnet}), adopt the same down-sampling stride ($\times 1/8$) and have same number of inner channels at each convolutional stage. The late-fusion approaches (DSCD-e and SSCD-l) employ an identical CD block, i.e., a CNN block with 6 stacked residual units.

The quantitative results are reported in Table.\ref{Table.CompareArchs}. The DSCD-e architecture obtains the lowest accuracy, although it requires the lowest computations. Meanwhile, DSCD-l achieves much higher metric values, showing that the separate embedding of temporal features and CD features is essential in SCD. Compared to the DSCD-l, the SSCD-e slightly increases the mIoU values at the cost of taking much heavier computations. This shows that the embedding of CD features from scratch is in-efficient. On the contrary, the proposed SSCD-l re-uses and merges the temporal features for CD, which leads to significant performance improvements. It surpasses the SSCD-e by 4.81\% in Sek and 4.26\% in $F_{scd}$, whereas its computational cost is only slightly higher than that of the DSCD-l.

Fig.\ref{Fig.ResultsArch} qualitatively compares the results obtained by the SSCD-e and SSCD-l architectures. The SCD maps are generated by masking the LC maps with the change maps, as illustrated in Sec.\ref{sc3.archs}. The intermediate results reveal one of the major limitations of the SSCD-e architecture. Since the change information is separately embedded in the SSCD-e architecture, the segmented changes are often inconsistent with those in the bi-temporal LC maps. For example, in Fig.\ref{Fig.ResultsArch}(a) the LC change of 'ground' turning into 'low vegetation' is indicated in the LC maps but it is not represented in the change map. Also a LC change from 'building' to 'building' is indicated in the SCD map, which is self—contradictory. In Fig.\ref{Fig.ResultsArch}(c) the emergence of a building is also omitted. However, these LC changes are easily captured by the SSCD-l architecture. Through direct modelling of the LC features, the CD unit is aware of the semantic changes. Additionally, some of the non-salient differences are adaptively omitted in the results of SSCD-l, such as the transitions between between 'ground' and 'vegetation' in Fig.\ref{Fig.ResultsArch}(b).

These experimental comparisons indicate that the proposed SSCD-l provides the most accurate results in SCD among the compared architectures. Its advantages in embedding the semantic information (indicated by the SeK) is particularly dominant. However, there are also many variants of these architectures exploited in previous studies. In Sec.\ref{sc5.compareSOTA} these methods are further compared to summarize common features of the different SCD architectures.

\subsection{Ablation Study}

After verifying the effectiveness of the SSCD-l, we further perform an ablation study to evaluate the auxiliary components in the proposed Bi-SRNet. The quantitative results are presented in Table \ref{Table.Ablation}. First we test the effectiveness of the SCLoss by adding it as an auxiliary loss to train the SSCD-l. This boosts the accuracy by around 0.82\% in SeK and 0.76\% in $F_{scd}$, indicating that the semantic embedding of features is improved. Taking this method (SSCD-l with SCLoss) as the baseline, we further assess the performance of each SR blcok. The Siam-SR blocks, which are placed on each temporal branch, lead to noticeable accuracy improvements (0.4\% in mIoU and 0.43 in $F_{scd}$). Meanwhile, the Cot-SR block that models the temporal coherence improves the SeK (by over 0.41\%) and the $F_{scd}$ (by over 0.46\%). This indicates that both the Siam-SR and the Cot-SR improve the semantic embedding of temporal features, while the former also improves the detection of change information. Finally, we evaluate the Bi-SRNet which contains all these auxiliary designs. Compared to the standard SSCD-l, its improvements are around 0.81\% in mIoU, 1.36\% in SeK and 1.39 \% in $F_{scd}$. These results indicate that the Bi-SRNet integrating all the designs brings an increase in accuracy.

\begin{figure*}
\centering
    {\includegraphics[width=14cm]{Pics/ST_colorbar.png}}\\
    \setlength{\tabcolsep}{1pt}
    \begin{tabular}{>{\centering\arraybackslash}m{0.6cm}>{\centering\arraybackslash}m{2.1cm}>{\centering\arraybackslash}m{2.1cm}>{\centering\arraybackslash}m{2.1cm}>{\centering\arraybackslash}m{2.1cm}>{\centering\arraybackslash}m{2.1cm}>{\centering\arraybackslash}m{2.1cm}>{\centering\arraybackslash}m{2.1cm}}
        (a1) &
        \includegraphics[width=2.1cm]{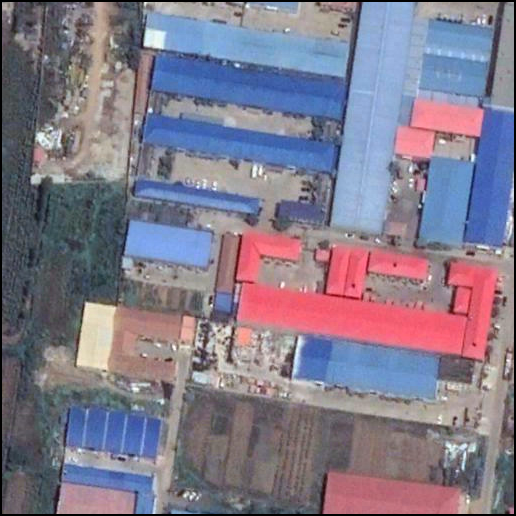} &
        \includegraphics[width=2.1cm]{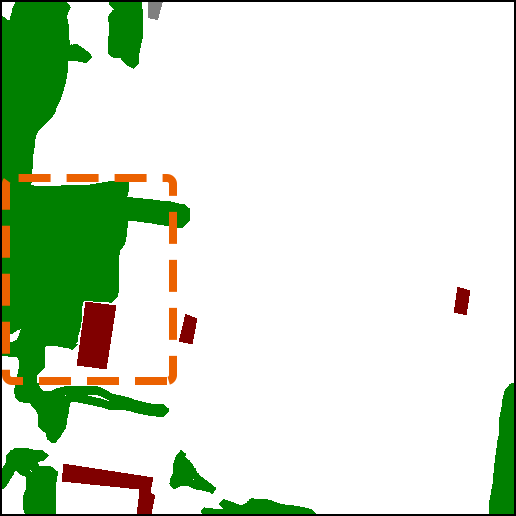} &
        \includegraphics[width=2.1cm]{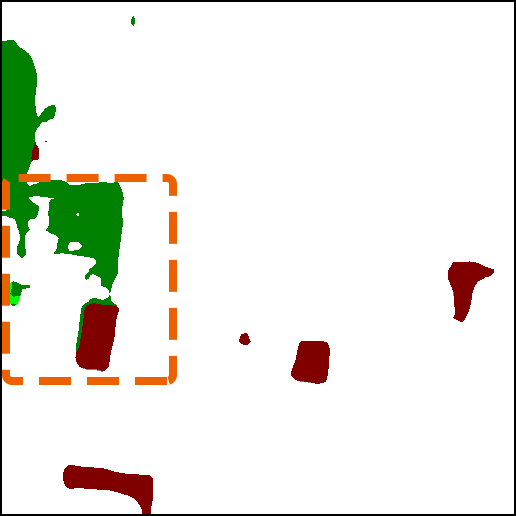} &
        \includegraphics[width=2.1cm]{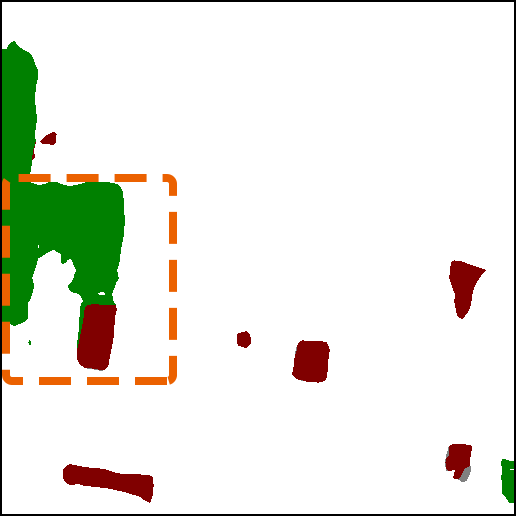} &
        \includegraphics[width=2.1cm]{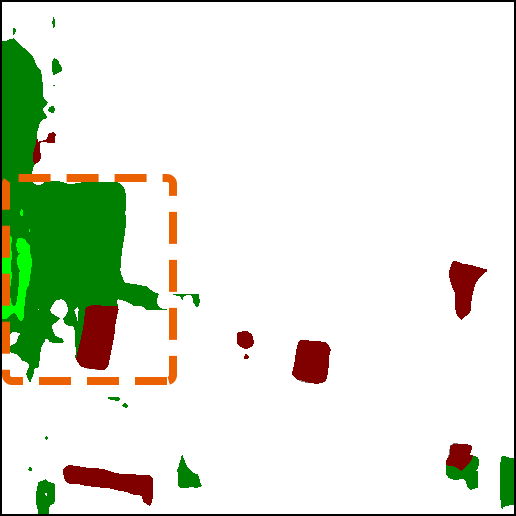} &
        \includegraphics[width=2.1cm]{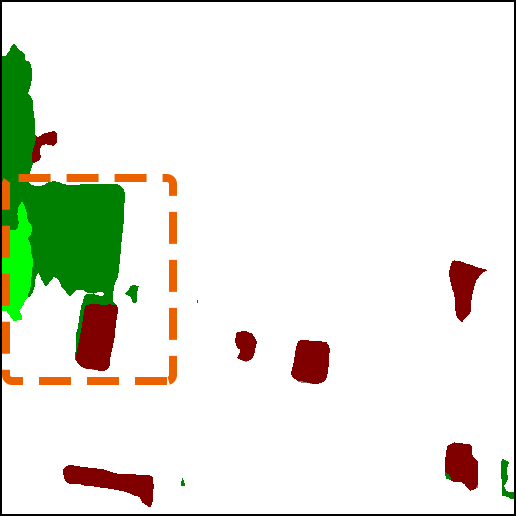} &
        \includegraphics[width=2.1cm]{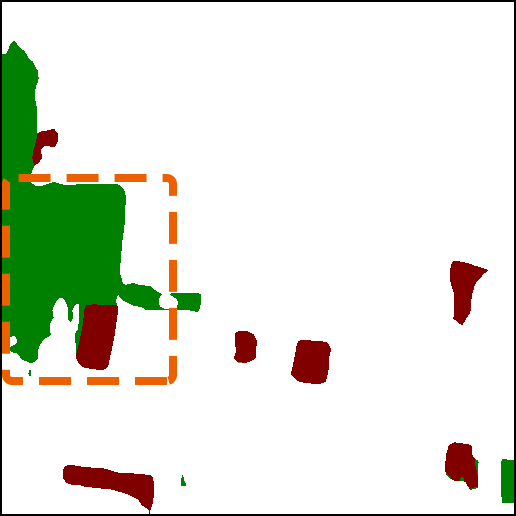} \\
        (a2) &
        \includegraphics[width=2.1cm]{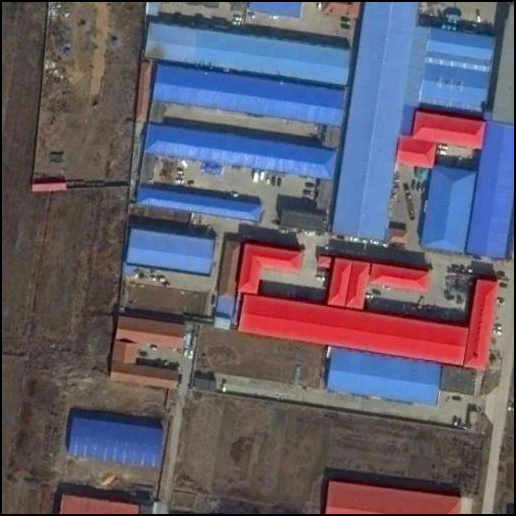} &
        \includegraphics[width=2.1cm]{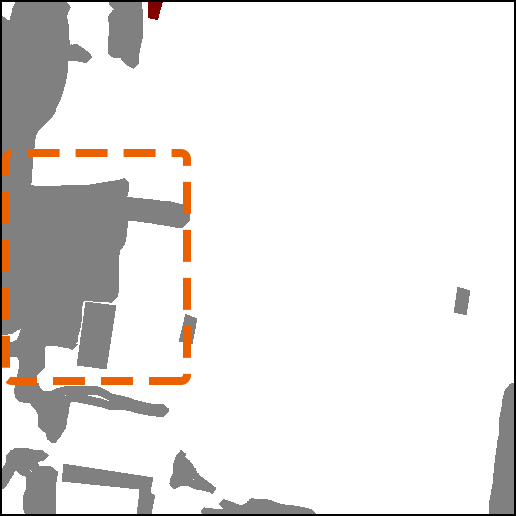} &
        \includegraphics[width=2.1cm]{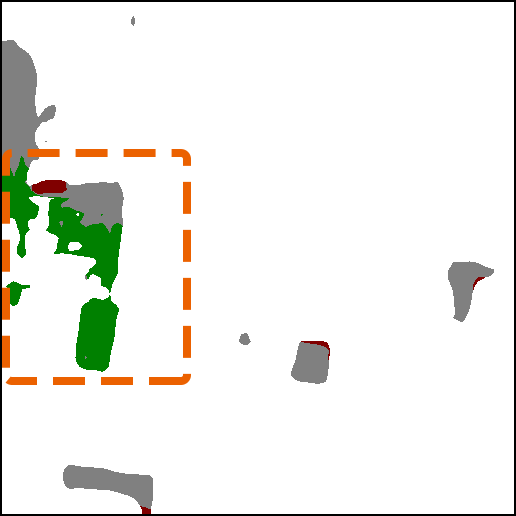} &
        \includegraphics[width=2.1cm]{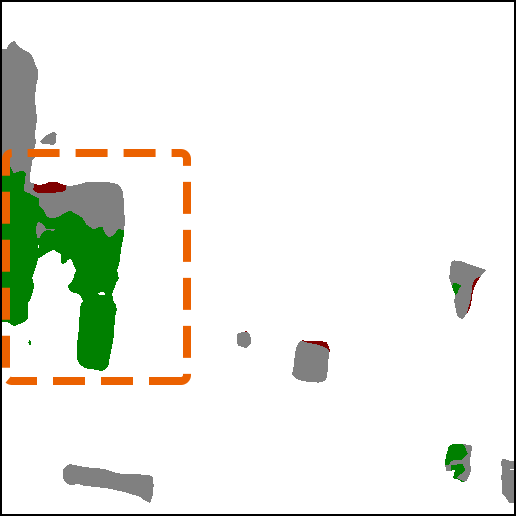} &
        \includegraphics[width=2.1cm]{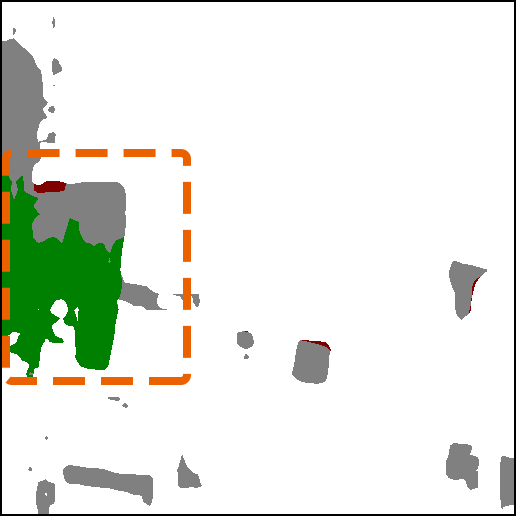} &
        \includegraphics[width=2.1cm]{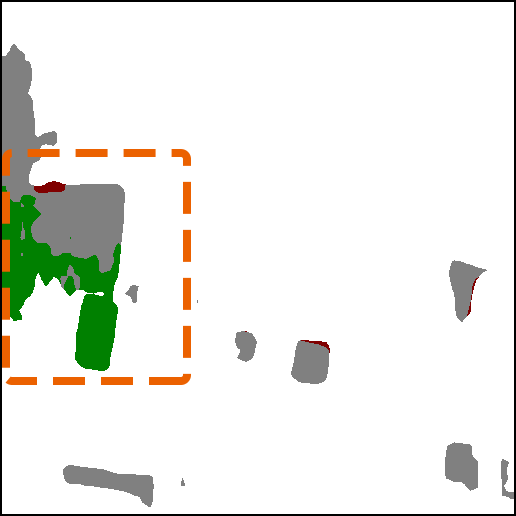} &
        \includegraphics[width=2.1cm]{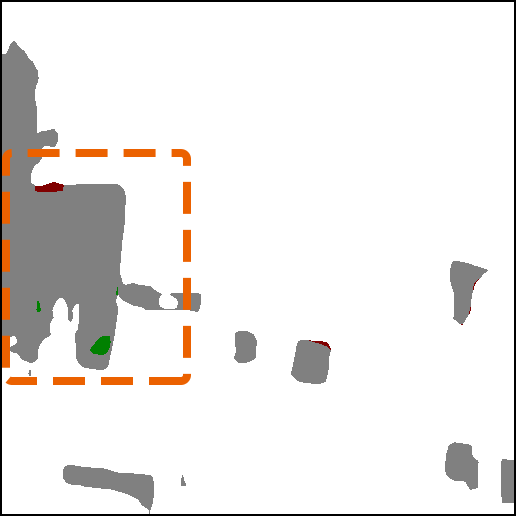} \\
        \hline\\
        (b1) &
        \includegraphics[width=2.1cm]{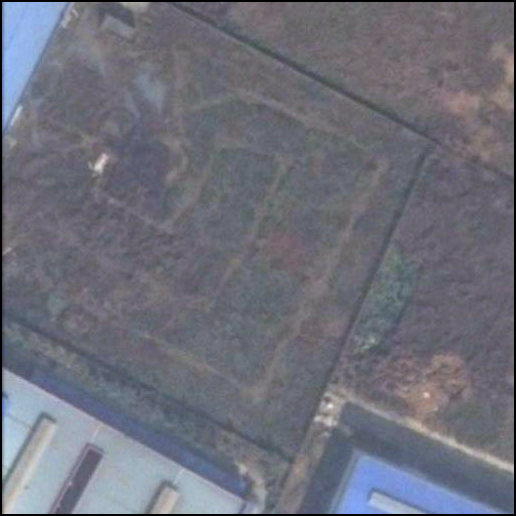} &
        \includegraphics[width=2.1cm]{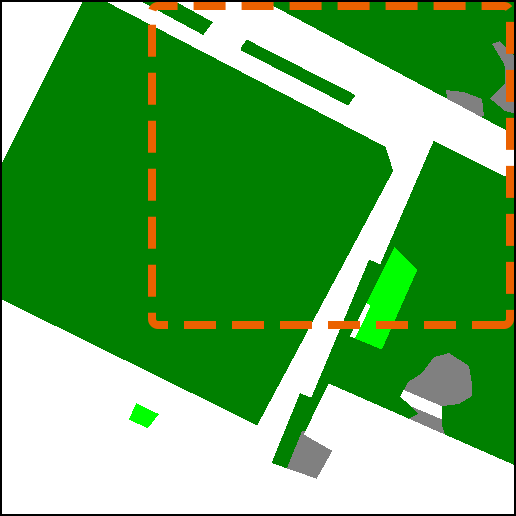} &
        \includegraphics[width=2.1cm]{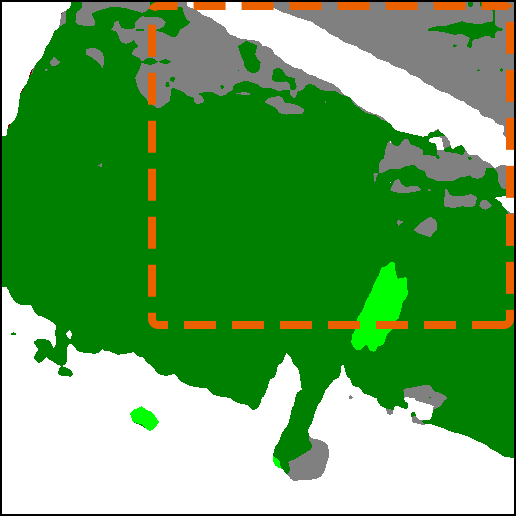} &
        \includegraphics[width=2.1cm]{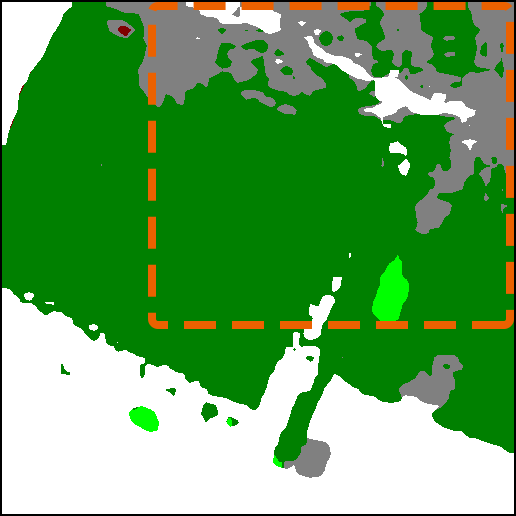} &
        \includegraphics[width=2.1cm]{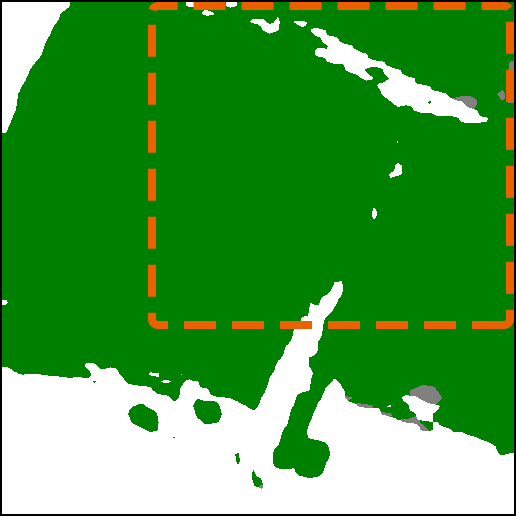} &
        \includegraphics[width=2.1cm]{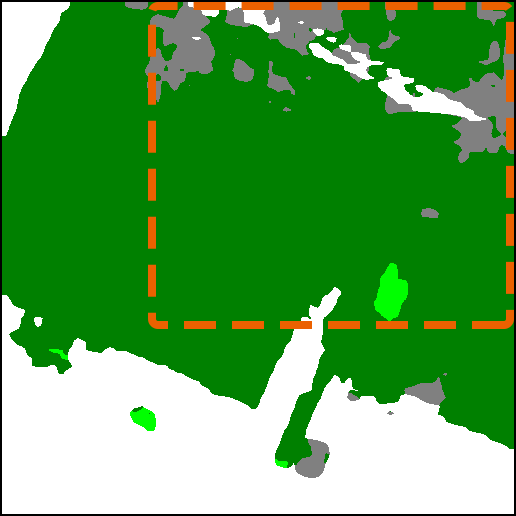} &
        \includegraphics[width=2.1cm]{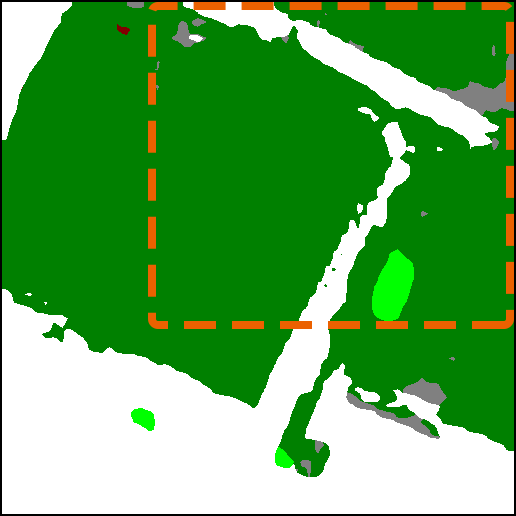} \\
        (b2) &
        \includegraphics[width=2.1cm]{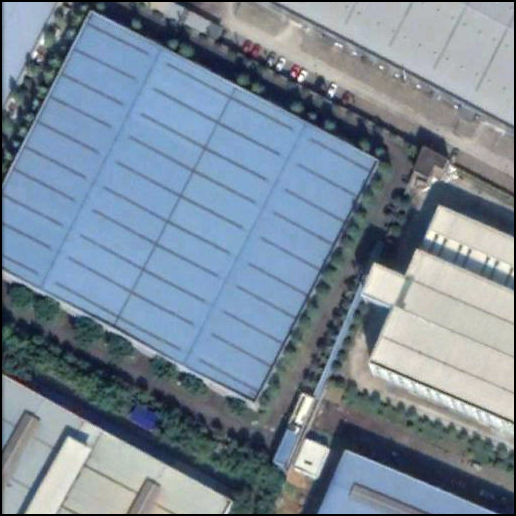} &
        \includegraphics[width=2.1cm]{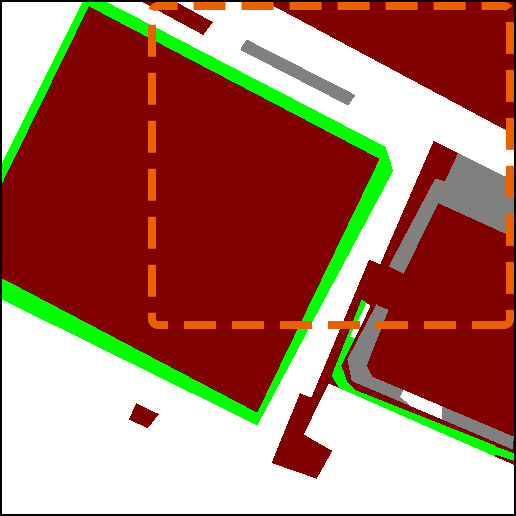} &
        \includegraphics[width=2.1cm]{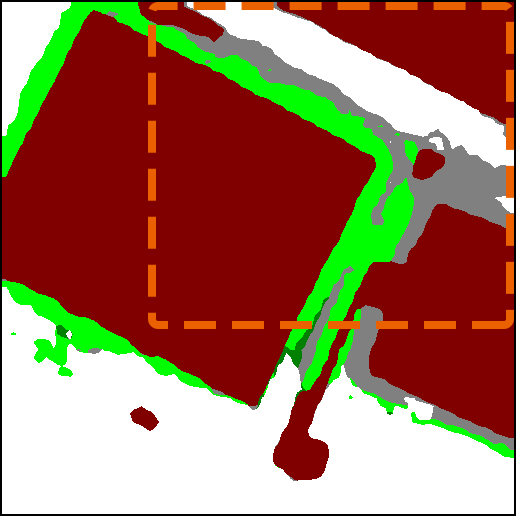} &
        \includegraphics[width=2.1cm]{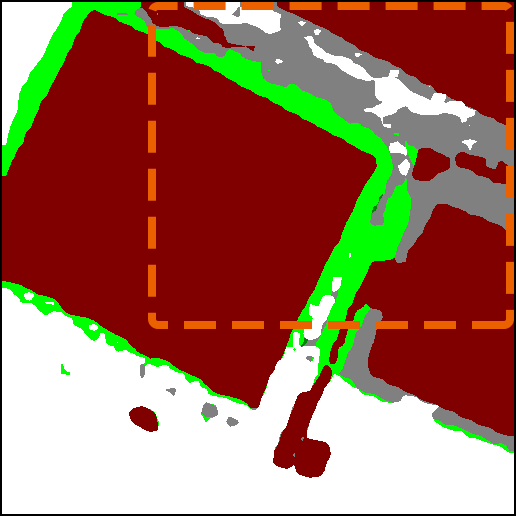} &
        \includegraphics[width=2.1cm]{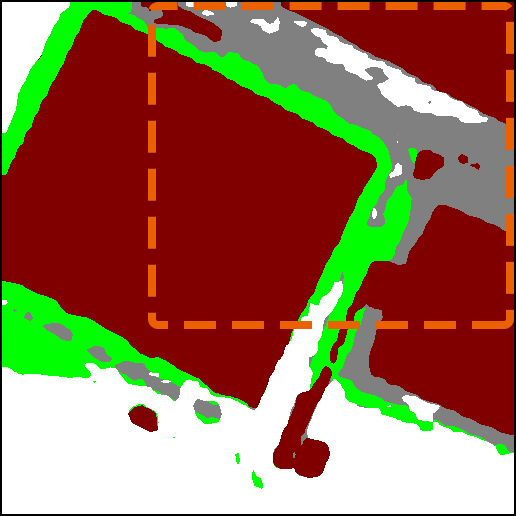} &
        \includegraphics[width=2.1cm]{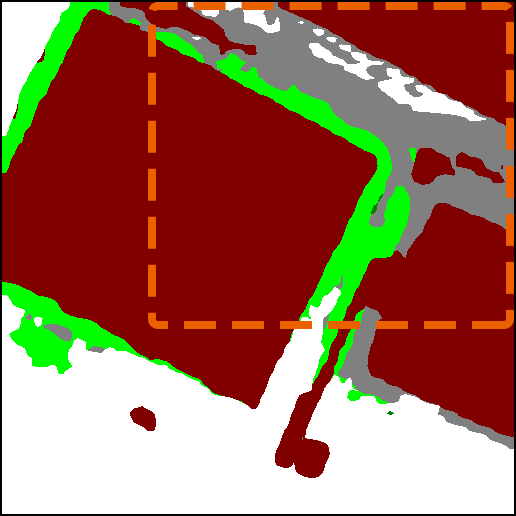} &
        \includegraphics[width=2.1cm]{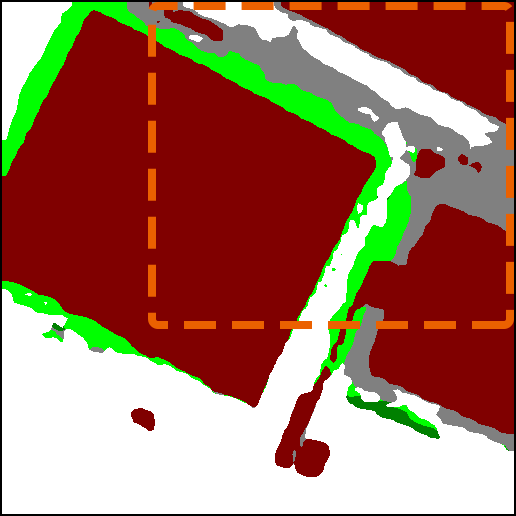} \\
        \hline\\
        (c1) &
        \includegraphics[width=2.1cm]{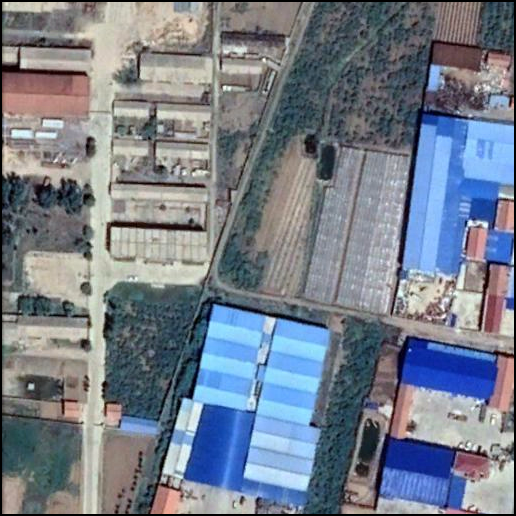} &
        \includegraphics[width=2.1cm]{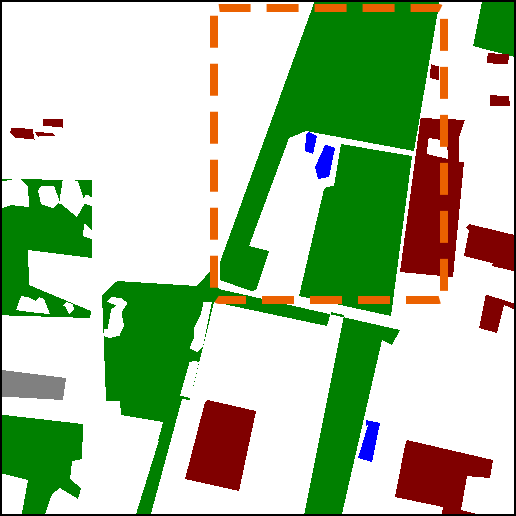} &
        \includegraphics[width=2.1cm]{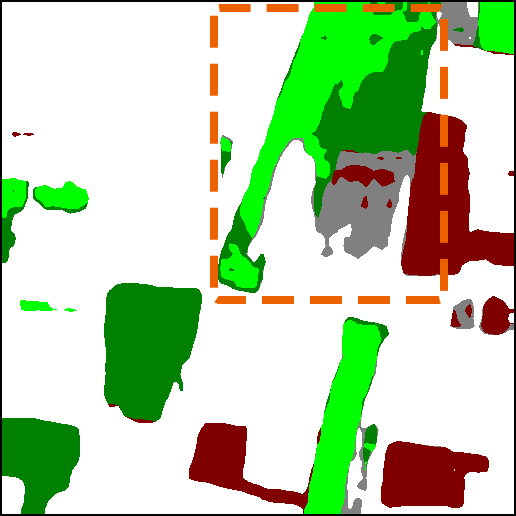} &
        \includegraphics[width=2.1cm]{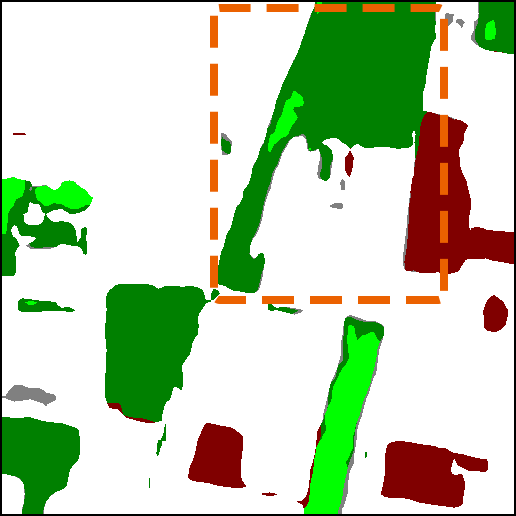} &
        \includegraphics[width=2.1cm]{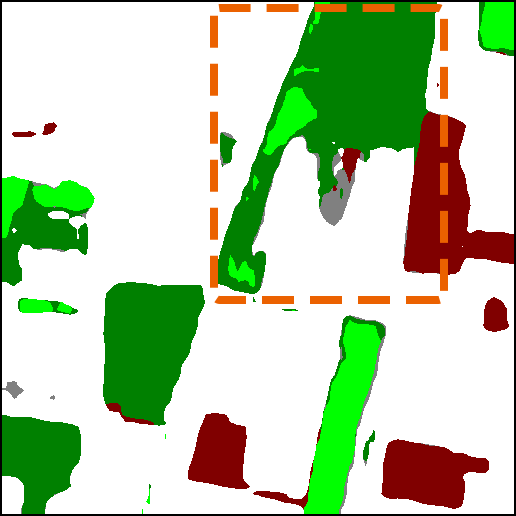} &
        \includegraphics[width=2.1cm]{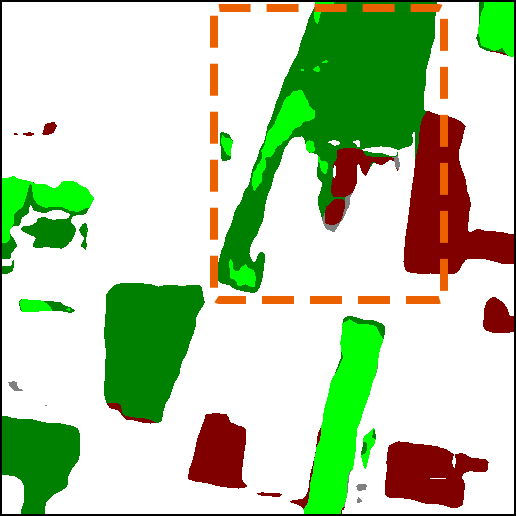} &
        \includegraphics[width=2.1cm]{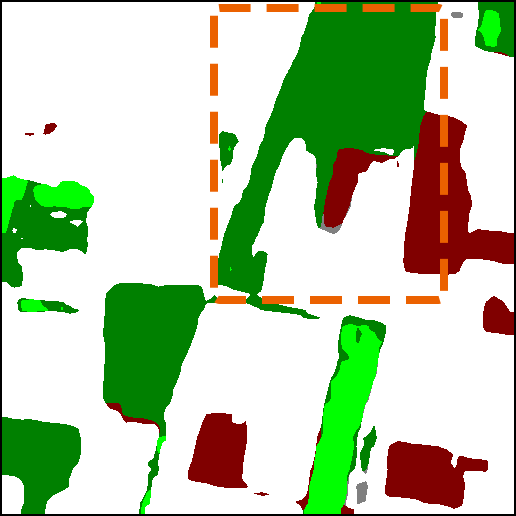} \\
        (c2) &
        \includegraphics[width=2.1cm]{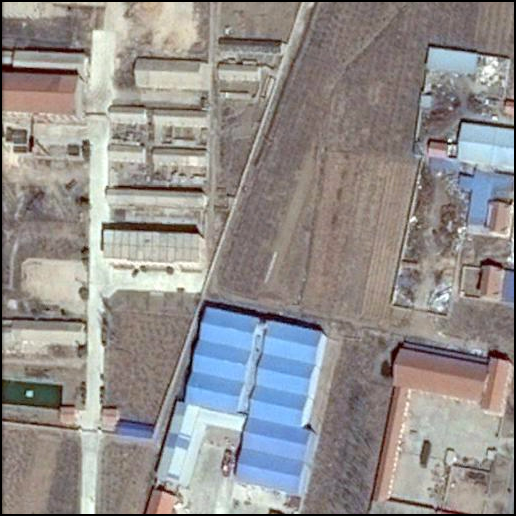} &
        \includegraphics[width=2.1cm]{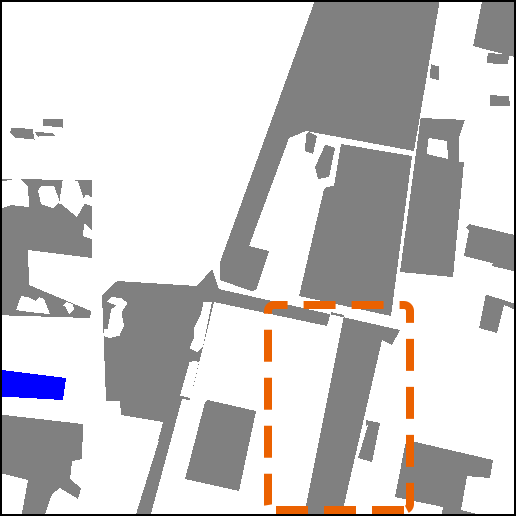} &
        \includegraphics[width=2.1cm]{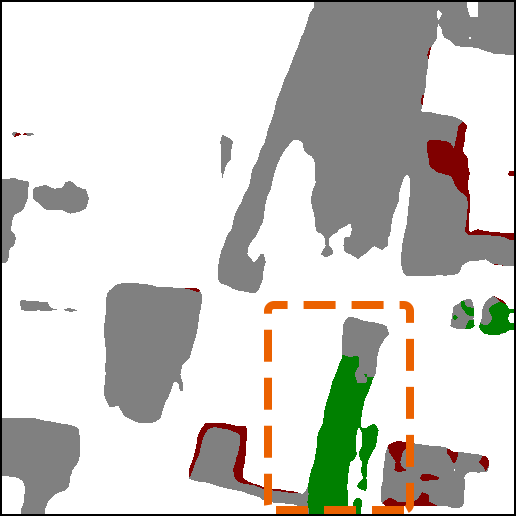} &
        \includegraphics[width=2.1cm]{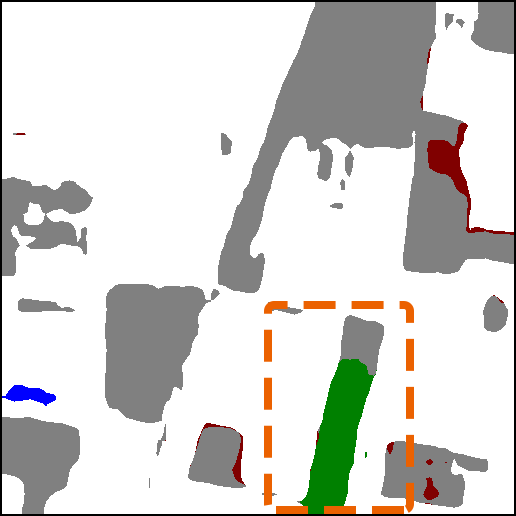} &
        \includegraphics[width=2.1cm]{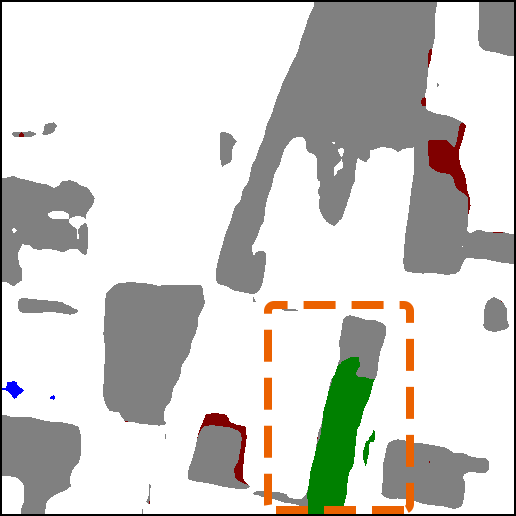} &
        \includegraphics[width=2.1cm]{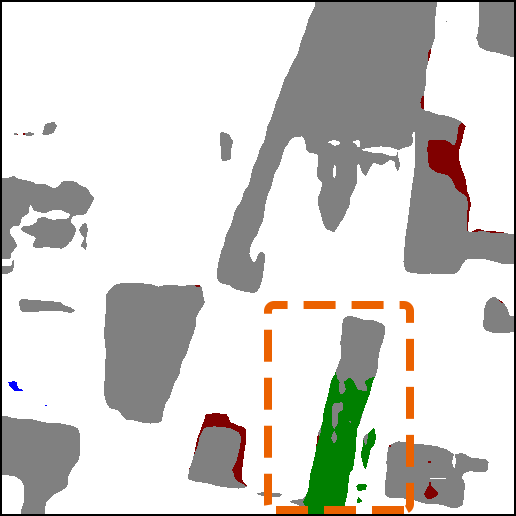} &
        \includegraphics[width=2.1cm]{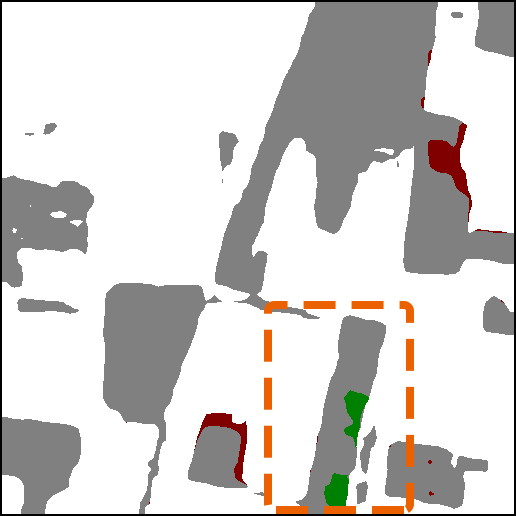} \\
        & Test image & GT & SSCD-l & SCLoss-SSCDl & SiamSR-SSCDl & CotSR-SSCDl & Bi-SRNet\\
    \end{tabular}
    \caption{Example of results provided by different proposed methods in the ablation study. The major differences are highlighted in orange rectangles.} \label{Fig.Ablation}
\end{figure*}

The qualitative results in some testing areas are presented in Fig.\ref{Fig.Ablation}. The prediction maps from left to right are provided for the proposed methods in Table \ref{Table.Ablation}, which are organized in the sequence of number of contained components. Compared to the results of the standard SSCD-l, the predicted LC categories after introducing the SCLoss and SR blocks are gradually improved. The Bi-SRNet exhibits advantages in discriminating the critical areas. For example, in Fig.\ref{Fig.Ablation}(a2),(b1),(c1) and (c2), identification of the \textit{ground}, \textit{low vegetation} and \textit{tree} classes is greatly improved.

Through this ablation study we find that: i) as indicated by increases in SeK values, all the tested auxiliary components result in improvements in semantic embedding; ii) the semantic reasoning designs in the Bi-SRNet improve not only the discrimination of LC categories, but also the detection of changes.

\begin{table*}[t]
\centering
    \caption{Quantitative results of the ablation study.}
    \resizebox{1\linewidth}{!}{%
        \begin{tabular}{r|cccc|cccc}
        \toprule
            \multirow{2}*{Proposed Method} & \multicolumn{4}{c|}{Components}  & \multicolumn{4}{c}{Accuracy}\\
            \cline{2-9}
            & CD block & siam-SR block & Cot-SR block & SCLoss & OA(\%) & mIoU(\%) & Sek(\%) & $F_{scd}$(\%) \\
            \hline
            SSCD-l & $\surd$ &  &  &  & 87.19 & 72.60 & 21.86 & 61.22 \\
            SCLoss-SSCDl & $\surd$ &  &  & $\surd$ & 87.48 & 73.06 & 22.68 & 61.98 \\
            SiamSR-SSCDl & $\surd$  & $\surd$ & & $\surd$ & 87.73 & 73.45 & 23.15 & 62.41 \\
            CotSR-SSCDl & $\surd$  &  & $\surd$ & $\surd$ & 87.67 & 73.38 & 23.09 & 62.44 \\
            Bi-SRNet & $\surd$ & $\surd$ & $\surd$ & $\surd$ & \textbf{87.84} & \textbf{73.41} & \textbf{23.22} & \textbf{62.61} \\
        \bottomrule
        \end{tabular} }\label{Table.Ablation}
\end{table*}

\subsection{Comparative Experiments}\label{sc5.compareSOTA}

\begin{table*}[t]
    \centering
    \caption{Comparison of the proposed methods with literature methods for the SCD.}
    \resizebox{1\linewidth}{!}{%
        \begin{tabular}{r|c|ccc|cccc}
        \toprule
            \multirow{2}*{Method} & \multirow{2}*{Arch. Type} & \multicolumn{3}{c|}{Computational Costs} & \multicolumn{4}{c}{Accuracy} \\
            \cline{3-9}
            & & Params (Mb) & FLOPs (Gbps)) & Infer. Time (s/100e) & OA(\%) & mIoU(\%) & Sek(\%) & $F_{scd}$(\%)\\
            \hline
            FC-EF \cite{daudt2018fully} & DSCD-e & 1.66 & 17.75 & 0.73 & 85.18 & 64.25 & 9.98 & 48.45\\
            UNet++ \cite{peng2019end} & DSCD-e & 9.16 & 139.82 & 3.46 & 85.18 & 63.83 & 9.90 & 48.04\\
            HRSCD-str.2 \cite{daudt2019multitask} & DSCD-e & 6.39 & 14.29 & 2.33 & 85.49 & 64.43 & 10.69 & 49.22 \\
            ResNet-GRU \cite{mou2018learning} & DSCD-l & 21.45 & 182.36 & 2.73 & 85.09 & 60.64 & 8.99 & 45.89 \\
            ResNet-LSTM \cite{mou2018learning} & DSCD-l & 21.48 & 182.63 & 2.73 & 86.77 & 67.16 & 15.96 & 56.90\\
            FC-Siam-conv. \cite{daudt2018fully} & DSCD-l & 2.74 & 35.01 & 1.21 & 86.92 & 68.86 & 16.36 & 56.41 \\
            FC-Siam-diff \cite{daudt2018fully} & DSCD-l & 1.66 & 21.41 & 1.02 & 86.86 & 68.96 & 16.25 & 56.20 \\
            IFN \cite{zhang2020deeply} & DSCD-l & 35.73 & 329.10 & 9.41 & 86.47 & 68.45 & 14.25 & 53.54 \\
            HRSCD-str.3 \cite{daudt2019multitask} & SSCD-e & 12.77 & 42.67 & 6.26 & 84.62 & 66.33 & 11.97 & 51.62\\
            HRSCD-str.4 \cite{daudt2019multitask} & SSCD-e & 13.71 & 43.69 & 6.37 & 86.62 & 71.15 & 18.80 & 58.21 \\
            \hline
            SSCD-l (proposed) & SSCD-l & 23.31 & 189.57 & 2.75 & 87.19 & 72.60 & 21.86 & 61.22 \\
            Bi-SRNet (proposed) & SSCD-l & 23.39 & 189.91 & 3.42 & \textbf{87.84} & \textbf{73.41} & \textbf{23.22} & \textbf{62.61}\\
        \bottomrule
        \end{tabular} }\label{Table.CompareSOTA}
\end{table*}

\begin{figure*}
\centering
    {\includegraphics[width=14cm]{Pics/ST_colorbar.png}}\\
    \setlength{\tabcolsep}{1pt}
    \begin{tabular}{>{\centering\arraybackslash}m{0.6cm}>{\centering\arraybackslash}m{2.0cm}>{\centering\arraybackslash}m{2.0cm}>{\centering\arraybackslash}m{2.0cm}>{\centering\arraybackslash}m{2.0cm}>{\centering\arraybackslash}m{2.0cm}>{\centering\arraybackslash}m{2.0cm}>{\centering\arraybackslash}m{2.0cm}>{\centering\arraybackslash}m{2.0cm}}
        (a1) &
        \includegraphics[width=2.0cm]{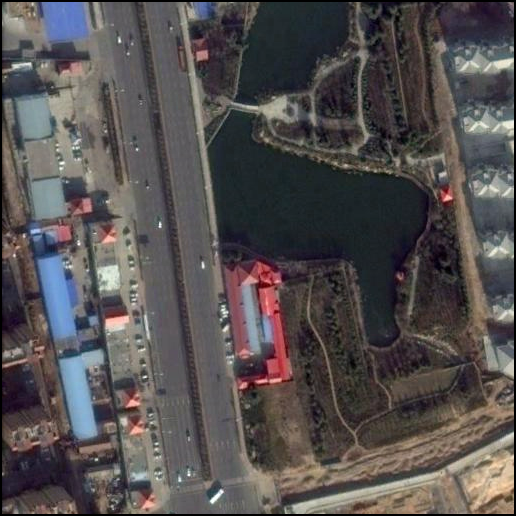} &
        \includegraphics[width=2.0cm]{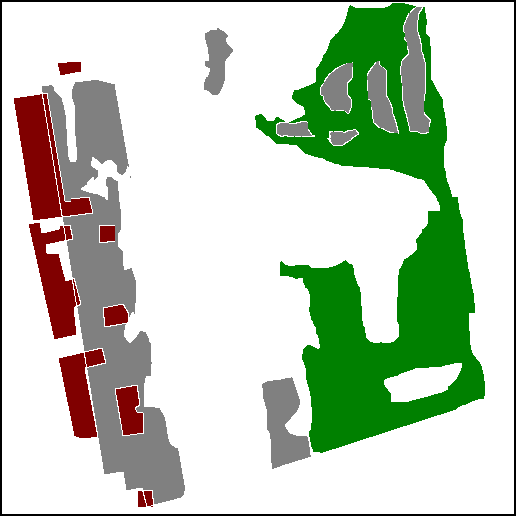} &
        \includegraphics[width=2.0cm]{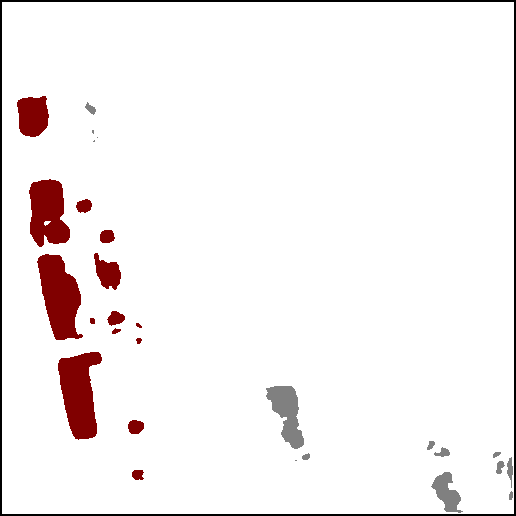} &
        \includegraphics[width=2.0cm]{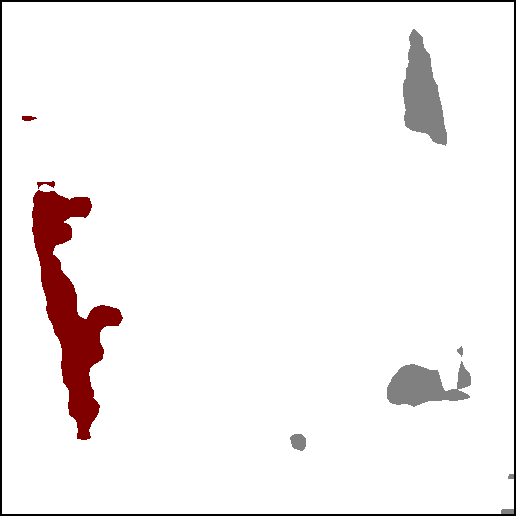} &
        \includegraphics[width=2.0cm]{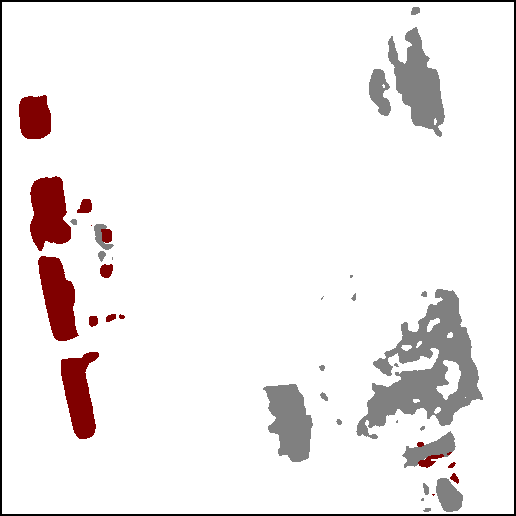} &
        \includegraphics[width=2.0cm]{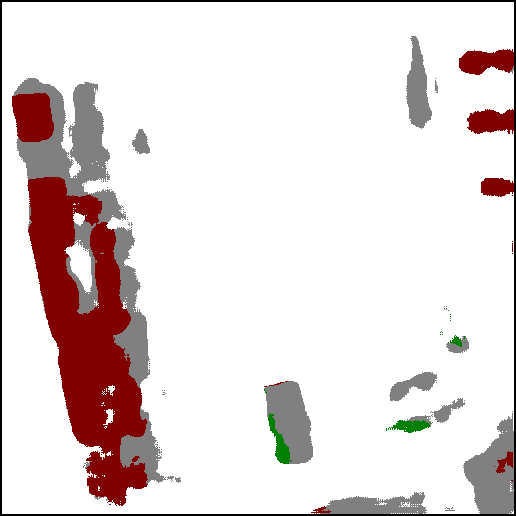} &
        \includegraphics[width=2.0cm]{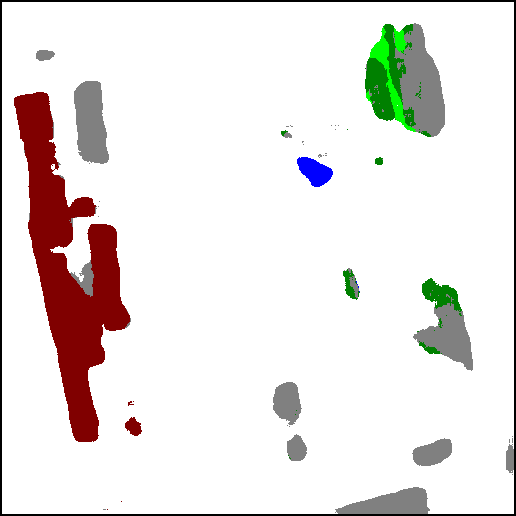} &
        \includegraphics[width=2.0cm]{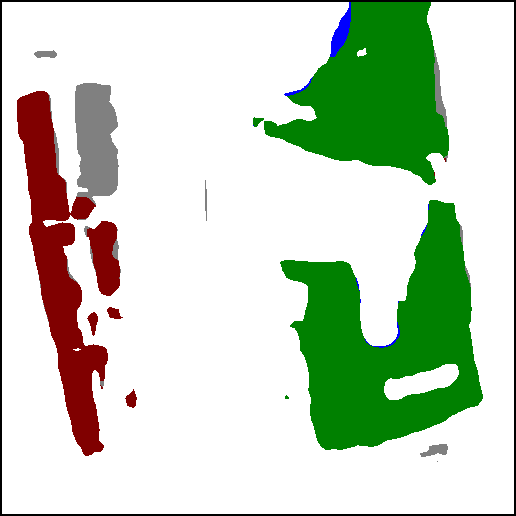} \\
        (a2) &
        \includegraphics[width=2.0cm]{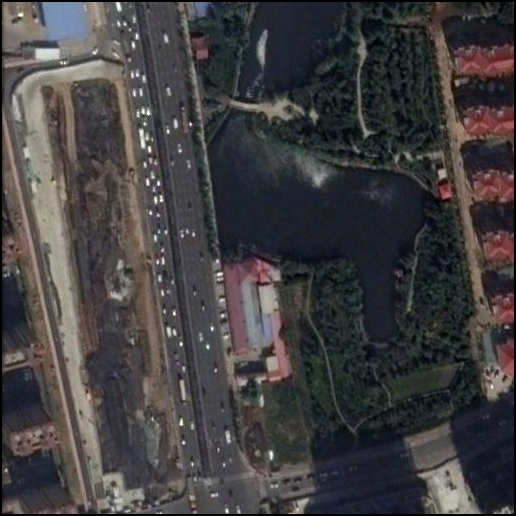} &
        \includegraphics[width=2.0cm]{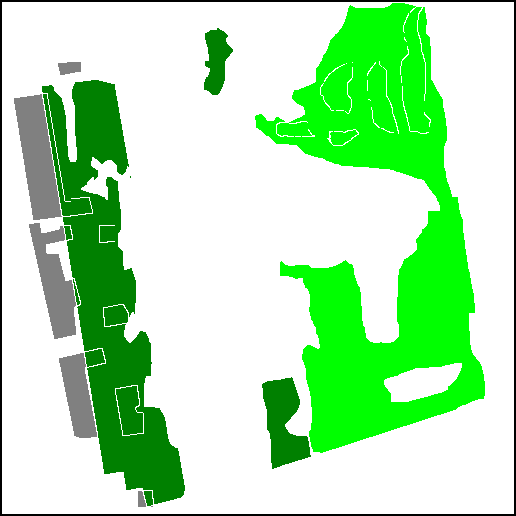} &
        \includegraphics[width=2.0cm]{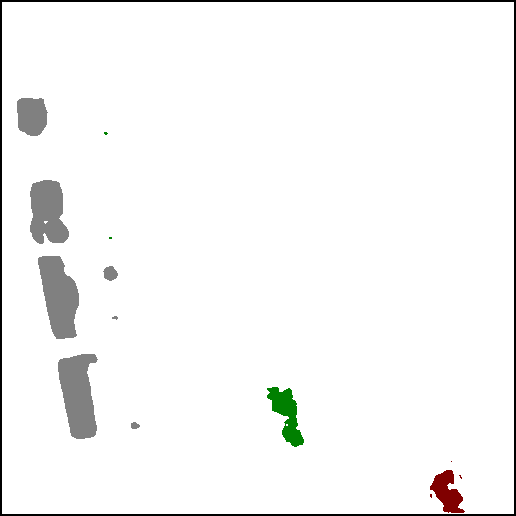} &
        \includegraphics[width=2.0cm]{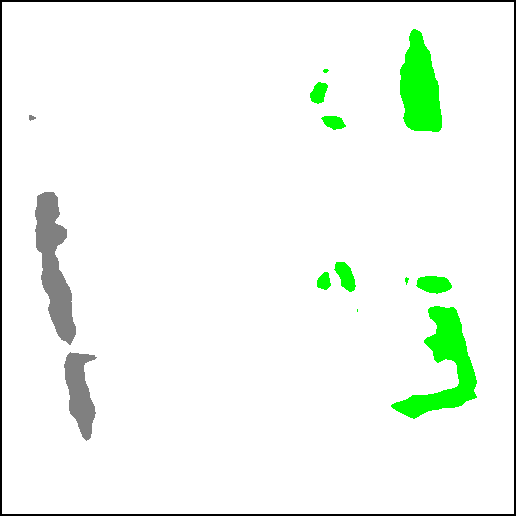} &
        \includegraphics[width=2.0cm]{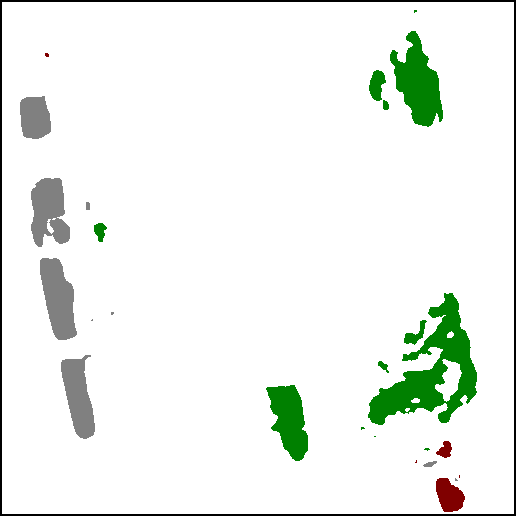} &
        \includegraphics[width=2.0cm]{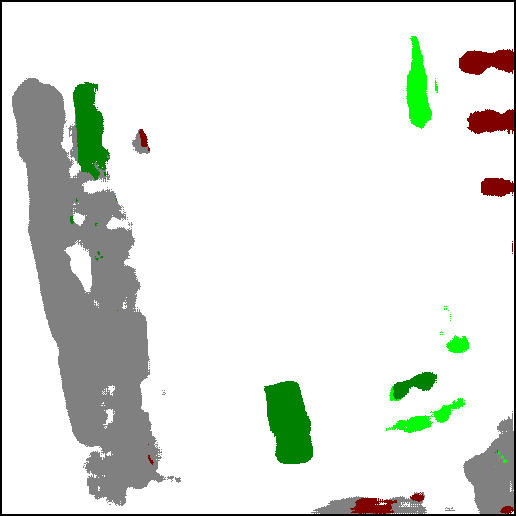} &
        \includegraphics[width=2.0cm]{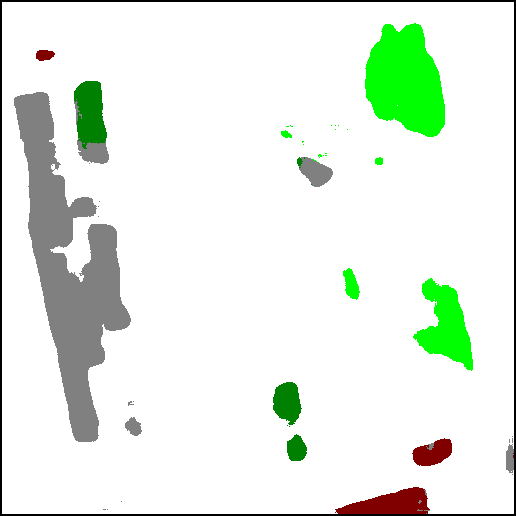} &
        \includegraphics[width=2.0cm]{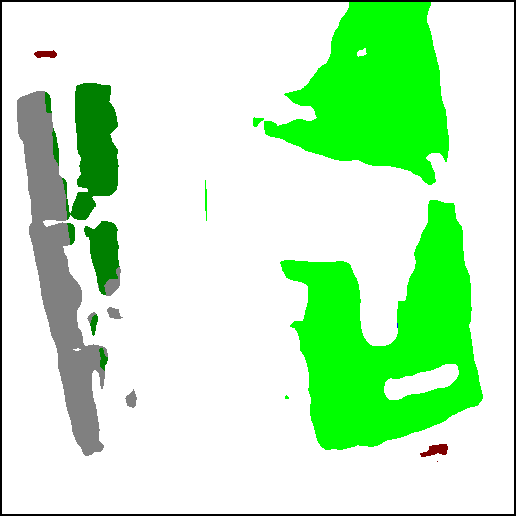} \\
        \hline\\
        (b1) &
        \includegraphics[width=2.0cm]{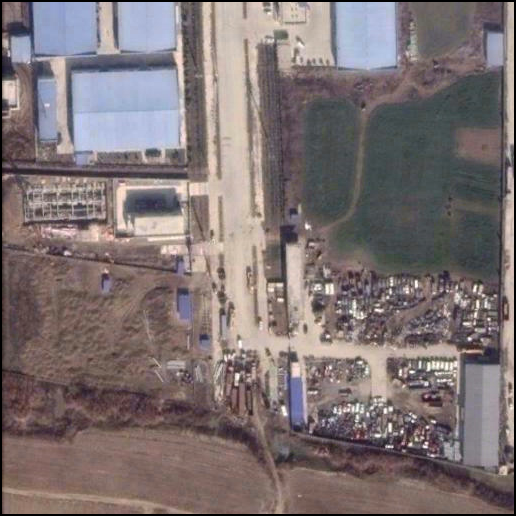} &
        \includegraphics[width=2.0cm]{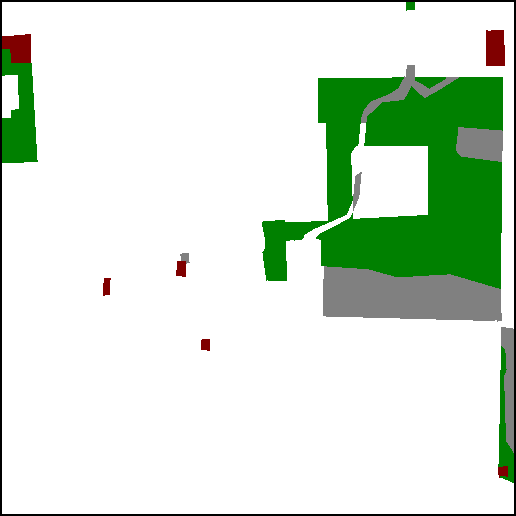} &
        \includegraphics[width=2.0cm]{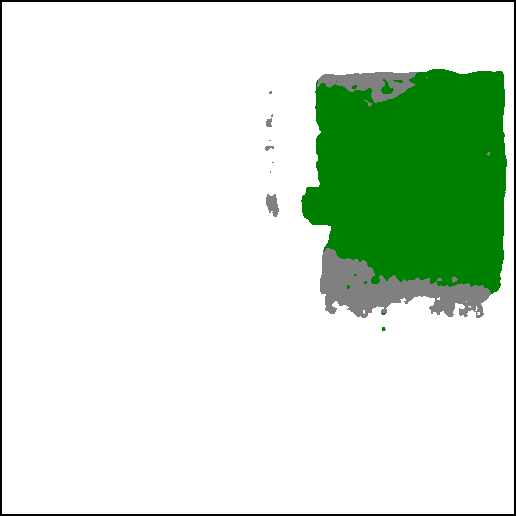} &
        \includegraphics[width=2.0cm]{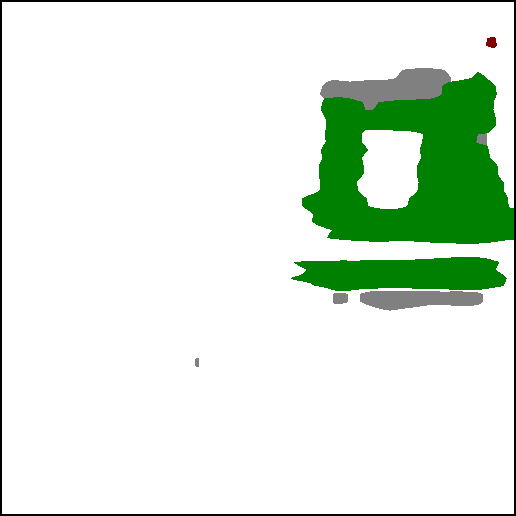} &
        \includegraphics[width=2.0cm]{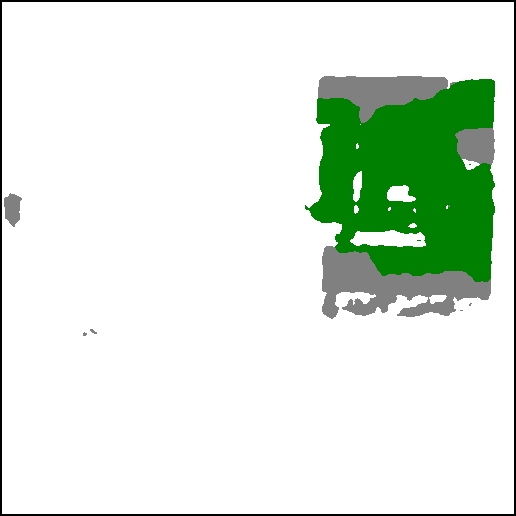} &
        \includegraphics[width=2.0cm]{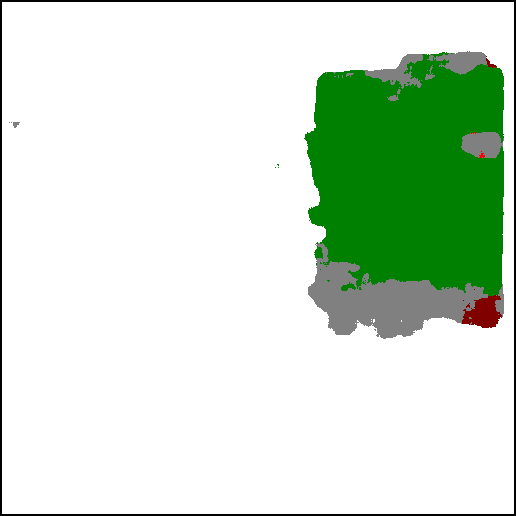} &
        \includegraphics[width=2.0cm]{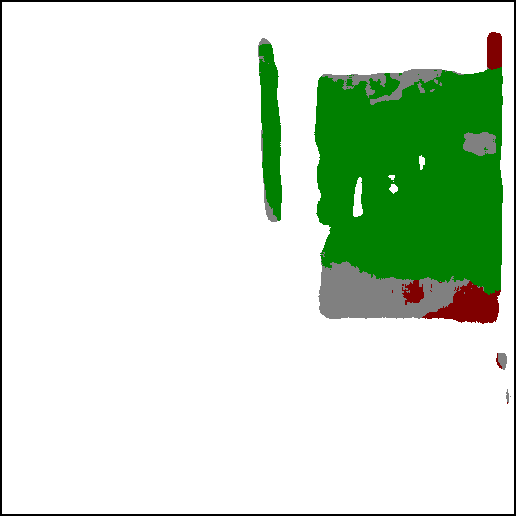} &
        \includegraphics[width=2.0cm]{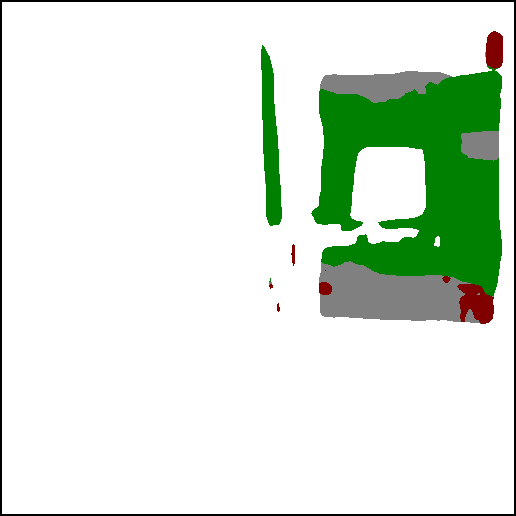} \\
        (b2) &
        \includegraphics[width=2.0cm]{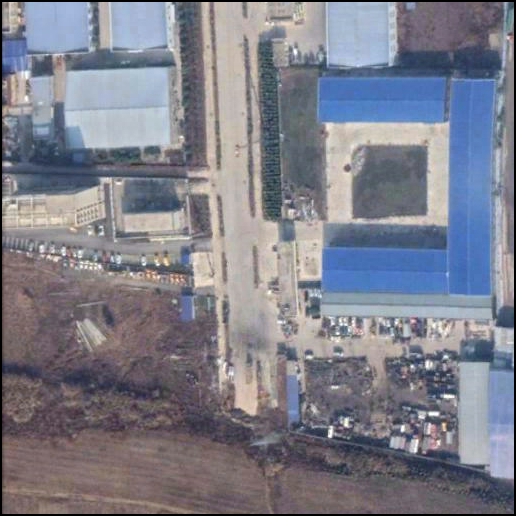} &
        \includegraphics[width=2.0cm]{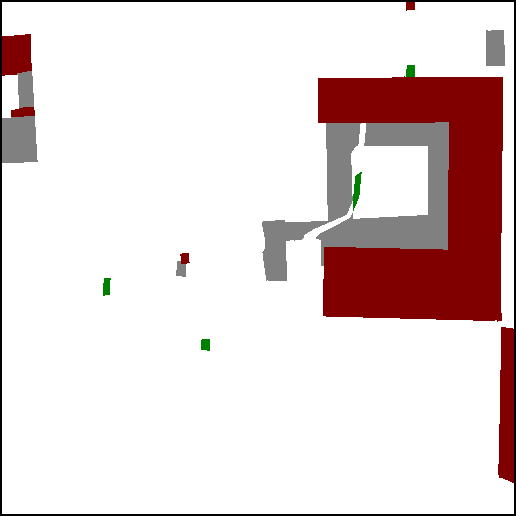} &
        \includegraphics[width=2.0cm]{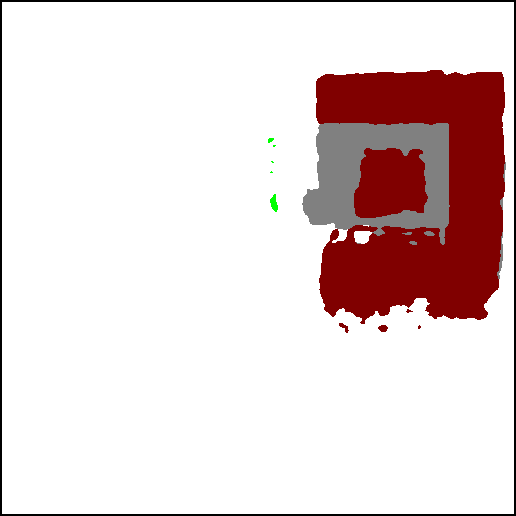} &
        \includegraphics[width=2.0cm]{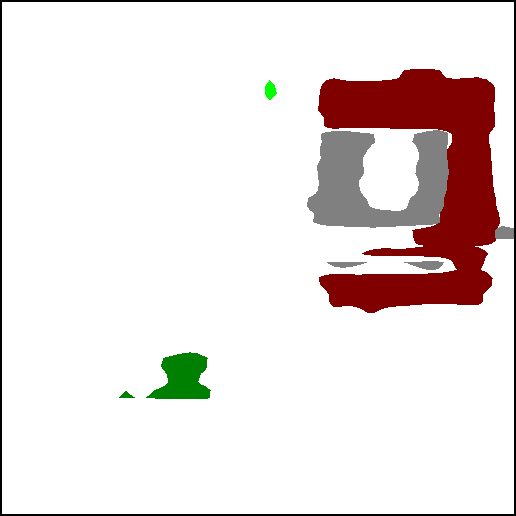} &
        \includegraphics[width=2.0cm]{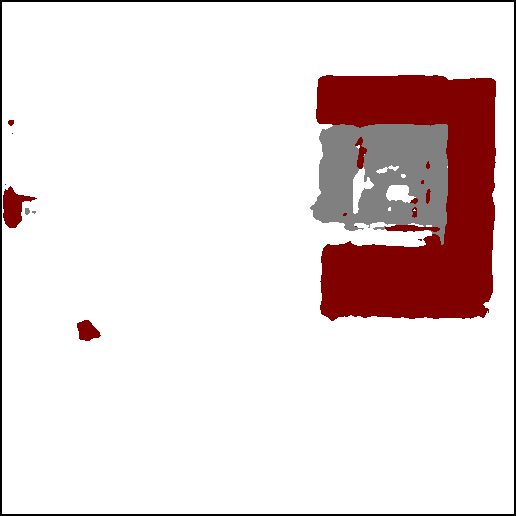} &
        \includegraphics[width=2.0cm]{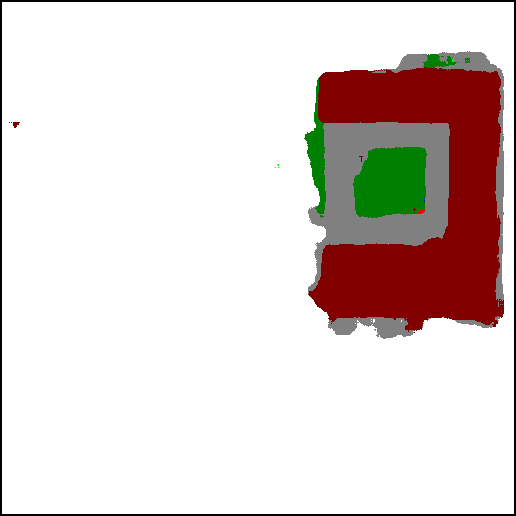} &
        \includegraphics[width=2.0cm]{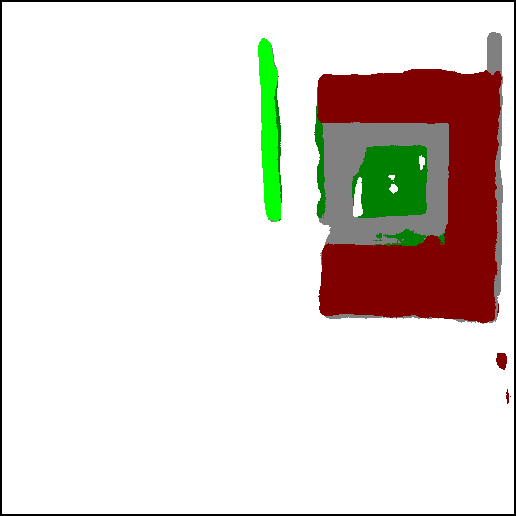} &
        \includegraphics[width=2.0cm]{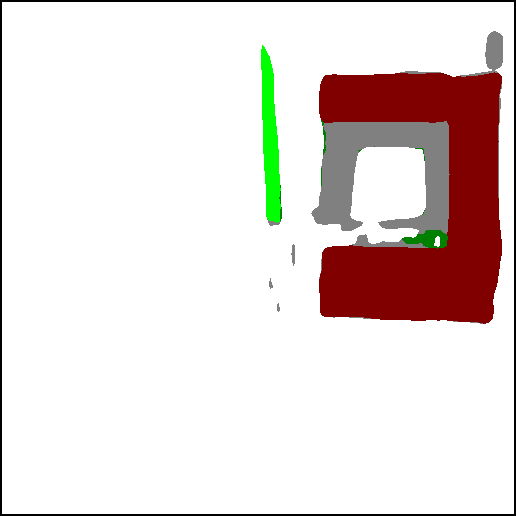} \\
        \hline\\
        (c1) &
        \includegraphics[width=2.0cm]{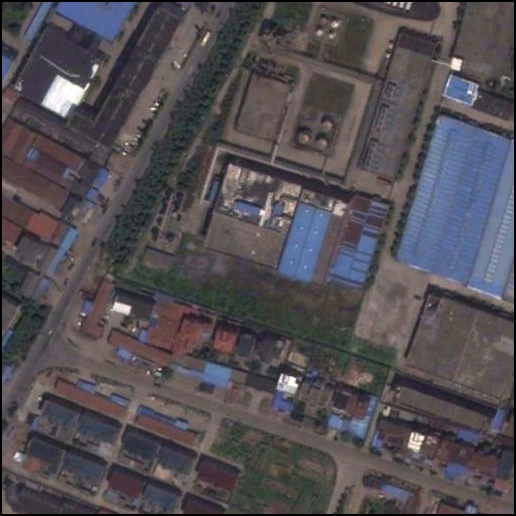} &
        \includegraphics[width=2.0cm]{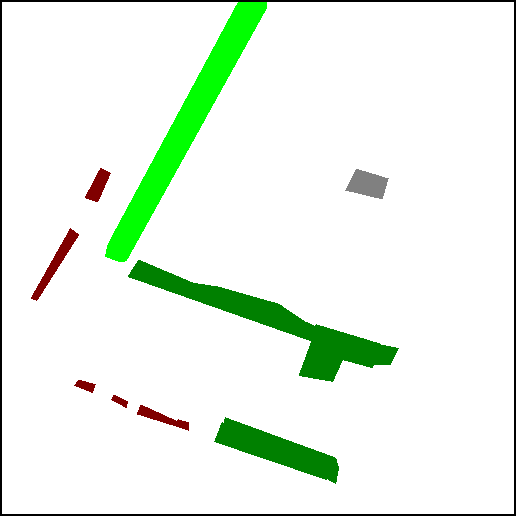} &
        \includegraphics[width=2.0cm]{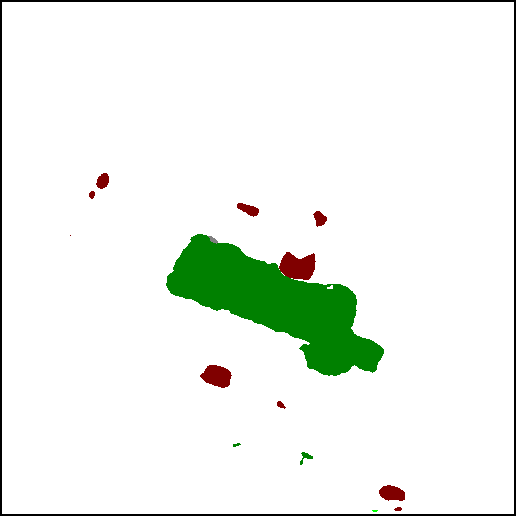} &
        \includegraphics[width=2.0cm]{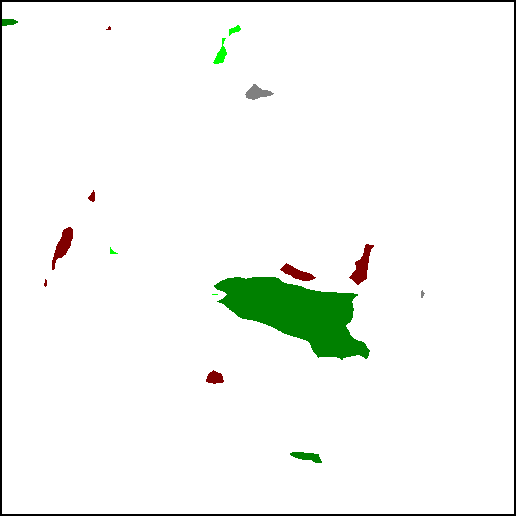} &
        \includegraphics[width=2.0cm]{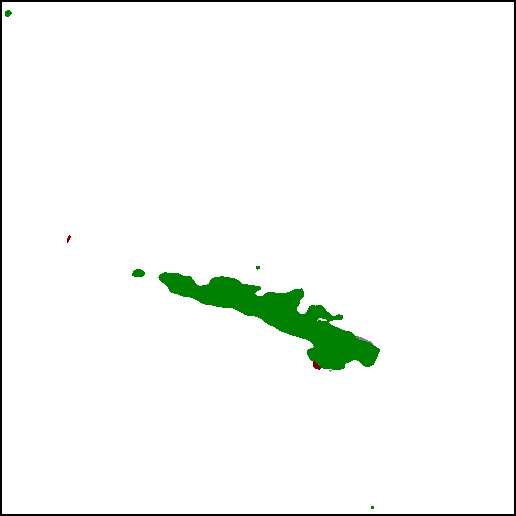} &
        \includegraphics[width=2.0cm]{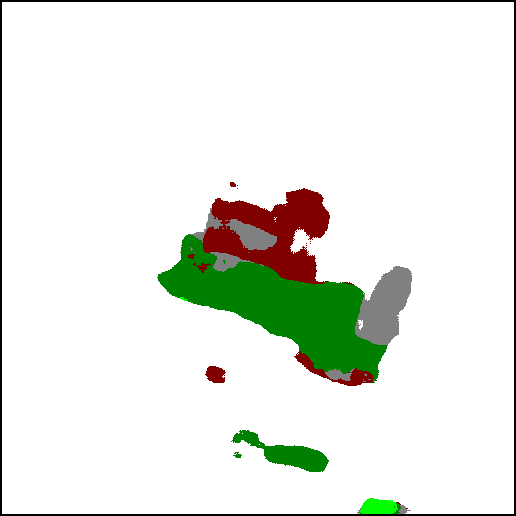} &
        \includegraphics[width=2.0cm]{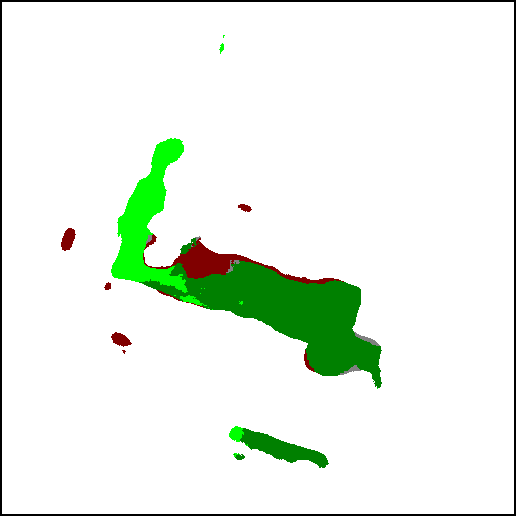} &
        \includegraphics[width=2.0cm]{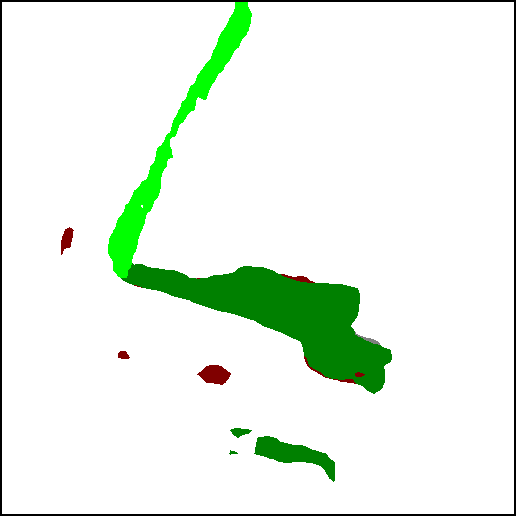} \\
        (c2) &
        \includegraphics[width=2.0cm]{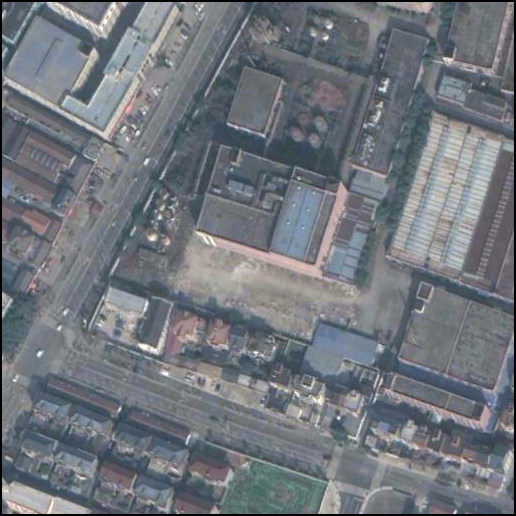} &
        \includegraphics[width=2.0cm]{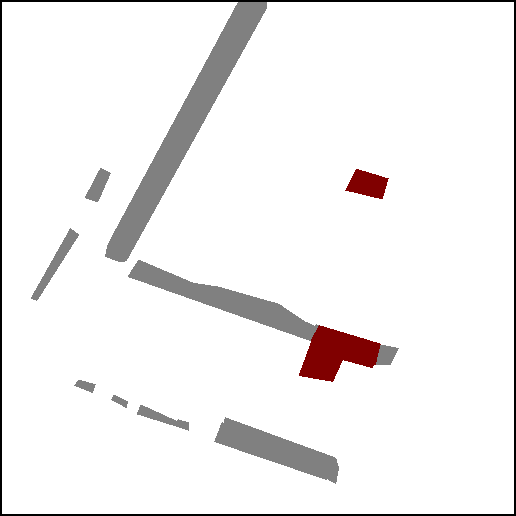} &
        \includegraphics[width=2.0cm]{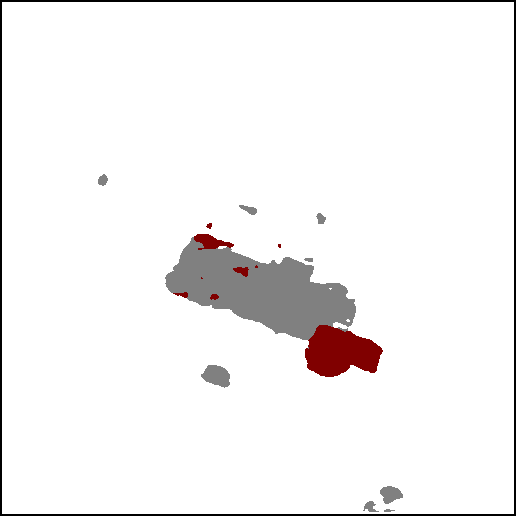} &
        \includegraphics[width=2.0cm]{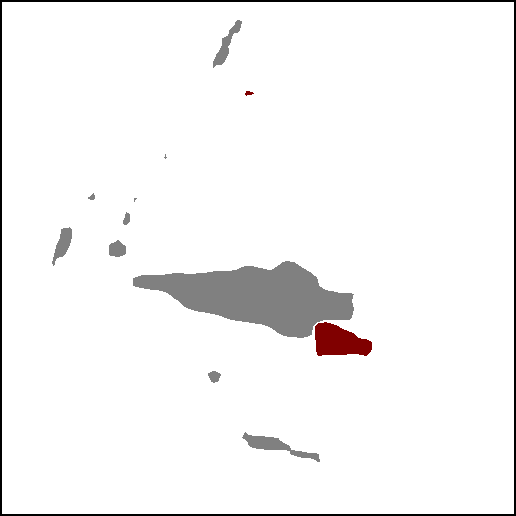} &
        \includegraphics[width=2.0cm]{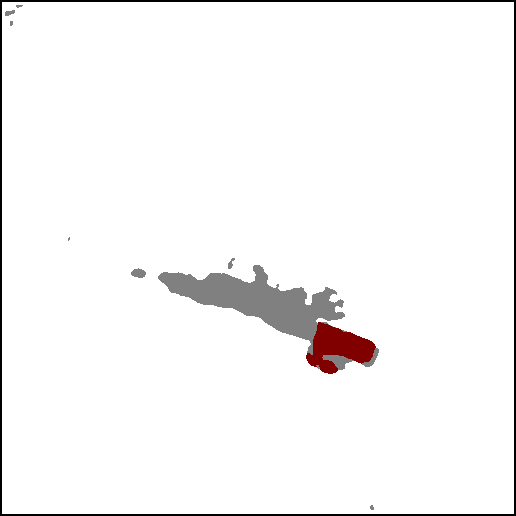} &
        \includegraphics[width=2.0cm]{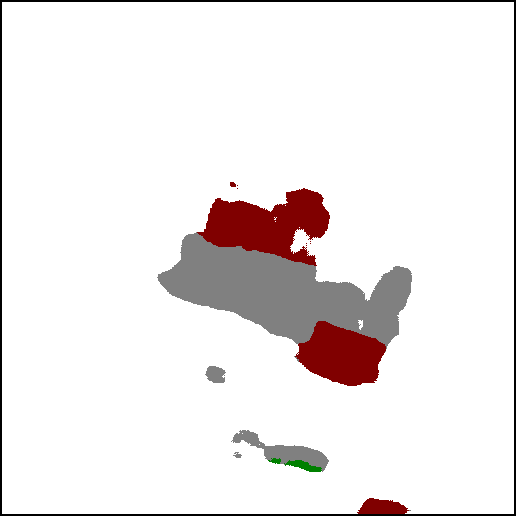} &
        \includegraphics[width=2.0cm]{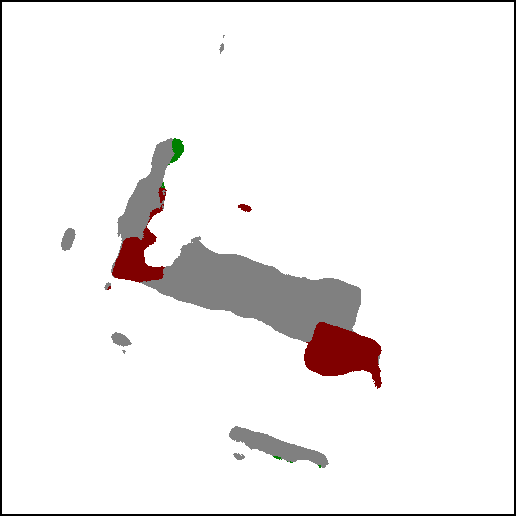} &
        \includegraphics[width=2.0cm]{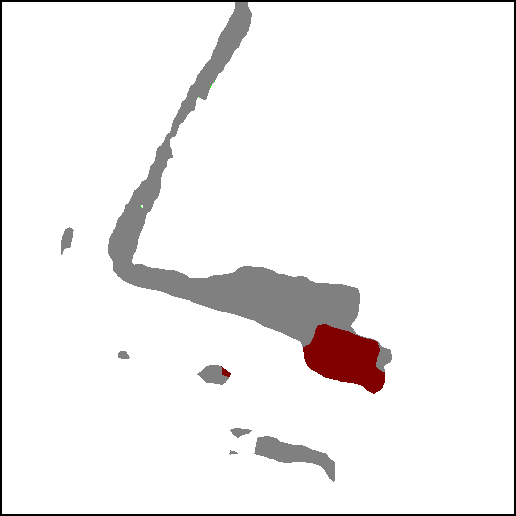}\\
        \hline\\
        (d1) &
        \includegraphics[width=2.0cm]{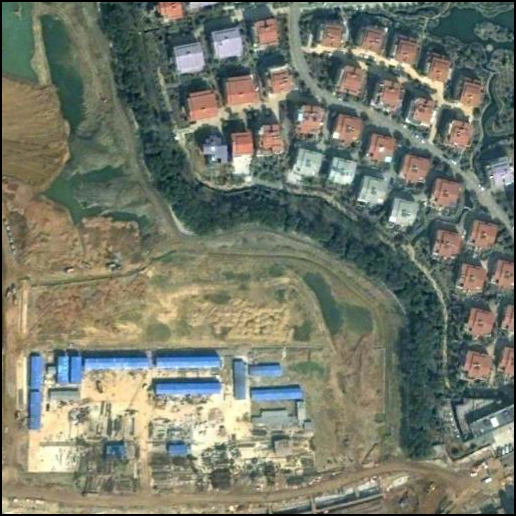} &
        \includegraphics[width=2.0cm]{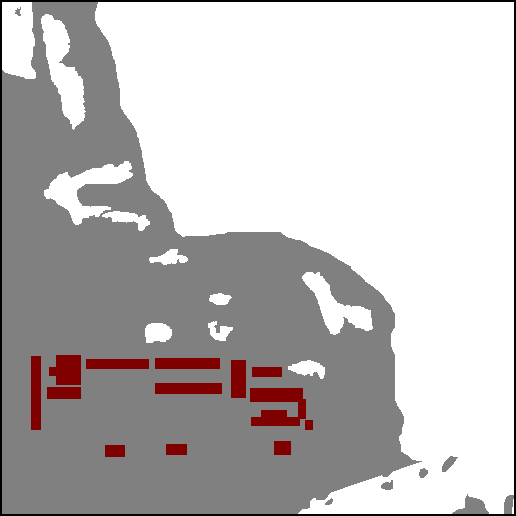} &
        \includegraphics[width=2.0cm]{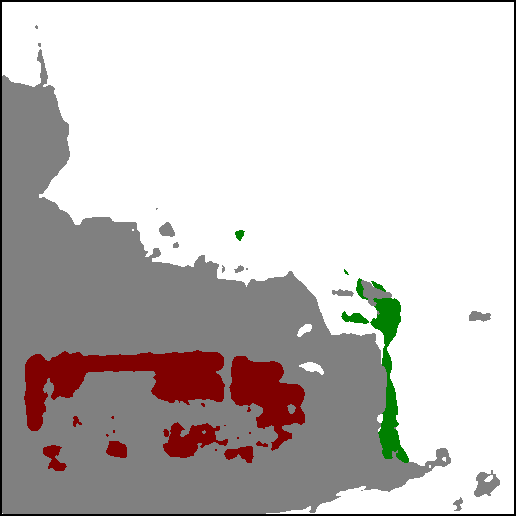} &
        \includegraphics[width=2.0cm]{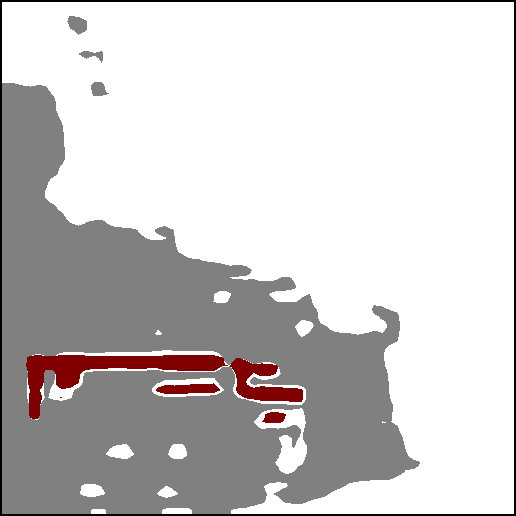} &
        \includegraphics[width=2.0cm]{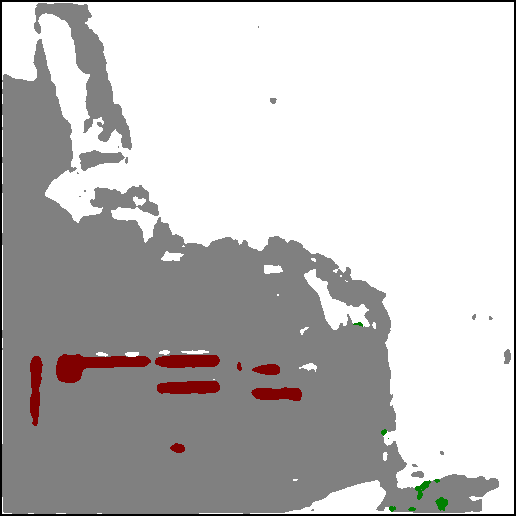} &
        \includegraphics[width=2.0cm]{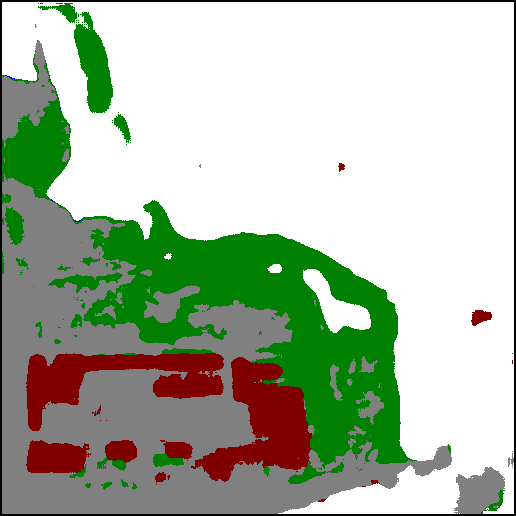} &
        \includegraphics[width=2.0cm]{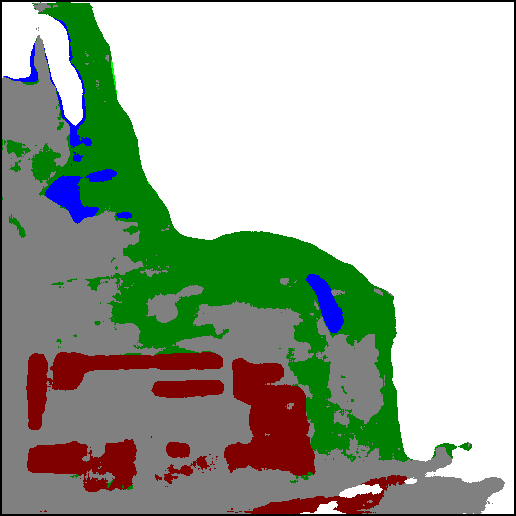} &
        \includegraphics[width=2.0cm]{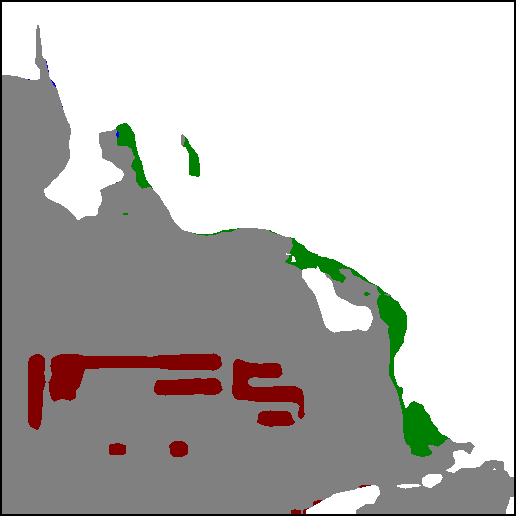} \\
        (d2) &
        \includegraphics[width=2.0cm]{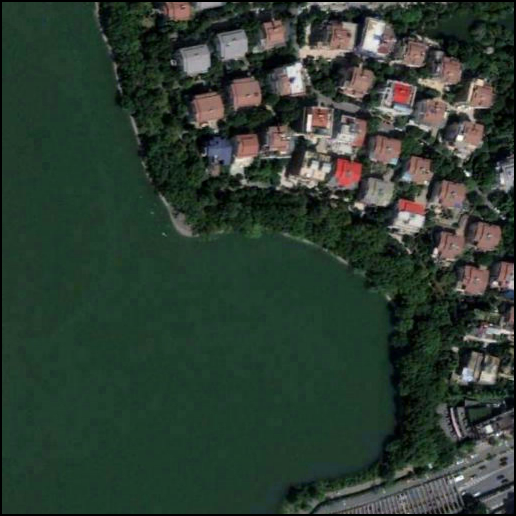} &
        \includegraphics[width=2.0cm]{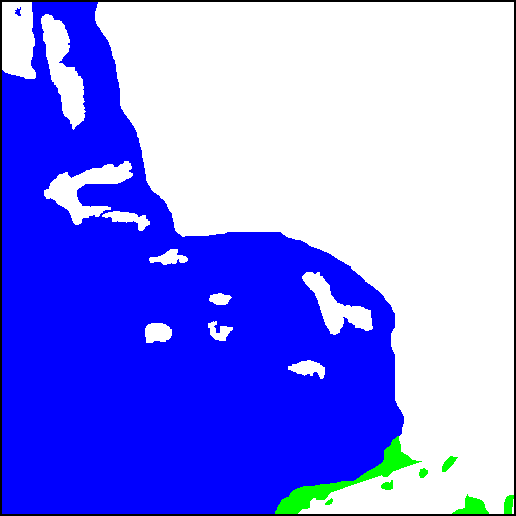} &
        \includegraphics[width=2.0cm]{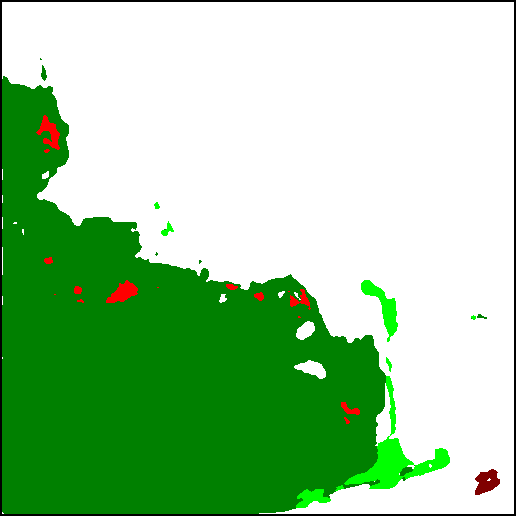} &
        \includegraphics[width=2.0cm]{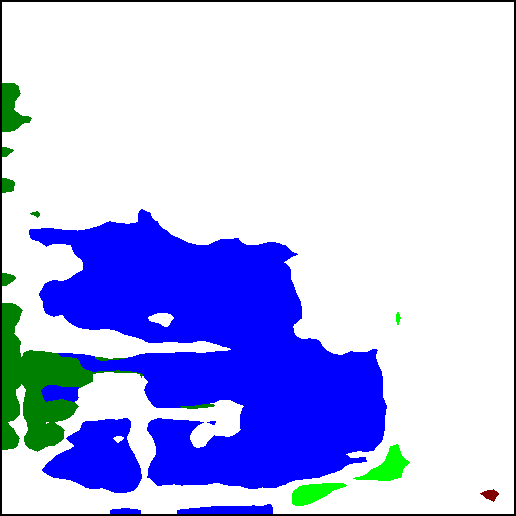} &
        \includegraphics[width=2.0cm]{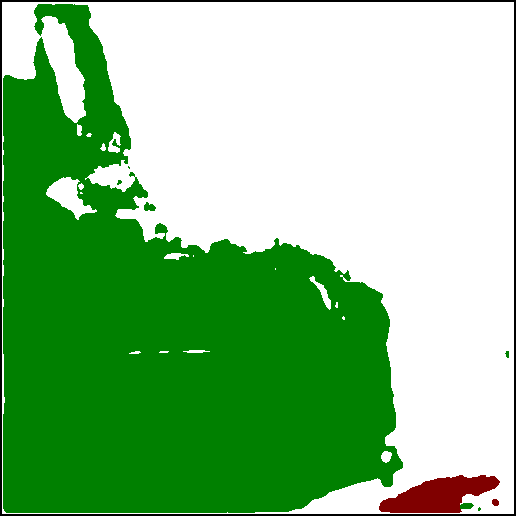} &
        \includegraphics[width=2.0cm]{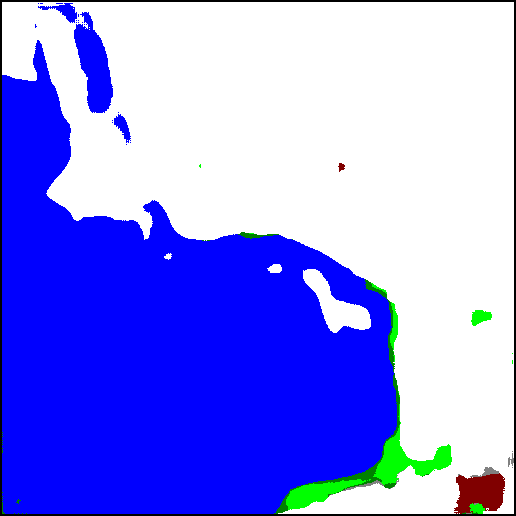} &
        \includegraphics[width=2.0cm]{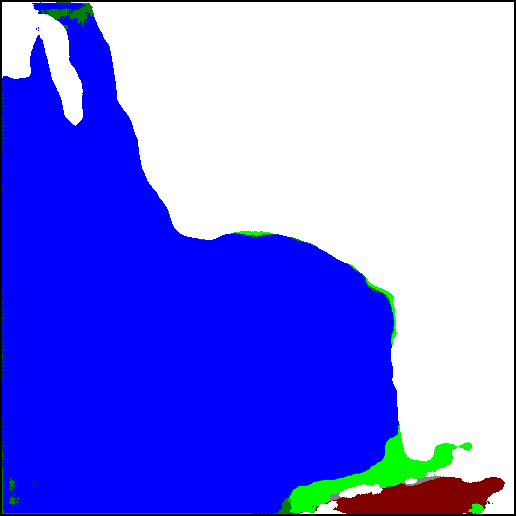} &
        \includegraphics[width=2.0cm]{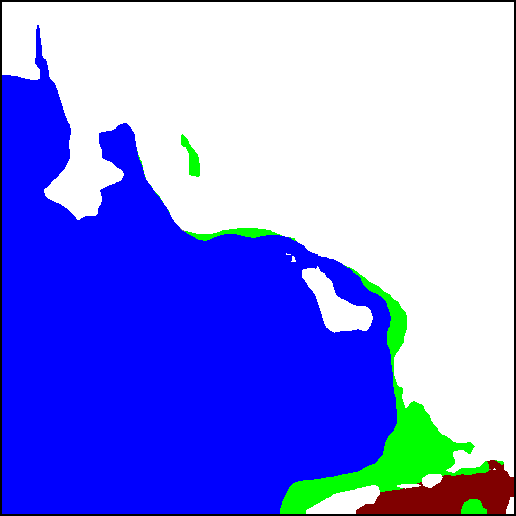} \\
        & Test image & GT & UNet++ & ResNet-LSTM & IFN & HRSCD-str.3 & HRSCD-str.4 & Bi-SRNet (proposed)\\
    \end{tabular}
    \caption{Example of results provided by different methods in the comparative experiments.} \label{Fig.SOTA}
\end{figure*}

To comprehensively evaluate the performance of the proposed SSCD-l architecture and the Bi-SRNet, we further compare them with several SOTA methods in both CD and SCD tasks. The compared methods include:

\textit{1) The FC-EF, FC-Siam-conv and FC-Siam-diff \cite{daudt2018fully}.} These are three UNet-like\cite{ronneberger2015unet} CNNs for binary CD. The FC-EF concatenates temporal images as inputs, whose architecture can be divided into the DSCD-e. The  FC-Siam-conv and FC-Siam-diff both contain siamese encoders while their decoders also serve as CD blocks, which can be divided into the DSCD-l.

\textit{2) The UNet++ \cite{peng2019end}.} This model is a variant of the UNet and can be divided into the DSCD-e. Note that the deep supervision function introduced in \cite{peng2019end} is not suitable for the SCD task, thus it is removed.

\textit{3) The HRSCD-str.2, HRSCD-str.3 and HRSCD-str.4 \cite{daudt2019multitask}.} These are three methods introduced for the SCD, both containing residual blocks \cite{he2016resnet} and encoder-decoder structures. The HRSCD-str.2 directly produces semantic change maps, thus belongs to the type DSCD-e. The HRSCD-str.3 and HRSCD-str.4 both contain triple encoding branches, thus belong to the type SSCD-e.

\textit{4) The ResNet-GRU and ResNet-LSTM \cite{mou2018learning}.} These methods are derived from the methods in \cite{mou2018learning} that combines CNN and RNNs for CD. Since the original methods are designed to classify low-resolution RSIs and contain only few convolutional layers, which are not suitable for processing HR RSIs, we further change their encoders into the ResNet34 \cite{he2016resnet}. The RNN units serve as CD blocks, thus these methods belong to the DSCD-l.

\textit{5) The IFN \cite{zhang2020deeply}.} This method contains a VGGNet \cite{Simonyan2015vgg} encoder and an attention-based decoder. The features are merged in the decoder, thus it belongs to the DSCD-l.

Among the above-mentioned methods, the ones in 3) are aimed to SCD, whereas those in 2), 4) and 5) are originally designed for binary CD. To apply these CD methods for the SCD, we slightly modified their last convolutional layers to meet the required numbers of output maps and channels. The quantitative results are reported in Table \ref{Table.CompareSOTA}. One can observe that the methods using the DSCD-e architecture generally produce unsatisfactory results. In these methods, the modelling of semantic information and change information is entangled, which leads to low SeK values. The FC-Siam-conv and the FC-Siam-diff obtain higher accuracy, which concatenate and merge the semantic features through their decoders. The ResNet-LSTM obtains the highest accuracy among the DSCD-l based methods due to its temporal modelling design. The HRSCD-str.4 (based on the SSCD-e architecture) obtains the highest accuracy amoung literature methods. It contains skip-connections between the temporal branches and the CD branch, which alleviates the drawbacks in the standard SSCD-e. The proposed methods based on the SSCD-l architectures obtain the best accuracy. Without using any CNN decoder or specialized encoder, the proposed Bi-SRNet outperform SOTA by a large margin in all the metrics (1.45\%, 3.06\% and 3\% in mIoU, SeK and $F_{scd}$, respectively). This confirms that the SSCD-l is a better CNN architecture for the SCD.

The computational costs of different methods are reported in Table \ref{Table.CompareSOTA}. Generally, the models with ResNet34 encoders (ResNet-GRU, ResNet-LSTM, SSCD-l and Bi-SRNet) have more parameters and require more calculations. However, their inference time is close to some UNet-based methods (e.g., UNet++ and HRSCD-str.2), since most of the calculations are performed on down-scaled feature maps (at Stage 3 and 4 in the ResNet). The HRSCD-str.3 and HRSCD-str.4 that contain intense calculations in the decoder even require more inference time. The IFN that contains also cascaded attention blocks in the decoder requires highest computation and inference time. Inference time of the proposed methods (SSCD-l and Bi-SRNet) are at the middle level among the compared methods.

To visually assess the results, in Fig.\ref{Fig.SOTA} we present comparisons of the results provided by different methods in several sample areas. One can observe that most of these methods are sensitive to the semantic changes between \textit{building} and \textit{low vegetation}. However, the DSCD-based methods commonly omit some other LC classes. Specifically, the UNet++, the ResNet-LSTM and the IFN omitted many \textit{tree} and \textit{water} areas. The detection of changes is much improved in SSCD-based results. However, the SSCD-e based approaches (HRSCD-str.3 and HRSCD-str.4) mis-classified the LC classes in some critical areas (e.g., confusion between \textit{ground} and \textit{low vegetation} in Fig.\ref{Fig.SOTA}(d1)) and omitted some minor changes (e.g., the \textit{low vegetation} to \textit{tree} changes in Fig.\ref{Fig.SOTA}(a) and the \textit{tree} to \textit{ground} changes in Fig.\ref{Fig.SOTA}(c)). Meanwhile, these change types are all captured by the proposed Bi-SRNet. This further confirms its advantages in both CD and semantic exploitation.

\section{Conclusions}\label{sc6}

In this study, we investigated to improve the SCD task. First, we summarized the existing CNN architectures for SCD and identified the main limitations in semantic embedding and CD. Accordingly, we proposed a novel SCD architecture, the SSCD-l, where the semantic temporal features are merged in a deep convolutional CD block. Then, we further extend the SSCD-l architecture into the Bi-SRNet by introducing several semantic modelling designs. Three auxiliary designs are introduced, including two Siam-SR blocks to augment temporal information, a Cot-SR block to model temporal correlations, as well as a SCLoss to enhance temporal coherence. Finally, a set of experiments have been conducted to evaluate the effectiveness of the proposed methods.

Through experiments we found that: i) The proposed SSCD-l architecture outperforms other standard SCD architectures by a large margin; ii) The proposed Bi-SRNet containing semantic reasoning designs further improves the SSCD-l not only in the segmentation of LC classes, but also in the detection of changes; iii) The SSCD-l based methods (standard SSCD-l and the Bi-SRNet) outperform SOTA methods and obtain the highest accuracy metrics on the SECOND. The advantages of the Bi-SRNet are two-fold. First, in the SSCD-l architecture, the sub-tasks in SCD (SS and CD) are disentangled (with separate outputs and loss functions) but are deeply integrated (through re-use of the semantic features in the CD block). This leads to more accurate CD results. Second, the spatial and temporal correlations are modelled in the Bi-SRNet with both the SR blocks and the SCLoss. This further bridges the sub-tasks in SCD and enables the Bi-SRNet to discriminate better the LC classes in critical areas.

One of the remaining problems is to model the temporal correlation of LCLU classes especially in the changed areas, which is not fully exploited in the proposed method. The learning of LCLU transition types can potentially contribute to the recognition of semantic classes \cite{bruzzone1997iterative}. To model these transitions, more connections can be established between the CD unit and the temporal branches (in the DSCD-l architecture), which is left for future studies.

\bibliographystyle{IEEEtran}
\bibliography{refs}

\end{document}